%% file: UnifiedNeuralNetworks.tex
\documentclass[preprint,12pt]{elsarticle}

\usepackage{amsfonts,amssymb,amsmath,amsthm}
\usepackage{algorithmic}
\usepackage{algorithm}
\usepackage{array}
\usepackage{textcomp}
\usepackage{stfloats}
\usepackage{url}
\usepackage{verbatim}
\usepackage{graphicx}
\usepackage{subcaption}
\usepackage{color,pdfcolmk}
\usepackage[strings]{underscore}
\usepackage{dsfont}

\input{macrosJMLR}

\newtheorem{theorem}{Theorem}
\newtheorem{lemma}{Lemma}
\newtheorem{remark}{Remark}
\newtheorem{proposition}{Proposition}

\newtheorem{definition}{Definition}

\usepackage[shortlabels]{enumitem}

\begin{document}

\begin{frontmatter}



\title{A Unified and Constructive Framework for the Universality of Neural Networks}

\author[UT-mech,Oden]{Tan Bui-Thanh}
\ead{tanbui@oden.utexas.edu}

\affiliation[UT-mech]{organization={Department of Aerospace Engineering and
   Engineering Mechanics, The University of Texas at Austin},
             city={Austin},
             state={Texas},
             country={USA}}
             
\affiliation[Oden]{organization={Oden Institute for Computational Engineering and Sciences, The University of Texas at Austin},
             city={Austin},
             state={Texas},
             country={USA}}



\begin{abstract}
One of the reasons why many neural networks are capable of replicating complicated tasks or functions is their universal property. Though the past few decades have seen tremendous advances in theories of neural networks, a single constructive framework for neural network universality remains unavailable. This paper is the first effort to provide a unified and constructive framework for the universality of a large class of activation functions including most of existing ones. At the heart of the framework is the concept of neural network approximate identity (nAI). The main result is: {\em any nAI activation function is universal}. It turns out that most of existing activation functions are nAI, and thus universal in the space of continuous functions on compacta. The framework 
induces {\bf several advantages} over the contemporary
  counterparts.
First, it is constructive with elementary means from functional analysis, probability theory, and numerical analysis. Second, it is the first unified attempt that is valid for most of existing activation functions. Third, as a by product, the framework provides the first universality proof for some of the existing activation functions including Mish, SiLU, ELU, GELU, and etc. Fourth, it provides new proofs for most activation functions. Fifth, it discovers new activation functions with guaranteed universality property. 
    Sixth, for a given activation and error tolerance, the framework provides precisely the architecture of the corresponding one-hidden neural network with predetermined number of neurons, and the values of weights/biases. Seventh, the framework allows us to abstractly present the first universal approximation with favorable non-asymptotic rate.
\end{abstract}


\begin{highlights}
\item  Unlike existing universality approaches, our framework is constructive using elementary means from functional analysis, probability theory, non-asymptotic analysis, and numerical analysis. 
\item While existing approaches are either technical or specialized for particular activation functions, our framework is the first unified attempt that is valid for most of the existing activation functions and beyond.
\item The framework provides the first universality proof for some of the existing activation functions including Mish, SiLU, ELU, GELU, and etc;
\item The framework provides new proofs for all activation functions
\item The framework discovers and facilitates the discovery of new activation functions with guaranteed universality property. Indeed, any activation\textemdash whose $\k$th derivative, with $\k$ being an integer, is integrable and essentially bounded\textemdash is universal. 
In that case, the activation function and all of its $j$th derivative, $j = 1,\hdots,\k$ are not only a valid activation but also universal.
\item For a given activation and error tolerance, the framework provides precisely the architecture of the corresponding one-hidden neural network with a predetermined number of neurons, and the values of weights/biases;
\item The framework is the first that allows for abstractly presenting the universal approximation with the favorable non-asymptotic rate of $\N^{-\half}$, where $\N$ is the number of neurons. This provides not only theoretical insights into the required number of neurons but also guidance on how to choose the number of neurons to achieve a certain accuracy with controllable successful probability. Perhaps more importantly, it shows that neural network may fail, with non-zero probability, to provide the desired testing error in practice when an architecture is selected;
\item Our framework also provides insights into the developments, and hence providing constructive derivations, of some of the existing approaches. 
\end{highlights}


\end{frontmatter}


\section{Introduction}
{H}{uman}  brain consists of networks of billions  of neurons, each of which, roughly speaking, receives information\textemdash electrical pulses \textemdash from other neurons via dendrites, processes the information using soma, is activated by difference of electrical potential, and
passes  the output along its axon to other neurons through synapse. Attempt to understand the extraordinary ability of the brain in categorizing, classifying, regressing, processing, etc information has inspired numerous scientists to develop computational models to mimic brain functionalities.
Most well-known is perhaps the McCulloch-Pitts model \cite{McCulloch90}, which is also called perceptron by Rosenblatt who extended McCulloch-Pitts model to networks of artificial neurons capable of learning from data  \cite{Rosenblatt58}. {\em The question is if such a network could mimic some of the brain capability}, such as learning to classify. The answer lies in the fact that perceptron networks can represent Boolean logic functions exactly. From a mathematical point of view, perceptron networks with Heaviside activation functions compute step functions, linear combinations of which form the space of simple functions which in turn is dense in the space of measurable functions \cite{Durrett2010,Royden18}. That is, linear combination of perceptions can approximate a measurable function to any desired accuracy \cite{Lippmann87, Cotter90}: {\em the first universal approximation for neural networks.} The universal approximation capability partially ``explains" why human brains can be trained to virtually learn any tasks.

For training one-hidden layer networks, Widrow-Hoff rule \cite{widrow60} can be used for supervised learning. For many-layer ones, the most popular approach is back-propagation \cite{Rumelhart87} which requires the derivative of activation functions. Heaviside function is, however, not differentiable in the classical sense. A popular smooth approximation is the standard logistic (also called sigmoidal) activation function which is infinitely differentiable. {\em The question is now: are neural networks of sigmoidal function universal?} An answer to this question was first presented by Cybenko \cite{Cybenko1989} and Funahashi \cite{FUNAHASHI1989}. The former, though non-constructive (a more constructive proof was then provided in \cite{Chen92} and revisited in \cite{Costarelli14}), elegantly used the Hahn-Banach theorem to show that sigmoidal neural networks (NNs) with one hidden layer is dense in the space of continuous functions on compacta. The latter  used Fourier transform and an integral formulation for  integrable function with bounded variation \cite{Irie88} to show the same results for NNs with continuous, bounded, monotone increasing sigmoidal activation function.
Recognizing NN output as an approximate back-projection operator, \cite{Carroll89} employed inverse Radon transformation to show the universal approximation property in space of squared integrable functions.
Making use of the classical Stone-Weierstrass approximation theorem \cite{HORNIK89} successfully  proved NNs with non-decreasing sigmoidal function are dense in space of continuous functions over compacta and dense in space of measurable functions.
Using the separability of space of  continuous functions over compacta  \cite{Maiorov1999LowerBF} constructed a strictly increasing analytic sigmoidal function that is universal. 
The work in \cite{KURKOVA92}, based on a Kolmogorov theorem, showed that any continuous function on hypercubes can be approximated well with two-hidden layer NNs of sigmoidal functions. 

Though sigmoidal functions are natural continuous approximation to Heaviside function, and hence mimicking the activation mechanism of neurons, they are not the only ones having universal approximation property. Indeed,
\cite{MHASKAR92} showed, using distributional theory, that NNs with $k$th degree sigmoidal function are dense in space of continuous functions over compacta. 
Meanwhile, \cite{Gallant88} designed a cosine squasher activation function so that the output of one-hidden layer neural network is truncated Fourier series and thus can approximate square integrable functions to any desired accuracy.
Following and extending \cite{Cybenko1989}, \cite{HORNIK91} showed that NN is dense in $\L^\p$ with bounded non-constant activations function and in space of continuous functions over compacta 
with continuous bounded nonconstant activation functions.
Using the  Stone-Weierstrass theorem, \cite{Cotter90} showed that any network with activations (e.g. exponential, cosine squasher, modified sigma-pi, modified logistic, and step functions) that can transform product of functions into sum of functions is universal in the space of bounded measurable functions over compacta. The work in \cite{Stinchcombe90} provided universal approximation for bounded weights (and biases) with piecewise polynomial and superanalytic  activation functions (e.g. sine, cosine, logistic functions and piecewise continuous functions are superanalytic at some point with positive radius of convergence).  

One hidden-layer NN is in fact dense in space of continuous functions and $\L^\p$ if the activation function is not polynomial almost everywhere \cite{LESHNO93, pinkus1999}. Using Taylor expansion and Vandermonde determinant, \cite{Attali97} provided an elementary proof of  universal approximation for 
NNs  with $C^\infty$ activation functions. Recently,
\cite{YAROTSKY17} provides universal approximation theorem for multilayer NNs with ReLU activation functions using partition of unity and Taylor expansion for functions in Sobolev spaces. Universality of multilayer ReLU NNs can also be obtained using approximate identity and rectangular quadrature rule \cite{Moon21}. Restricting in Barron spaces \cite{Barron93,klusowski18}, the universality of ReLU NNs is a straightforward application of the law of large numbers \cite{Weinan19}. The universality of ReLU NNs can also be achieved by emulating finite element approximations \cite{he20,Opschoor20}.

  Unlike others, \cite{CARDALIAGUET1992} introduced $\B$ellshape function (as derivative of squash-type activation function, for example)  as a means to explicitly construct one-hidden layer NNs to approximate continuous function over compacta in multiple dimensions. More detailed analysis was then carried out in \cite{CHEN2009, CHEN2015}.     
  The idea was revisited and extended to establish universal approximation  in the uniform norm for tensor product sigmoidal and hyperbolic tangent activations \cite{Anastassiou01,Anastassiou00,ANASTASSIOU2011,ANASTASSIOU2011b,ANASTASSIOU2011tanhnD}. 
  Similar approach was also taken in \cite{COSTARELLI2015} using cardinal B-spline
  and in \cite{Costarelli2015sigmoid} using the distribution function as sigmoid but for a family of sigmoids with a certain of decaying-tail conditions. 
  
  While it is sufficient for most universal approximation results to hold when each weight (and bias) varying over the whole real line $\R$, this is not necessary. In fact, universal approximation results for continuous function on compacta can also be obtained using finite set of weights \cite{CHUI1992,ITO91,Ismailov12}.
It is quite striking that one-hidden layer with only one neuron is enough for universality: \cite{Guliyev16} constructs a smooth, sigmoidal, almost monotone activation function so that one-hidden layer with one neuron can approximate any continuous function over any compact subset in $\R$ to any desired accuracy.

 Universal theorems with convergence rate for sigmoidal and others NNs have also been established.  Modifying the proofs in \cite{Chen92,MHASKAR92}, \cite{Debao93} showed that, in one dimension, the error  incurred by sigmoidal NNs with $\N$ hidden units scales as $\mc{O}\LRp{\N^{-1}}$ in the uniform norm over compacta. The result was then extended to $\R$  for bounded continuous function \cite{HONG02}. Similar results were obtained for functions with bounded variations \cite{ GAO93} and with bounded $\phi$-variations \cite{Lewicki03}. For multiple dimensions,  \cite{Barron93} provided universal approximation for sigmoidal NNs in the space of functions with bounded first moment of the magnitude distribution of their Fourier transform with rate $\mc{O}\LRp{\N^{-1/2}}$ in the $\L^2$-norm, independent of dimensions. This result can be generalized to Lipschitz functions (with additional assumptions) \cite{LEWICKI04}. Using explicit NN construction in \cite{CARDALIAGUET1992},
 \cite{Anastassiou01,Anastassiou00,ANASTASSIOU2011,ANASTASSIOU2011b,ANASTASSIOU2011tanhnD,CHEN2009, CHEN2015,Costarelli2015sigmoid} provided convergence rate of $\mc{O}\LRp{\N^{-\alpha}}$, $0 < \alpha < 1$ in the uniform norm for H\"older continuous functions with exponent $\alpha$.
 Recently,
\cite{SIEGEL2020313} has revisited universal approximation theory with rate  $\mc{O}\LRp{\N^{-1/2}}$ in Sobolev norms for smooth activation functions with polynomial decay condition on all derivatives. This improves/extends the previous similar work for sigmoidal functions in \cite{Barron93} and exponentially decaying activation functions in \cite{HornikEtAl94}. The setting in \cite{SIEGEL2020313} is valid for  sigmoidal, arctan, hyperbolic tangent, softplus, ReLU, Leaky ReLU, and $k$th power of ReLU,  as their central differences satisfy polynomial decaying condition. However, the result is only valid for smooth functions in high-order Barron spaces. For activation functions without decay but essentially bounded and having bounded Fourier transform on some interval (or having bounded variation), the universal approximation results with slower rate  $\mc{O}\LRp{\N^{-1/4}}$ in the $\L^2$-norm can be obtained for first order Barron space. These rates can be further improved 
using stratified sampling \cite{MAKOVOZ96,klusowski18}. Convergence rates for ReLU NNs in Sobolev spaces has been recently established using finite elements \cite{he20,Opschoor20}.

The main objective of this paper is to provide the first constructive and unified frameworks for a large class of activation functions including most of existing ones. At the heart of this new framework is the introduction of the neural network approximate identity (nAI) concept. The important consequence is: {\em any nAI activation function is universal}.
The following are the main contributions: i) Unlike existing works, our framework is constructive using elementary means from functional analysis, probability theory, non-asymptotic analysis, and numerical analysis. 
ii) While existing approaches are either technical or specialized for particular activation functions, our framework is the first unified attempt that is valid for most of the existing activation functions and beyond;
         iii) The framework provides the first universality proof for some of the existing activation functions including Mish, SiLU, ELU, GELU, and etc;
         iv) The framework provides new proofs for all activation functions;
         v) The framework discovers and facilitates the discovery of new activation functions with guaranteed universality property. Indeed, any activation\textemdash whose $\k$th derivative, with $\k$ being an integer, is integrable and essentially bounded\textemdash is universal. 
In that case, the activation function and all of its $j$th derivative, $j = 1,\hdots,\k$ are not only a valid activation but also universal.
         vi) For a given activation and error tolerance, the framework provides precisely the architecture of the corresponding one-hidden neural network with a predetermined number of neurons, and the values of weights/biases;
         vii) The framework is the first that allows for abstractly presenting the universal approximation with the favorable non-asymptotic rate of $\N^{-\half}$, where $\N$ is the number of neurons. This provides not only theoretical insights into the required number of neurons but also guidance on how to choose the number of neurons to achieve a certain accuracy with controllable successful probability. Perhaps more importantly, it shows that neural network may fail, with non-zero probability, to provide the desired testing error in practice when an architecture is selected;
         viii) Our framework also provides insights into the developments, and hence providing constructive derivations, of some of the existing approaches.

The paper is organized as follows. Section \secref{notations} introduces conventions and notations used in the paper. Elementary facts about convolution and approximate identities that are useful for our purposes are presented in section \secref{cAI}. Section \secref{quadrature} recalls quadrature rules in terms of Riemann sums for continuous functions on compacta and their error analysis using moduli of continuity. This follows by a unified abstract framework for universality in section \secref{aAI}. The key to achieve this is to introduce the concept of neural network approximate identity (nAI), which immediately provides an abstract universality result in Lemma \lemref{AI}. This abstract framework reduces universality proof to nAI proof.
Section \secref{manyActivations} shows that most of existing activations are nAI. This includes the family of rectified polynomial units (RePU) in which the parametric and
leaky ReLUs are a member,  a family of generalized sigmoidal functions in which the standard sigmoidal, hyperbolic tangent and softplus are a member, the exponential
linear unit (ELU), the Gaussian error linear unit (GELU), 
the sigmoid linear unit (SiLU), and the Mish.  It is the nAI proof of the
Mish activation function that guides us to devise a general framework for a large class of nAI functions (including all activations in this paper and beyond)
in section \secref{generalAI}. Abstract universal approximation result with non-asymptotic rates is presented in section \secref{rates}.
Section \secref{conclusion} concludes the paper.

\section{Notations}
\seclab{notations}
This section describes notations used in the paper.
We reserve lower case roman letters for scalars or scalar-valued function. Boldface lower case roman letters are for vectors with components denoted by subscripts.  We denote by $\R$ the set of real numbers and by $*$ the convolution operator. For $\x \in \Rn$, where $\n \in \mathbb{N}$ is the ambient dimension, $\snor{\x}_p := \LRp{\sum_{i=1}^\n\snor{\x_i}^\p}^{1/\p}$ denotes the standard $\ell^p$ norm in $\Rn$. We conventionally write $\f\LRp{\x} := \f\LRp{\x_1,\hdots,\x_n}$
    and  $\nor{\f}_\p := \LRp{\intR \snor{f\LRp{\x}}^\p\,d\mu\LRp{\x}}^{1/\p}$ for $1\le \p <\infty$ and $\mu$ is the Lebesgue measure in $\R^n$. For $p = \infty$, $\nor{\f}_\infty := \text{ess}\sup_{\Rn}\snor{f} := \LRc{M > 0: \mu\LRc{\x:\snor{f(x)} > M} = 0}$. For simplicity, we use $d\x$ in place of $d\mu\LRp{\x}$. Note that we also use $\nor{f}_\infty$ to denote the uniform norm of continuous function $f$. We 
    define $\L^\p:=\L^\p\LRp{\Rn} := \LRc{f: \nor{f}_p < \infty}$ for $1\le \p \le \infty$, $\C\LRp{\K}$ as the space of continuous functions on $\K \subseteq\Rn$, $\C_0\LRp{\Rn}$ as the space of continuous functions vanishing at infinity,  $\C_b\LRp{\Rn}$ as the space of bounded functions in $\C\LRp{\Rn}$, and 
    $\C_c\LRp{\K}$ 
    as the space of functions in $\C\LRp{\K}$ with compact support.

\section{Convolution and Approximate Identity}
\seclab{cAI}
The mathematical foundation for our framework is  approximate identity, which relies on convolution. Convolution has been used by many authors including \cite{pinkus1999,COSTARELLI2015,LESHNO93} to assist in proving universal approximation theorems. Approximate identity generated by ReLU function has been used in \cite{Moon21} for showing ReLU universality in $\L^\p$.
In this section we collect some important results from convolution and approximate identity that are useful for our developments (see, e.g., \cite{folland1999real} for a comprehensive treatment). Let $\tau_\y\f\LRp{\x} := \f\LRp{\x-\y}$ be the translation operator. The following is standard. 
\begin{proposition}
$\f$ is uniformly continuous iff $\lim_{\y\to 0}\nor{\tau_\y\f - f}_\infty = 0$.
\propolab{translationOp}
\end{proposition}

Let $\f,\g: \Rn \to \R$ be two measurable functions in $\Rn$, their convolution is defined as
\begin{equation*}
    \f*\g := \intR \f\LRp{\x - \y}\g\LRp{\y}\,d\y,
\end{equation*}
when the integral exists. 
We are interested in the conditions under which $\f\LRp{\x-\y} \g\LRp{\y}$ (or $\f\LRp{\y} \g\LRp{\x-\y}$) is integrable for almost every (a.e.) $\x$, as our approach requires a discretization of the convolution integral. Below 
is a relevant case. 
\begin{lemma}
If $\f \in \C_0\LRp{\Rn}, \g \in \L^1\LRp{\Rn}$ then $\f * \g \in \C_0\LRp{\Rn}$.
\lemlab{convolution}
\end{lemma}
We are interested in $\g$ which is an approximate identity in the following sense.

\begin{definition}
A family of functions $\Beps \in \L^1\LRp{\Rn}$, where $\theta > 0$, is called an approximate identity (AI) if: i) The family is bounded in the $\L^1$ norm, i.e., $\nor{\Beps}_1 \le C$ for some $C > 0$; ii) $\intR\Beps\LRp{\x}\,d\x =1$ for all $\theta > 0$; and iii) $\int_{\nor{\x} > \delta} \snor{\Beps\LRp{\x}}\,d\x \longrightarrow 0$ as $\theta \longrightarrow 0$, for any $\delta > 0$.
\defilab{AI}
\end{definition}
An important class of approximate identity is obtained by rescaling $\L^1$ functions.
\begin{lemma}
If $\g \in \L^1\LRp{\Rn}$ and $\intR\g\LRp{\x}\,d\x = 1$, then $\Beps\LRp{\x} := \frac{1}{\theta^\n}\g\LRp{\frac{\x}{\theta}}$ 
is an AI. 
\lemlab{AIscaling}
\end{lemma}

\begin{lemma}
For any $ \f \in \C_0\LRp{\Rn}$,  $\lim_{\theta \to 0}\nor{\f * \Beps - \f}_\infty = 0$, where $\Beps$ is an AI. Consequently, for any $\K$ being a compact subset of $\Rn$, $\lim_{\theta \to 0}\nor{\Ind_{\K}\LRp{\f * \Beps - \f}}_\infty = 0$, where $\Ind_\K$ is the indicator/characteristic function of $\K$.

\lemlab{AI}
\end{lemma}
\begin{proof}
We briefly present the proof  here as we will use a similar approach to estimate approximate identity error later. Since $f \in C_0\LRp{\Rn}$ it resides in $\L^\infty\LRp{\Rn}$ and is uniformly continuous. As a consequence of Proposition \proporef{translationOp}, for any $\varepsilon >0$, there exists $\delta > 0$ such that $\nor{\tau_\y\f - f}_\infty \le \varepsilon$ for $\nor{\y} \le \delta$.
We have
\begin{multline*}
    \nor{\f * \Beps - \f}_\infty  \le \intR\nor{\tau_\y\f - \f}_\infty\snor{\Beps\LRp{\y}}\,d\y 
    \le \varepsilon\int_{\nor{\y} \le \delta} \snor{\Beps\LRp{\y}}\,d\y \\+ 
    2\nor{\f}_\infty\int_{\nor{\y} > \delta} \snor{\Beps\LRp{\y}}\,d\y, 
\end{multline*}
where we have used the 
second property in Proposition \proporef{translationOp}. The assertion is now clear as $\varepsilon$ is arbitrarily small and by the third condition of approximate identity.
\end{proof}

\section{Quadrature rules for continuous functions on bounded domain}
\seclab{quadrature}
Recall that for a bounded function on a compact set, it is Riemann integrable if and only if it is continuous almost everywhere. In that case, Riemann sums converge to Riemann integrals as the corresponding partition size approaches 0. It is well-known (see, e.g. \cite{Baker68,DavisRabinowitz84}, and references therein) that most common numerical quadrature rules\textemdash including trapezoidal, some Newton-Cotes formulas, and Gauss-Legendre quadrature formulas\textemdash are Riemann sums. In this section, we use the Riemann sum interpretation of quadrature rule to approximate integrals of bounded functions. The goal is to characterize quadrature error in terms of modulus of continuity. We first discuss quadrature error in one dimension and then extend the result to $\n$ dimensions.

We assume the domain of interest is $\LRs{-1,1}^\n$. 
Let $\f \in {\C}\LRs{-1,1}^\n :={\C}\LRp{\LRs{-1,1}^\n}$, 
$\P^\m := \LRc{\z^1,\hdots,\z^{\m+1}}$ a partition of $\LRs{-1,1}$, and $\Q_\m := \LRc{\xi^1,\hdots,\xi^\m}$ the collection of all ``quadrature points" such that $-1\le \z^j\le \xi^j\le \z^{j+1}\le 1$ for $j=1,\hdots,\m$. We assume that 
\[
\sum_{j=1}^\m\g\LRp{\xi^j}\LRp{\z^{j+1}-\z^j}
\]
be a valid Riemann sum, e.g. trapezoidal rule, 
and thus converging to $\int_{-1}^1
\g\LRp{\z}\,d\z$ as $\m$ approaches $\infty$.
We define a quadrature rule for $\LRs{-1,1}^\n$ as the tensor product of the aforementioned one dimensional quadrature rule, and thus
\begin{equation}
\S\LRp{\m,\f} := \sum_{
j^1,\hdots,
j^n = 1
}^\m\f\LRp{\xi^{j^1},\hdots,\xi^{j^n}}\Pi_{i=1}^\n\LRp{\z^{j^i+1}-\z^{j^i}}
\eqnlab{RiemannSum}
\end{equation}
is a valid Riemann sum for $\int_{\LRs{-1,1}^\n}\f\LRp{\x}\,d\x$.

Recall the modulus of continuity $\omega\LRp{\f,\h}$ 
of a function $\f$:
\begin{equation}
\omega\LRp{\f,\h} := 
\sup_{\snor{\z}_2\le h}\nor{\tau_{\z}\f - \f}_\infty,
\eqnlab{modCon}
\end{equation}
$\omega\LRp{\f,0} = 0$, and $\omega\LRp{\f,\h}$ is continuous with respect to $h$ at $0$ (due to the uniform continuity of $\f$ on $\LRs{-1,1}^\n$).
The following error bound for the tensor product quadrature rule is an extension of the standard one dimensional case \cite{Baker68}.
\begin{lemma}
Let $\f \in \C\LRs{-1,1}^n$.
Then\footnote{The result can be marginally improved using the common refinement for $\P_\m$ and $\Q_\m$, but that is not important for our purpose.} 
\[
\snor{\S\LRp{\m,\f} - \int_{\LRs{-1,1}^\n}
\f\LRp{\x}\,d\x} \le 2^\n\omega\LRp{\f,\snor{\P_\m}},
\]
where $\snor{\P_\m}$ is the norm of the partition $\P_\m$.
\lemlab{QuadratureError}
\end{lemma}


\section{Error Estimation for Approximate Identity with quadrature}
\seclab{errorAI}
We are interested in approximating functions in $\C_c\LRs{-1,1}^\n$ 
using neural networks.  
How to extend the results to $\C\LRs{-1,1}^\n$ is given in \ref{extension}. At the heart of our framework is to first
approximate any function $\f$ in $\C_c\LRs{-1,1}^\n$ with an approximate identity $\Beps$ and then numerically integrate the convolution integral $\f * \Beps$ with quadrature (later with Monte Carlo sampling in Section \secref{rates}). This extends the similar approach in \cite{Moon21} for ReLU activation functions in several directions: 1) we rigorously account for the approximate identity and quadrature errors, 2) our unified framework holds for most activation functions (see Section \secref{manyActivations}) using the network approximate identity (nAI) concept introduced in Section \secref{aAI}, 4) we identify sufficient conditions under which an activation is nAI (see Section \secref{generalAI}), and 5) we provide non-asymptotic rate of convergence (see Section \secref{rates}). Moreover, this procedure is the key to our unification of neural network universality.


Let $\f \in \C_c\LRs{-1,1}^\n$ and $\Beps$ be an approximate identity. 
From Lemma \lemref{AI} we know that $\f * \Beps$ converges to $\f$ uniformly as $\theta \to 0$. We, however, are interested in estimating the error $\nor{\f * \Beps - \f}_\infty$ for a given $\theta$. From the proof of Lemma \lemref{AI}, for any $\delta > 0$, we have 
\begin{multline*}
    \nor{\f * \Beps - \f}_\infty  \le \intR\nor{\tau_\y\f - \f}_\infty\snor{\Beps\LRp{\y}}\,d\y \\
    \le \omega\LRp{\f,\delta}\int_{\nor{\y} \le \delta} \snor{\Beps\LRp{\y}}\,d\y + 
    2\nor{\f}_\infty\int_{\nor{\y} > \delta} \snor{\Beps\LRp{\y}}\,d\y 
     \le \omega\LRp{\f,\delta} + 2\nor{\f}_\infty \T\LRp{\Beps,\delta}, 
\end{multline*}
where $\T\LRp{\Beps,\delta} := \int_{\nor{\y} > \delta} \snor{\Beps\LRp{\y}}\,d\y$ is the tail mass of $\Beps$.

Next, for any $\x \in \LRs{-1,1}^\n$, we approximate
\[
\f * \Beps\LRp{\x} = \int_{\Rn}\Beps\LRp{\x - \y}\f\LRp{\y}\,d\y=  \int_{\LRs{-1,1}^\n}\Beps\LRp{\x - \y}\f\LRp{\y}\,d\y
\]
with the Riemann sum 
defined in \eqnref{RiemannSum}.
The following result is obvious by triangle inequality, continuity of $\omega\LRp{\f,\h}$ at $\h = 0$, the third condition of the definition of $\Beps$, and Lemma \lemref{QuadratureError}.
\begin{lemma}
Let $\f \in \C_c\LRs{-1,1}^\n$, $\Beps$ be an approximate identity, and $\S\LRp{\m,\f * \Beps}\LRp{\x}$ be the quadrature rule \eqnref{RiemannSum} for $\f * \Beps$ with the partition $\P^\m$ and quadrature point set $\Q^\m$. Then, for any $\delta > 0$, and $\varepsilon > 0$, there holds
\begin{equation}
      \f\LRp{\x} - \S\LRp{\m,\f * \Beps}\LRp{\x} = \mc{O}\LRp{\omega\LRp{\f,\delta} + \omega\LRp{\f,\snor{\P^\m}} + \T\LRp{\Beps,\delta} }.
  \eqnlab{error}
\end{equation}
In particular, for any $\varepsilon > 0$, there exist a sufficiently small $\snor{\P^\m}$  (and hence  large $\m$), and sufficiently small $\theta$ and $\delta$ such that 
\[
\nor{\f\LRp{\x} - \S\LRp{\m,\f * \Beps}\LRp{\x}}_\infty \le \varepsilon.
\]
\lemlab{abstractUniversal}
\end{lemma}
\begin{remark}
Aiming at using only elementary means, we use Riemann sum to approximate the integral, which does not give
the best possible convergence rates. To improve the rates, Section 
\secref{rates} resorts to Monte Carlo sampling to not only reduce the number of ``quadrature" points but also to obtain a total number of points independent of the ambient dimension $\n$. 
\end{remark}

\section{An abstract unified framework for universality}
\seclab{aAI}
\begin{definition}[Network Approximate Identity function]
A univariate function $\sigma\LRp{x}: \R \to \R$ admits a network approximate identity (nAI) if there exist $1\le \k \in \mathbb{N}$, $\LRc{\alpha_i}_{i=1}^\k \subset \R$,
$\LRc{\w^i}_{i=1}^\k \subset \R$, and $\LRc{\b_i}_{i=1}^\k \subset \R$ such that
\begin{equation}
\g\LRp{\xn} := \sum_{i=1}^\k\alpha_i\sigma\LRp{\w^i\xn + \b_i} \in \L^1\LRp{\R}.
\eqnlab{nAI}
\end{equation}
\defilab{nAI}
\end{definition}
Thus, up to a scaling, $\g\LRp{\xn}$ is an approximate identity.
\begin{lemma}
If a univariate function $\sigma\LRp{x}: \R \to \R$ is a nAI, then it
is universal in $\C_c\LRs{-1,1}$.
\end{lemma}
\lemlab{nAI}
\begin{proof}
From nAI definition \defiref{nAI}, there exist $\k\in\mathbb{N}$ and $\alpha_i,\w^i,\b_i$, $i=1,\hdots,\k$ such that \eqnref{nAI} holds. After rescaling $\g\LRp{\xn}/\nor{\g\LRp{\xn}}_{\L^1\LRp{\R}}$ we infer from  Lemma \lemref{AIscaling}  that  $\Beps\LRp{\xn} : = \frac{1}{\theta}\g\LRp{\frac{\xn}{\theta}}$ 
is an approximate identity. Lemma \lemref{abstractUniversal} then asserts that for any desired accuracy $\varepsilon > 0$, there exist a partition $\snor{\P^\m}$, and sufficiently small $\theta$ and $\delta$ such that 
\begin{equation}
 \S\LRp{\m,\f * \Beps}\LRp{\xn} =  \frac{1}{\theta}\sum_{j=1}^\m\LRp{\zn^{j+1}-\zn^{j}} \f\LRp{\xi^{j}}
\sum_{\ell=1}^\k\alpha_\ell\sigma\LRs{\frac{\w^\ell}{\theta}\LRp{\xn-{\xi^{j}}} + \b_\ell}
\eqnlab{NNf}
\end{equation}
is within $\varepsilon$ from any $\f\LRp{\xn} \in \C_c\LRs{-1,1}$ in the uniform norm.
\end{proof}

We have explicitly constructed a one-hidden layer neural network $\S\LRp{\m,\f * \Beps}\LRp{\xn}$ with an arbitrary nAI activation $\sigma$ in \eqnref{NNf} to approximate well any continuous function $\f\LRp{\xn}$ with compact support in $\LRs{-1,1}$. A few observations are in order: i) if we know the modulus of continuity of $\f\LRp{\xn}$ and the tail behavior of $\Beps$ (from property of $\sigma$), we can precisely determine the total number of quadrature points $\m$, the scaling $\theta$, and the cut-off radius $\delta$ in terms of $\varepsilon$ (see Lemma 
\lemref{abstractUniversal}). That is, the topology of the network is completely determined; ii) The weights and biases of the network are also readily available from the nAI property of $\sigma$, the quadrature points, and $\theta$; iii) the coefficients of the output layer is also pre-determined by nAI (i.e. $\alpha_\ell$) and the values of the unknown function $\f$ at the quadrature points; and iv) {\bf any nAI activation function is universal in $\C_c\LRs{-1,1}$}.

Clearly the Gaussian activation function $e^{-\xn^2}$ is an nAI with $\k = 1$, $\alpha_1 = 1$, $\w_1 = 1$ and $\b_1 = 0$.
The interesting fact is that, as we shall prove in Section \secref{manyActivations}, most existing activation functions, though not integrable by themselves, are nAI. 

\begin{remark}
    It is important to point out that the universality in Lemma \lemref{nAI} for any nAI activation is in one dimension. In order to extend the universality result to $n$ dimensions for an nAI activation function, we shall deploy a special $\n$-fold composition of its one-dimensional approximate identity \eqnref{nAI}. Section \secref{manyActivations} provides explicit constructions of such $\n$-fold composition for many nAI activation functions, and Section \secref{generalAI} extends the result to abstract nAI function with appropriate conditions. 
\end{remark}

\begin{remark}
Our framework also helps understand 
some of the existing universal approximation approaches in a more constructive manner. For example, the constructive proof in \cite{Chen92} 
is not entirely constructive as the key step in equation (4), where the authors introduced the neural network approximation function $\g\LRp{\xn}$, is not constructive. Our framework can provide a constructive derivation of this step. Indeed, by applying a summation by part, one can see that $\g\LRp{\xn}$ resembles a quadrature approximation of the convolution of $\f\LRp{\xn}$ and the derivative (approximated by forward finite difference) of scaled sigmoidal function. Since the derivative of sigmoidal function, up to a scaling factor, is nAI (see Section \secref{generalAI}), 
the convolution of $\f\LRp{\xn}$ and a scaled sigmoidal derivative can approximate $\f\LRp{\xn}$ to any desired accuracy.

Another important example is the pioneered work in \cite{CARDALIAGUET1992} that has been extended in many directions in
 \cite{Anastassiou01,Anastassiou00,ANASTASSIOU2011,ANASTASSIOU2011b,ANASTASSIOU2011tanhnD,CHEN2009, CHEN2015,Costarelli2015sigmoid, Costarelli14}. Though the rest of the approach is constructive, the introduction of the neural network operator is rather mysterious. From our framework point of view, this is no longer the case as the neural network operator is nothing more than a quadrature approximation of the convolution of $\f\LRp{\xn}$ and $\B$ell shape functions constructed from  activation functions under consideration. Since these $\B$ell shape functions fall under the nAI umbrella, the introduction of neural network operator is now completely justified and easily understood.
 {We would like to point out that our work does not generalize/extend the work in \cite{CARDALIAGUET1992}.  
    Instead,  we aim at developing a new analytical framework that achieves unified and broad results that are not possible with the contemporary counterparts, while being more constructive and simpler. 
    The
    {$\B$ell shape function idea is not sufficient to construct our framework}. Indeed, the work in \cite{CARDALIAGUET1992} and its extensions
 \cite{Anastassiou01,Anastassiou00,ANASTASSIOU2011,ANASTASSIOU2011b,ANASTASSIOU2011tanhnD,CHEN2009, CHEN2015,Costarelli2015sigmoid, Costarelli14} have been limited to a few special and standard squash-type activation functions such as sigmoid and arctangent functions. 
    Our work\textemdash thanks to other new ingredients such as convolution, numerical quadrature, Monte-Carlo sampling, and finite difference together with the new nAI concept\textemdash breaks the barrier.
 Our framework thus provides not only a new constructive method but also new insights into the work in \cite{CARDALIAGUET1992} as its special case. }
\end{remark}

\section{Many existing activation functions are nAI}
\seclab{manyActivations} 
We now exploit properties of
 each activation (and its family, if any) under consideration to
show that they are nAI for $\n=1$. That is, these activations
generate one-dimensional approximate identity, which in turn shows
that they are universal by Lemma \lemref{nAI}.  
To show that they are also universal in $\n$ dimensions\footnote{Note that by Proposition 3.3 of \cite{pinkus1999}, with an additional assumption on the denseness of the set of ridge functions in $\C\LRp{\Rn}$, we can conclude that they are also universal in $\n$ dimensions. This approach, however, does not provide an explicit construction of the corresponding neural networks.}
we extend a constructive and explicit approach from \cite{Moon21} that was done for ReLU activation function. 
We start
with the family of rectified polynomial units (RePU)\textemdash also parametric and
leaky ReLUs\textemdash with many
properties, 
then family of sigmoidal functions, then the exponential
linear unit (ELU), then the Gaussian error linear unit (GELU), then
the sigmoid linear unit (SiLU), and we conclude with the Mish
activation with least properties. 
The proofs for all results in this section,  together with figures demonstrating the $\B$ functions for each activation, can be found in Appendix
\secref{proofs} and Appendix \secref{figures}.

\subsection{Rectified Polynomial Units (RePU) is an nAI}
\seclab{relu}
Following \cite{RePU19, MHASKAR92} we define the rectified polynomial unit (RePU) as
\[
\RePU\LRp{\q; x} = 
\begin{cases}
x^\q & \text{if } x\ge 0, \\
0 & \text{otherwise},
\end{cases}
\quad \q \in \mbb{N},
\]
which reduces to the standard rectilinear (ReLU) unit \cite{ReLU10} when $\q = 1$, the Heaviside unit when $\q = 0$, and the rectified cubic unit \cite{deepRitz18} when $\q = 3$.

The goal of this section is to construct an integrable linear combination of \RePU\ activation function for a given $\q$. 
We begin with constructing  compact supported $\mc{B}$ell-shape functions\footnote{
$\mc{B}$ell-shape functions seemed to be first coined in \cite{CARDALIAGUET1992}.}  from $\RePU$ in one dimension. Recall the binomial coefficient notation 
$
\begin{pmatrix}
\r \\ k
\end{pmatrix}
= \frac{\r!}{\LRp{\r-k}!k!},
$
 and the
central finite differencing operator with stepsize $\h$
$
\deltah\LRs{\f}\LRp{\xn} := \f\LRp{\xn + \frac{\h}{2}} - \f\LRp{\xn-\frac{\h}{2}}
$
(see Remark \remaref{FD} for forward and backward finite differences).

\begin{lemma}[$\B$-function for \RePU\, in one dimension]
Let $\r := \q+1$. For any $\h \ge 0$, define
\[
{\B}\LRp{\xn,\h} := \frac{1}{\q!}\deltah^\r\LRs{\RePU}\LRp{\xn} = \frac{1}{\q !}\sum_{i=0}^{\r}\LRp{-1}^{i}\begin{pmatrix}
\r \\ i
\end{pmatrix}
\RePU\LRp{\q;\xn + \LRp{\frac{\r}{2}- i}\h}. 
\]
Then: i) $\B\LRp{\xn,\h}$ is piecewise polynomial of order at most $\q$ on each interval  $\LRs{k\h-\frac{\r\h}{2},(k+1)\h-\frac{\r\h}{2}}$ for $k = 0,\hdots,\r-1$; ii) $\B\LRp{\xn,\h}$ is $\LRp{\q-1}$-time differentiable for $\q \ge 2$, continuous for $\q = 1$, and discontinuous for $\q = 0$. Furthermore, $\supp\LRp{\B\LRp{\xn,\h}}$ is a subset $\LRs{-\frac{\r\h}{2},\frac{\r\h}{2}}$; and iii) $\B\LRp{\xn, \h}$ is even, non-negative, unimodal, 
    and $\int_{\R}\B\LRp{\xn, 1}\,d\xn = 1$.. 

\lemlab{RePU}
\end{lemma}

Though there are many approaches to construct $\B$-functions in $\n$ dimensions from $\B$-function in one dimension (see, e.g. \cite{CARDALIAGUET1992, Anastassiou01,Anastassiou00,ANASTASSIOU2011,ANASTASSIOU2011b,ANASTASSIOU2011tanhnD, COSTARELLI2015, Costarelli2015sigmoid}) 
that are realizable by neural networks,  inspired by \cite{Moon21} we construct $\B$-function in $\n$ dimensions by $\n$-fold composition of the one dimensional $\B$-function in Lemma \lemref{RePU}. By Lemma \lemref{abstractUniversal}, it follows that the activation under consideration is universal in $n$ dimensions. We shall carry out this $\n$-fold composition for not only \RePU{} but also the other activation functions. The same procedure for an abstract activation function will be presented in Section \secref{generalAI}.

\begin{theorem}[$\B$-function for \RePU\, in $\n$ dimensions]
Let $\r := \q+1$, and $\x =\LRp{\x_1,\hdots,\x_n}$. Define
$
\Bc\LRp{\x} = \B\LRp{\x_n,\B\LRp{\x_{\n-1},\hdots,\B\LRp{\x_1,1}}}.
$
The following hold:
\begin{enumerate}[i)]
    \item $\Bc\LRp{\x}$ is non-negative with compact support. In particular $\supp\LRp{\Bc\LRp{\x}} \subseteq X_{i=1}^\n \LRs{-b_i,b_i}$ where  $b_i  = \underbrace{\B\LRp{0,\hdots,\B\LRp{0,1}}}_{(i-1)-\text{time composition}}$, for $2\le i\le \n$, and $b_1 = \frac{\n}{2} $.
    
    \item $\Bc\LRp{\x}$ is even with respect to each component $\x_i$, 
    and unimodal with $\Bc\LRp{0} = \max_{\x\in\Rn}\Bc\LRp{\x}$.
    \item $\Bc\LRp{\x}$ is piecewise polynomial of order at most $\LRp{\n-i}\q$ in $\x_i$, $i =1,\hdots,\n$. Furthermore, $\Bc\LRp{\x}$ is $(\q-1)$-time differentiable in each $\x_i$, 
    $\intR \Bc\LRp{\x}\,d\x \le 1$ and $\Bc\LRp{\x} \in \L^1\LRp{\Rn}$.
\end{enumerate}
\theolab{RePUnD}
\end{theorem}

\begin{remark}
Note that Lemma \lemref{RePU} and Theorem \theoref{RePUnD} also hold for parametric ReLU \cite{he2015delving} and  leaky ReLU \cite{maas2013a} (a special case of parametric ReLU)  with $\r = 2$. In other words, parametric ReLU is an nAI, and thus universal.
\end{remark}




\subsection{Sigmoidal and related activation functions are nAI}
\seclab{sigmoid}
\subsubsection{Sigmoidal, hyperbolic tangent, and softplus activation functions}
Recall the sigmoidal 
and softplus functions \cite{glorot2011a} are, respectively, given by
\[
\sigma_s\LRp{\xn} := \frac{1}{1+e^{-x}}, \quad 
\text{ and }
\sigma_p\LRp{\xn} := \ln\LRp{1 + e^\xn}.
\]
It is the relationship between sigmoidal and hyperbolic tangent function, denoted as $\sigma_t\LRp{\xn}$,
\[
\sigma_s\LRp{\xn} = \half + \half\sigma_t\LRp{\frac{\xn}{2}}
\]
that allows us to construct their $\B$-functions a similar manner. In
particular, based on the bell shape geometry of the the derivative of sigmoidal and
hyperbolic tangent functions,
we apply the central finite differencing
with $\h > 0$ to obtain the corresponding $\B$-functions
\[
\B_s\LRp{\xn,\h} := \frac{1}{2}\deltah\LRs{\sigma_s}\LRp{\xn}, \quad
\text{and }
\B_t\LRp{\xn,\h} := \frac{1}{4}\deltah\LRs{\sigma_t}\LRp{\xn}.
\]
Since $\sigma_s\LRp{\xn}$ is the derivative of $\sigma_p\LRp{\xn}$, we apply central difference twice to obtain a $B$-function for $\sigma_p\LRp{\xn}$:
$\B_p\LRp{\xn,\h} = 
\deltah^2\LRs{\sigma_p}\LRp{\xn}$.
The following Lemma \lemref{sigmoid1D} and Theorem
\theoref{sigmoidnD} hold for $\B\LRp{\xn,\h}$ being either
$\B_s\LRp{\xn,\h}$ or $\B_t\LRp{\xn,\h}$ or $\B_p\LRp{\xn,\h}$.

\begin{lemma}[$\B$-function for sigmoid, hyperbolic tangent, and solfplus in one dimension]
For $0 < h < \infty$, there hold: i) $\B\LRp{\xn,\h} \ge 0$ for all $\xn \in \R$, $\lim_{\snor{\xn} \to \infty}\B\LRp{\xn,\h}$ = 0, and $\B\LRp{\xn,\h}$ is even; and ii) $\B\LRp{\xn,\h}$ is unimodal
$\int_\R\B\LRp{\xn,\h}\,d\xn < \infty$, and $\B\LRp{\xn,\h} \in \L^1\LRp{\R}$.
\lemlab{sigmoid1D}
\end{lemma}

Similar to Theorem \theoref{RePUnD}, we construct $\B$-function in $\n$ dimensions by $\n$-fold composition of the one dimensional $\B$-function in Lemma \lemref{sigmoid1D}.
\begin{theorem}[$\B$-function for sigmoid, hyperbolic tangent, and softplus in $\n$ dimensions]
Let $\x =\LRp{\x_1,\hdots,\x_n} \in \Rn$. Define
\[
\Bc\LRp{\x} = \B\LRp{\x_n,\B\LRp{\x_{\n-1},\hdots,\B\LRp{\x_1,1}}}.
\]
Then $\Bc\LRp{\x}$ is even with respect to each component $\x_i$, $i=1,\hdots,\n$, and unimodal with $\Bc\LRp{0} = \max_{\x\in\Rn}\Bc\LRp{\x}$. Furthermore, $\intR \Bc\LRp{\x}\,d\x \le 1$ and $\Bc\LRp{\x} \in \L^1\LRp{\Rn}$.
\theolab{sigmoidnD}
\end{theorem}

\subsubsection{Arctangent function}
We next discuss arctangent activation function 
$\sigma_a\LRp{\xn} := \arctan\LRp{\xn}$
whose shape is similar to sigmoid. Since its derivative has the bell shape geometry, we define the $\B$-function for arctangent function by approximating its derivative with central finite differencing, i.e.,
\[
\B_a\LRp{\xn,\h} = 
\frac{1}{2\pi}\deltah\LRs{\sigma_a}\LRp{\xn} =
\frac{1}{2\pi} \arctan\LRp{\frac{2\h}{1+\xn^2 -\h^2}},
\]
for $0 < \h \le 1$.
Then simple algebra manipulations show that Lemma \lemref{sigmoid1D} holds for $\B_a\LRp{\xn,\h}$ with
$\int_\R \B_a\LRp{\xn,\h}\,d\xn = \h$. Thus, Theorem \theoref{sigmoidnD} also holds for $\n$-dimensional arctangent $\B$-function
\[
\Bc\LRp{\x} = \B_a\LRp{\x_n,\B_a\LRp{\x_{\n-1},\hdots,\B_a\LRp{\x_1,1}}},
\]
as we have shown for the sigmoidal function.

\subsubsection{Generalized sigmoidal functions}

We extend the class of sigmoidal function in \cite{costarelli13} to  generalized sigmoidal (or ``squash") functions. 
\begin{definition}[Generalized sigmoidal functions]
We say $\sigma\LRp{\xn}$ is a generalized sigmoidal function if it satisfies the following conditions:
\begin{enumerate}[i)]
    \item $\sigma\LRp{\xn}$ is bounded, i.e.,   there exists a constant $\sigma_{\text{max}}$ such that $\snor{\sigma\LRp{\xn}} \le \sigma_{\text{max}}$ for all $\xn \in \R$.
    \item $\lim_{\xn \to \infty}\sigma\LRp{\xn} = L$ and $\lim_{\xn \to -\infty}\sigma\LRp{\xn} = \ell$.
    \item There exist $\xn^- < 0$ and $\alpha > 0$ such that ${\sigma\LRp{\xn} - \ell} = \mc{O}\LRp{\snor{\xn}^{-1-\alpha}}$ for $\xn < \xn^-$.
    \item There exist $\xn^+ > 0$ and $\alpha > 0$ such that ${L - \sigma\LRp{\xn} } = \mc{O}\LRp{\snor{\xn}^{-1-\alpha}}$ for $\xn > \xn^+$.
\end{enumerate}
\defilab{sigmoid}
\end{definition}

Clearly the standard sigmoidal, hyperbolic tangent, and arctangent activations
are members of the class of generalized sigmoidal functions.

\begin{lemma}[$\B$-function for generalized sigmoids in one dimension]
Let $\sigma\LRp{\xn}$ be a generalized sigmoidal function, and $\B\LRp{\xn,\h}$ be an associated $\B$-function defined as
\[
\B\LRp{\xn,\h} := \frac{1}{2\LRp{L -\ell}}\LRs{\sigma\LRp{\xn + \h} - \sigma\LRp{\xn -\h}}.
\]
\newcommand{\Z}{\mathbb{Z}}
Then the following hold for any $\h \in \R$: i) There exists a constant $C$ such that  $\snor{\B\LRp{\xn,\h}} \le C \LRp{\xn + \snor{\h}}^{-1-\alpha}$ for $\xn \ge \xn^+ + \snor{\h}$, and $\snor{\B\LRp{\xn,\h}} \le C \LRp{-\xn + \snor{\h}}^{-1-\alpha}$ for $\xn \le \xn^- - \snor{\h}$; and ii) For $x \in \LRs{m, M}$, $m, M \in \R$,  $\sum_{k =-\infty}^\infty\B\LRp{\xn + k\h,\h}$ converges uniformly to $\text{sign}\LRp{\h}$. Furthermore, $\int_{\R} \B\LRp{\xn,\h}\,d\xn = \h$, and $\B\LRp{\xn,\h} \in \L^1\LRp{\R}$.
\lemlab{generalizedSigmoid1D}
\end{lemma}

{\em Thus any generalized sigmoidal function is an nAI in one dimension. } Note that the setting in \cite{costarelli13}\textemdash in which
$\sigma\LRp{\xn}$ is non-decreasing, $L =1$, and $\ell = 0$\textemdash
is a special case of our general setting. In this less general
setting, it is clear that $\B\LRp{\xn,\h} \ge$ and thus $\int_{\R}
\B\LRp{\xn,\h}\,d\xn = \h$ is sufficient to conclude $\B\LRp{\xn,\h}
\in \L^1\LRp{\R}$ for any $\h > 0$. 
We will
explore this in the next theorem as it is not clear, at the moment of
writing this paper, how to show that the $B$-functions for generalized
sigmoids, as constructed below, reside in $\L^1\LRp{\Rn}$.

\begin{theorem}[$\B$-function for generalized sigmoids in $\n$ dimensions]
Suppose that $\sigma\LRp{\xn}$ is a non-decreasing generalized sigmoidal function.
Let $\x =\LRp{\x_1,\hdots,\x_n} \in \Rn$. Define
$\Bc\LRp{\x} = \B\LRp{\x_n,\B\LRp{\x_{\n-1},\hdots,\B\LRp{\x_1,1}}}$.
Then $\intR \Bc\LRp{\x}\,d\x = 1$ and $\Bc\LRp{\x} \in \L^1\LRp{\Rn}$.
\theolab{generalizedSigmoidnD}
\end{theorem}

\subsection{The Exponential Linear Unit (ELU) is nAI}
Following \cite{clevert2016fast,klambauer2017selfnormalizing} we define the Exponential Linear Unit (ELU) as
\[
\sigma\LRp{\xn} = 
\begin{cases}
\alpha \LRp{e^{\xn} - 1} &  \xn \le 0 \\
\xn &  \xn > 0,
\end{cases}
\]
for some $\alpha \in \R$. The goal of this section is to show that ELU is an nAI. Since the unbounded part of ELU is linear, Section \secref{relu} suggests us to define, for $\h \in \R$,
\[
 \B\LRp{\xn,\h} := \frac{\deltah^2\LRs{\sigma}\LRp{\xn}}{\gamma}, \text{ where } \gamma = \max\LRc{1+4\snor{\alpha},2 + \snor{\alpha}}.
\]

\begin{lemma}[$\B$-function for ELU in one dimension]
Let $\alpha, \h \in \R$, then
$\int_{\R} \B\LRp{\xn,\h}\,d\xn = \frac{\h^2}{\gamma}, \text{ and } \B\LRp{\xn,\h} \in \L^1\LRp{\R} $
Furthermore: $\B\LRp{\xn,\h} \le 1$ for $\snor{\h} \le 1$.
\lemlab{ELU1D}
\end{lemma}

\begin{theorem}[$\B$-function for ELU in $\n$ dimensions]
Let $\x =\LRp{\x_1,\hdots,\x_n} \in \Rn$. Define 
$\Bc\LRp{\x} = \B\LRp{\x_n,\B\LRp{\x_{\n-1},\hdots,\B\LRp{\x_1,1}}}$.
Then $\intR \snor{\Bc\LRp{\x}}\,d\x \le 1$ and thus  $\Bc\LRp{\x} \in \L^1\LRp{\Rn}$.
\theolab{ELUnD}
\end{theorem}

\subsection{The Gaussian Error Linear Unit (GELU) is nAI}
GELU, introduced in \cite{hendrycks2020gaussian}, is defined as
\[
\sigma\LRp{\xn} := \xn \Phi\LRp{\xn},
\]
where $\Phi\LRp{\xn}$ is the cummulative distribution function of standard normal distribution. Since the unbounded part of GELU is essentially linear, Section \secref{relu} suggests us to define 
\[
\B\LRp{\xn,\h}:= \deltah^2\LRs{\sigma}\LRp{\xn}, \quad \h \in \R.
\]

\begin{lemma}[$\B$-function for GELU in one dimension]
\begin{enumerate}[i)]
    \item $\B\LRp{\xn,\h}$ is an even function 
    in both $\xn$ and $\h$.
    \item $\B\LRp{\xn,\h}$ has two symmetric roots $\pm\xn^*$ with $\h < \xn^*  < \max\LRc{2,2\h}$, and $\snor{\B\LRp{\xn,\h}} \le \frac{1}{\sqrt{2\pi}} \h^2$,
\[
\int_{\R} \B\LRp{\xn,\h}\,d\xn = \h^2,
\text{ and } \int_{\R} \snor{\B\LRp{\xn,\h}}\,d\xn \le \frac{37}{10} \h^2.
\]
\end{enumerate}
\lemlab{GELU1D}
\end{lemma}

\begin{theorem}[$\B$-function for GELU in $\n$ dimensions]
Let $\x =\LRp{\x_1,\hdots,\x_n} \in \Rn$. Define
$
\Bc\LRp{\x} = \B\LRp{\x_n,\B\LRp{\x_{\n-1},\hdots,\B\LRp{\x_1,1}}}$,
then $\Bc\LRp{\x} \in \L^1\LRp{\Rn}$.
\theolab{GELUnD}
\end{theorem}

\subsection{The Sigmoid Linear Unit (SiLU) is an nAI}
SiLU, also known as sigmoid shrinkage or swish
\cite{hendrycks2020gaussian,ELFWING20183,ramachandran2017searching,Atto08}, is defined as
$
\sigma\LRp{\xn} := \frac{\xn}{1+e^{-\xn}}.
$
By inspection, the second derivative of $\sigma\LRp{\xn}$ is bounded
and its graph is quite close to bell shape. This suggests us to
define
\[
\B\LRp{\xn,\h}:= \deltah^2\LRs{\sigma}\LRp{\xn}, \quad \h \in \R.
\]
The proofs of the following results are similar to those of Lemma \lemref{GELU1D} and Theorem \theoref{GELUnD}.
\begin{theorem}[$\B$-function for SiLU in $n$ dimension]
\begin{enumerate}[i)]
    \item $\B\LRp{\xn,\h}$ is an even function 
    in both $\xn$ and $\h$.
    \item $\B\LRp{\xn,\h}$ has two symmetric roots $\pm\xn^*$ with $\h < \xn^*  < \max\LRc{3,2\h}$, and  $\snor{\B\LRp{\xn,\h}} \le \half \h^2$.
    \item 
$
\int_{\R} \B\LRp{\xn,\h}\,d\xn = \h^2,
\text{ and } \int_{\R} \snor{\B\LRp{\xn,\h}}\,d\xn \le \frac{26}{5} \h^2.
$
\item Let $\x =\LRp{\x_1,\hdots,\x_n}$.
Define
$\Bc\LRp{\x} = \B\LRp{\x_n,\B\LRp{\x_{\n-1},\hdots,\B\LRp{\x_1,1}}}$,
then $\Bc\LRp{\x} \in \L^1\LRp{\Rn}$.
\end{enumerate}
\theolab{SiLU}
\end{theorem}

\subsection{The Mish unit is an nAI} 
\seclab{Mish}
Mish unit, introduced in \cite{misra2020mish}, is defined as
\[
\sigma\LRp{\xn} := \xn\tanh\LRp{\ln\LRp{1+e^\xn}}.
\]
Due to its similarity with SiLU, we define
\[
\B\LRp{\xn,\h}:= \deltah^2\LRs{\sigma}\LRp{\xn}.
\]
Unlike any of the preceding activation functions 
it is not straightforward to manipulate
Mish analytically. This motivates us to devise a new approach to show
that Mish is nAI. As shall be shown in Section \secref{generalAI}, this approach allows us to unify
the nAI property for all activation functions.
We begin with the following result on the second
derivative of Mish.
\begin{lemma}
The second derivative of Mish, $\sigma^{(2)}\LRp{\xn}$,
is continuous, bounded, and integrable. That is, $\snor{\sigma^{(2)}\LRp{\xn}} \le M$ for some positive constant $M$ and  $\sigma^{(2)}\LRp{\xn} \in \L^1\LRp{\R}$.
\lemlab{Mish2ndDiff}
\end{lemma}

\begin{theorem}
Let $\h \in \R$, then the following hold: 
\begin{enumerate}[i)]
    \item  There exists 
    $0 < M< \infty$ such that $\snor{\B\LRp{\xn,\h}} \le M \h^2$,
 and
$
\int_{\R}\snor{\B\LRp{\xn,\h}}\,d\xn \le \nor{\sigma^{(2)}}_{\L^1\LRp{\R}} \h^2.
$
\item Let $\x =\LRp{\x_1,\hdots,\x_n}$.
Define
$
\Bc\LRp{\x} = \B\LRp{\x_n,\B\LRp{\x_{\n-1},\hdots,\B\LRp{\x_1,1}}}
$,
then $\Bc\LRp{\x} \in \L^1\LRp{\Rn}$.
\end{enumerate}
\theolab{Mish}
\end{theorem}

\section{A general framework for nAI}
\seclab{generalAI}
Inspired by the nAI proof of the Mish activation function in Section \secref{Mish}, we develop a general framework for nAI that requires only two conditions on the $\k$th derivative of any activation $\sigma\LRp{\xn}$. The beauty of the general framework is that it provides a single nAI proof that  is valid for a large class of functions including all activation functions that we have considered. The trade-off is that less can be said about the corresponding $\B$-functions. To begin, suppose that there exists $\k \in \mathbb{N}$ such that
\begin{enumerate}[label = \textbf{C\arabic*}]
\item \label{enum:AC} \underline{Integrability}: $\sigma^{(\k)}\LRp{\xn}$ is integrable, i.e., $\sigma^{(\k)}\LRp{\xn} \in \L^1\LRp{\R}$, and
\item \label{enum:UB} \underline{Essential boundedness}: there exists $M < \infty$ such that $\nor{\sigma^{(\k)}}_\infty \le M$.
\end{enumerate}
Note that 
if the two conditions \ref{enum:AC} and \ref{enum:UB} hold for $\k =0$ (e.g. Gaussian activation functions) then the activation is obviously an nAI. Thus we only consider $\k \in \mathbb{N}$, and the $\B$-function for $\sigma\LRp{\xn}$ can be defined via the $\k$-order central finite difference:
\begin{equation*}
\B\LRp{\xn,\h} := \deltah^\k\LRs{\sigma}\LRp{\xn} = \sum_{i=0}^\k\LRp{-1}^i
\begin{pmatrix}
\k \\ i
\end{pmatrix}
\sigma\LRp{\xn+\LRp{\frac{\k}{2}-i}\h}.
\end{equation*}

\begin{lemma}
For any $\h \in \R$, there holds:
\begin{multline*}
    \B\LRp{\xn,\h} = \frac{\h^\k}{\LRp{\k-1}!}
\sum_{i=0}^\k\LRp{-1}^i
\begin{pmatrix}
\k \\ i
\end{pmatrix}
\LRp{\frac{\k}{2}-i}^\k 
\int_0^1\sigma^{(\k)}\LRp{\xn + s\LRp{\frac{\k}{2}-i}\h}\LRp{1-s}^{\k-1}\,ds.
\end{multline*}
\lemlab{BellGeneral}
\end{lemma}
\begin{proof}
Applying the Taylor theorem gives
\begin{multline*}
\sigma\LRp{\xn + \LRp{\frac{\k}{2}-i}\h} = 
\sum_{j=0}^{\k-1}\sigma^{(j)}\LRp{\xn}\LRp{\frac{\k}{2}-i}^j\frac{\h^j}{j!} + 
\frac{\h^\k}{\LRp{\k-1}!}\\
\times \LRp{\frac{\k}{2}-i}^\k\int_0^1\sigma^{(\k)}\LRp{\xn + s\LRp{\frac{\k}{2}-i}\h}\LRp{1-s}^{\k-1}\,ds.    
\end{multline*}
The proof is concluded if we can show that 
\begin{multline*}
\sum_{i=0}^\k\LRp{-1}^i
\begin{pmatrix}
\k \\ i
\end{pmatrix}
\sum_{j=0}^{\k-1}\sigma^{(j)}\LRp{\xn}\LRp{\frac{\k}{2}-i}^j\frac{\h^j}{j!} = \sum_{j=0}^{\k-1}\sigma^{(j)}\LRp{\xn}\frac{\h^j}{j!}
\sum_{i=0}^\k\LRp{-1}^i
\begin{pmatrix}
\k \\ i
\end{pmatrix}
\LRp{\frac{\k}{2}-i}^j= 0,
\end{multline*}
but this is clear by the alternating sum identity
\[
\sum_{i=0}^\k\LRp{-1}^i
\begin{pmatrix}
\k \\ i
\end{pmatrix}
i^j = 0, \quad \text{ for } j=0,\hdots,\k-1.
\]
\end{proof}

\begin{theorem}
Let $\h \in \R$. Then: i) There exists $N < \infty$ such that $\snor{\B\LRp{\xn,\h}} \le N\snor{\h}^\k$; ii) There exists $C < \infty$ such that $\intR\snor{\B\LRp{\xn,\h}}\,d\xn \le C\snor{\h}^\k$; and iii) Let $\x =\LRp{\x_1,\hdots,\x_n} \in \Rn$. Define
$
\Bc\LRp{\x} = \B\LRp{\x_n,\B\LRp{\x_{\n-1},\hdots,\B\LRp{\x_1,1}}},
$
then $\Bc\LRp{\x} \in \L^1\LRp{\Rn}$.
\theolab{generalLCAI}
\end{theorem}
\begin{proof}
The first assertion is straightforward by invoking assumption \ref{enum:UB}, Lemma \lemref{BellGeneral}, and defining
$
N = \frac{M}{\k!}\sum_{i=0}^\k
\begin{pmatrix}
\k \\ i
\end{pmatrix}
\snor{\frac{\k}{2}-i}^\k.
$
For the second assertion, using triangle inequalities and the Fubini theorem yields
\begin{multline*}
    \intR\snor{\B\LRp{\xn,\h}}\,d\xn \le  \frac{\snor{\h}^\k}{\LRp{\k-1}!}\sum_{i=0}^\k
\begin{pmatrix}
\k \\ i
\end{pmatrix}
\snor{\frac{\k}{2}-i}^\k \\ \times \int_0^1\LRp{1-s}^{\k-1}
\intR\snor{\sigma^{(\k)}\LRp{\xn + s\LRp{\frac{\k}{2}-i}\h}}d\xn\,ds
= \frac{N}{M} \nor{\sigma^{(\k)}}_{\L^1\LRp{\R}}\snor{\h}^\k,
\end{multline*}
and, by defining $C = \frac{N}{M} \nor{\sigma^{(\k)}}_{\L^1\LRp{\R}}$, the result follows owing to assumption \ref{enum:AC}. The proof of the last assertion is the same as the proof of Theorem \theoref{GELUnD}. In particular, we have
\[
\int_{\R^\n}\snor{\Bc\LRp{\x}}\,d\x \le C^\k M^{\frac{\n^\k-1}{\n-1}-\k} < \infty.
\]
\end{proof}
\begin{remark}
Note that Theorem \theoref{generalLCAI} is valid for all activation functions considered in Section \secref{manyActivations} with appropriate $\k$: for example, $\k = \q + 1$  for  RePU of order $\q$, $\k = 1$ for generalized sigmoidal functions, $\k = 2$ for ELU, GELU, SiLU and Mish, $\k = 0$ for Gaussian, and etc. 
\end{remark}

\begin{remark}
Suppose a function $\sigma\LRp{\x}$ satisfies both conditions \ref{enum:AC} and \ref{enum:UB}. Theorem \theoref{generalLCAI} implies that $\sigma\LRp{\x}$ and all of its $j$th derivative, $j = 1,\hdots,\k$, are not only a valid activation function but also universal. For example, any differentiable and non-decreasing sigmoidal function satisfies the conditions with $\k = 1$, and therefore its derivative and itself are universal. 
\remalab{sigmoid}
\end{remark}

\begin{remark}
In one dimension, if $\sigma$ is of bounded variation, i.e. its total variation $TV\LRp{\sigma}$ is finite, then $\B\LRp{\xn,\h}$ 
resides in $\L^1\LRp{\R}\cap\L^\infty\LRp{\R}$ by taking $\k = 1$. A simple proof of this fact can be found in \cite[Corollary 3]{SIEGEL2020313}. Thus, $\sigma$ is an nAI.
\end{remark}

\begin{remark}
We have used central finite differences for convenience, but Lemma \lemref{BellGeneral}, and hence Theorem \theoref{generalLCAI}, also holds for $\k$th-order forward and backward finite differences. The proofs are indeed almost identical.
\remalab{FD}
\end{remark}

\section{Universality 
with non-asymptotic rates}
\seclab{rates}
This section explores Lemma \lemref{abstractUniversal} and Lemma
\lemref{nAI} to study the convergence of the neural network
\eqnref{NNf}
to the ground truth
function $\f\LRp{\x} \in \C_c\LRs{-1,1}^\n$ as a function of the number of neurons. 
From \eqnref{error}, we need to estimate
the modulus of continuity of $\f\LRp{\x}$ (the first and the second
terms) and the decaying property of the tail (the third term) of the
approximate identity $\Beps$ (and hence the tail of the activation
$\sigma\LRp{\x}$). 
For
the former, we further assume that $\f\LRp{\x}$ is Lipschitz so that the first
and the second terms can be estimated as $\omega\LRp{\f,\delta} =
\mc{O}\LRp{\delta}$ and $\omega\LRp{\f,\snor{\P^\m}} =
\mc{O}\LRp{\snor{\P^\m}}$. To balance these two terms,
we need to pick the partition $\P^\m$ such that $\snor{\P^\m} \approx
\delta$. From \eqnref{RiemannSum} and Lemma \lemref{QuadratureError},
we conclude $\m = \mc{O}\LRp{\delta^{-1}}$ and the total number of
quadrature points scales as $\m = \mc{O}\LRp{\delta^{-n}}$. It follows
from \eqnref{NNf} that the total number of neurons is $\N =
\mc{O}\LRp{k\delta^{-n}}$. Conversely, for a given number of neurons
$\N$, the error from the first two terms scales as $\delta =
\mc{O}\LRp{\N^{-1/n}}$, which is quite pessimistic. This is due to the
tensor product quadrature rule. 
The result can be improved using Monte Carlo estimation
of integrals at the expense of deterministic estimates. Indeed, let $\xib^i$, $i = 1,\hdots,\N$, be independent and identically distributed (i.i.d.)
by the uniform distribution on $\LRs{-1,1}^\n$. A Monte Carlo counterpart of \eqnref{NNf} is given as
\begin{equation}
\tilde{\S}\LRp{\m,\f * \Beps}\LRp{\x} = \frac{1}{\N}
\sum_{j=1}^\N\f\LRp{\xib^{j}}
 \sum_{\ell=1}^\k\alpha_\ell\sigma\LRs{\frac{\wb^\ell}{\theta}\cdot\LRp{\x-{\xib^j}}
   + \b_\ell},
 \eqnlab{NNfmc}
\end{equation}
 which, by the law of large numbers, converges almost surely (a.s.) to $\f * \Beps\LRp{\x}$ for every $\x$. However, the Monte Carlo mean square error estimate
 \[
 \E_{\xib^1,\hdots,\xib^\N}\LRs{\snor{\tilde{\S}\LRp{\m,\f * \Beps}\LRp{\x} - \f * \Beps\LRp{\x}}^2} = \mc{O}\LRp{\N^{-1}}
 \]
 is not useful for our $\infty$-norm estimate without further technicalities in exchanging expectation and $\infty$-norm
 (see, e.g., \cite[Theorem 14.60]{RockafellarWetts98} and \cite{HafsaMandallena03}).
 
 While most, if not all, literature concerns deterministic and asymptotic rates, we pursue a non-asymptotic direction as it precisely captures the behavior of the Monte Carlo estimate \eqnref{NNfmc} of $\f * \Beps\LRp{\x}$, and hence of $\f\LRp{\x}$.
 Within the non-asymptotic setting, we now show that, with high probability, any neural network with the nAI property and $\k\N$ neurons converges with rate $\N^{-\half}$ to any Lipschitz continuous function $\f\LRp{\x}$ with compact support in $\LRs{-1,1}^\n$.
 
\begin{theorem}
 Let $\xib^i$, $i = 1,\hdots,\N$, be i.i.d. samples from
the uniform distribution on $\LRs{-1,1}^\n$. Suppose that $\f\LRp{\x}$ is Lipschitz and $\Beps\LRp{\x}$ is continuous. Furthermore, let $\delta = \mc{O}\LRp{\frac{C}{\sqrt{\N}}}$ and $\theta = \mc{O}\LRp{\delta^2}$ for some $C > 0$. There exist three absolute constants $\alpha$, $\ell$, and $L$:
\begin{equation}
  \nor{\tilde{\S}\LRp{\m,\f * \Beps}\LRp{\x} - \f\LRp{\x}}_\infty \le   \frac{C}{\sqrt{\N}} 
  \eqnlab{errorMC}
\end{equation}
holds with probability at least $1 - 2 e^{-2\frac{\LRp{C-\alpha\N}^2}{\LRp{L-\ell}^2}}$.
\theolab{universalRates}
\end{theorem}
\begin{proof}
If we define
\[
\g^j\LRp{\x} := \f\LRp{\xib^{j}}\Beps\LRp{\x - \xib^j}
\]
then $\g^j\LRp{\x}$, $j = 1,\hdots,\N$, are independent random variables and $\tilde{\S}\LRp{\m,\f * \Beps}\LRp{\x} = \frac{1}{\N}
\sum_{j=1}^\N\g^j\LRp{\x}$. Owing to the continuity of $\f$ and $\Beps$, we know that there exist two absolute constants $\ell$ and $L$ such that $\ell \le \g^j\LRp{\x} \le L$. Let $\h\LRp{\x} := \snor{\tilde{\S}\LRp{\m,\f * \Beps}\LRp{\x} -  \f * \Beps\LRp{\x}}$, then by Hoeffding inequality \cite{MohriRostamizadehTalwalkar18,Shalev-ShwartzBen-David2014} we have
\begin{equation}
\Prob\LRs{
\h\LRp{\x} > \varepsilon 
}  \le 2e^{-2\N\frac{\varepsilon^2}{\LRp{L-\ell}^2}}
\eqnlab{tailBound}
\end{equation}
for
each $\x$, where $\Prob\LRs{\cdot}$ stands for probability. An application of triangle inequality gives
\[
\snor{\h\LRp{\x} - \h\LRp{\y}} \le 2\LRp{L + \sup_{\x \in \LRs{-1,1}^\n}\f * \Beps\LRp{\x}} =: \alpha,
\]
where 
$\alpha$ is meaningful as $\f * \Beps\LRp{\x}$ is continuous. 
Thus, for a given $\y$, there holds
\[
\h\LRp{\y} \ge \h\LRp{\x} - \alpha \quad \forall \x \in \LRs{-1,1}^\n,
\]
that is,
\[
\h\LRp{\y} \ge \nor{\h\LRp{\x}}_\infty - \alpha.
\]
Consequently, using the tail bound \eqnref{tailBound} yields
\[
  \Prob\LRs{\nor{\h\LRp{\x}}_\infty > \frac{C}{\sqrt{\N}}} \le
  \Prob\LRs{\h\LRp{\y} > \frac{C}{\sqrt{\N}} - \alpha} \le 
2e^{-2\frac{\LRp{C -\alpha\N}^2}{\LRp{L-\ell}^2}}.
\]
It follows that
\begin{equation}
\nor{\tilde{\S}\LRp{\m,\f * \Beps}\LRp{\x} -  \f * \Beps\LRp{\x}}_\infty \le  \frac{C}{\sqrt{\N}}
\eqnlab{hoeffding}
\end{equation}
with probability at least $1 - 2 e^{-2\frac{\LRp{C-\alpha\N}^2}{\LRp{L-\ell}^2}}$ for any $C$. Clearly, we need to pick either $C\ll \alpha\N$ or $C \gg \alpha\N$. The former is more favorable as it makes the error in \eqnref{hoeffding} small with high probability. The latter could lead to large error with high probability.
Now, choosing $\delta = \mc{O}\LRp{\frac{C}{\sqrt{\N}}}$ and following the proof of Lemma \lemref{abstractUniversal} we arrive at
\[
\nor{\tilde{\S}\LRp{\m,\f * \Beps}\LRp{\x} - \f\LRp{\x}}_\infty =  \mc{O}\LRp{\frac{C}{\sqrt{\N}}
  + \T\LRp{\Beps,\frac{C}{\sqrt{\N}}}}.
\]
We now estimate 
$\T\LRp{\Beps,\frac{C}{\sqrt{\N}}} =  \int_{\nor{\y} > \frac{C}{\sqrt{\N}}} \snor{\Beps\LRp{\y}}\,d\y$.
By Markov inequality we have
\[
\T\LRp{\Beps,\frac{C}{\sqrt{\N}}} \le \frac{\theta\sqrt{\N}}{C},
\]
where we have used the fact that $\nor{\B\LRp{\x}}_{\L^1\LRp{\R^\n}} = 1$. Now taking $\theta = \mc{O}\LRp{\delta^2} = \mc{O}\LRp{\frac{C^2}{\N}}$ yields
\[
\T\LRp{\Beps,\frac{C}{\sqrt{\N}}} = \mc{O}\LRp{\frac{C}{\sqrt{\N}}},
\]
and this concludes the proof.
\end{proof}
\begin{remark}
The continuity of $\Beps\LRp{\x}$ in Theorem \theoref{universalRates} is only sufficient. 
All we need is the boundedness of $\Beps\LRp{\x}$. Theorem \theoref{universalRates} is thus valid for all activation functions that we have discussed, including those in Section \secref{generalAI}.
{Theorem \theoref{universalRates} provides not only theoretical insights into the required number of neurons but also a guide on how to choose the number of neurons to achieve a certain accuracy with controllable successful probability. Perhaps more importantly, it shows that neural networks may fail, with non-zero probability, to provide a desired testing error in practice when an architecture, and hence a number of neurons, is selected. }
\end{remark}

\section{Conclusions}
\seclab{conclusion}
We have presented a constructive framework for neural network universality. At the heart of the framework is the neural network approximate identity (nAI) concept that allows us to unify most of activations under the same umbrella. Indeed, we have shown that  most of existing activations are nAI, and thus universal in the space of continuous of functions on compacta. We have shown that for an activation to be nAI, it is sufficient to verify that its $\k$th derivative, $\k \ge 0$, is essentially bounded and integrable.
The framework induces several advantages over contemporary approaches. First, our approach is constructive with elementary means from functional analysis, probability theory, and numerical analysis. Second, it is the first attempt to unify the universality for most of existing activation functions. Third, as a by product, the framework provides the first universality proof for some of the existing activation functions including the Mish, SiLU, ELU, GELU, etc. Fourth, it provides a new universality proof for most activation functions. Fifth,
it discovers new activation functions with guaranteed universality property. Sixth, for each activation, the framework provides precisely the architecture of the one-hidden neural network with predetermined number of of neurons, and the values of weights/biases.  Seventh, the framework facilitates us to develop the first abstract universal result with favorable non-asymptotic rates of $\N^{-\half}$, where $\N$ is the number of neurons. Our framework also provides insights into the derivations of some of the existing approaches. {Ongoing work is to build upon our framework to study the universal approximation properties of convolutional neural networks and deep neural networks. Part of the future work is also to exploit the unified nature of the framework to study which activation is better, in which sense, for a wide range of classification and regression tasks.}


\section*{Funding}
This work is partially supported by National Science Foundation awards NSF-OAC-2212442, NSF-2108320, NSF-1808576, and NSF-CAREER-1845799; by Department of Energy awards DE-SC0018147 and DE-SC0022211; and by a 2021 UT-Portugal CoLab award, and we are grateful for the support. This paper describes objective technical results and analysis. Any subjective views or opinions that might be expressed in the paper do not necessarily represent the views of the U.S. National Science Foundation or the U.S. Department of Energy or the United States Government. We would like to thank Professor Christoph Schwab and Professor Ian Sloan for pointing out a technical mistake in the preprint.

\appendix

\section{Extension to $\C\LRs{-1,1}^\n$}
\label{extension}

In this section, we present an approach to extend the results in Sections \secref{errorAI} and \secref{rates} to continuous functions. To directly build upon the results in these sections to our extension, without loss of generality,  we considers $\C\LRs{-b,b}^\n$
such that $0 < \sqrt{2}b <1$.
Consider $\g \in \C\LRs{-b,b}^\n$
and let $\hat{\g} \in \C\LRp{\Rn}$ be an extension of $\eval{\hat{\g}}_{{\LRs{-b,b}^\n}} = \g$, and $\nor{\hat{\g}}_\infty = \nor{\g}_\infty$ \cite{Milman1997}. Next, let us take $\varphi \in \C_b\LRs{-1,1}^n$ such that: i) $\eval{\varphi}_{\LRs{-b,b}^\n} = 1$ and  ii) $\omega\LRp{\varphi,\h} = \mc{O}\LRp{\omega\LRp{\g,\h}}$ for any $\h > 0$. Define $\f := \hat{\g}\varphi$, then it is easy to see that:
\begin{itemize}
    \item $\eval{\f}_{\LRs{-b,b}^\n} = g$,
    \item $\f \in \C_c\LRs{-1,1}^\n$,
    \item $\nor{\f}_\infty = \mc{O}\LRp{\nor{\g}_\infty}$, and
    \item $\omega\LRp{\f,\h} = \mc{O}\LRp{\omega\LRp{\g,\h}}$.
\end{itemize}
Now, applying Lemma \lemref{abstractUniversal} we obtain
\begin{multline*}
     \f\LRp{\x} - \S\LRp{\m,\f * \Beps}\LRp{\x} = \mc{O}\LRp{\omega\LRp{\f,\delta} + \omega\LRp{\f,\snor{\P^\m}} + \T\LRp{\Beps,\delta} } \\ = 
     \mc{O}\LRp{\omega\LRp{\g,\delta} + \omega\LRp{\g,\snor{\P^\m}} + \T\LRp{\Beps,\delta} } 
\end{multline*}
for all $\x \in \Rn$. Thus, by restricting $\x \in \LRs{-b,b}^\n$ we arrive at
\[
 \g\LRp{\x} - \S\LRp{\m,\f * \Beps}\LRp{\x} =  \mc{O}\LRp{\omega\LRp{\g,\delta} + \omega\LRp{\g,\snor{\P^\m}} + \T\LRp{\Beps,\delta} }.
\]
This, together with Lemma \lemref{nAI}, ensures the universality of any nAI activation in $\C\LRs{-b,b}^\n$. The extension for Lipschitz continuous functions on $\LRs{-b,b}^\n$ for Section \secref{rates} follows similarly, again using the key extension results in \cite{Milman1997}.


\section{Proofs of results in Section 7}
\seclab{proofs}
This section  presents the detailed proofs of results in Section \secref{manyActivations}.

\begin{proof}[Proof of Lemma \lemref{RePU}]
We start by defining $\X_\r := \sum_{i=1}^\r\U_i$, where $\U_i$ are independent identically distributed uniform random variables on $\LRs{-\frac{1}{2},\frac{1}{2}}$. Following \cite{feller1,Irwin27,Hall27}, $\X_r$ is distributed by the Irwin-Hall distribution with  the probability density function
\[
\f_\r\LRp{\xn} := \frac{1}{\q!}\sum_{i=1}^{\floor{\xn}}(-1)^i
\begin{pmatrix}
\r \\ i
\end{pmatrix}
\LRp{\xn + \frac{\r}{2}- i}^{\q},
\]
where $\floor{\xn}$ denotes the largest integer smaller than $\xn$. Using the definition of \RePU, it is easy to see that $\f_r\LRp{\xn}$ can be also written in terms of \RePU\, as follows
\[
\f_\r\LRp{\xn} := \frac{1}{\q!}\sum_{i=1}^{\r}(-1)^i
\begin{pmatrix}
\r \\ i
\end{pmatrix}
\RePU\LRp{\q; \xn + \frac{\r}{2}- i},
\]
which in turns implies
\begin{equation}
\B\LRp{\xn,\h} = \h^\q\f_\r\LRp{\frac{\xn}{\h}}.
\eqnlab{Bandf}
\end{equation}
In other words, $\B\LRp{\xn,\h}$ is a dilated version of $\f_\r\LRp{\xn}$. Thus, all the properties of $\f_\r\LRp{\xn}$ holds for $\B\LRp{\xn,\h}$. In particular, all the assertions of Lemma \lemref{RePU} hold.
Note that the compact support can be alternatively shown using the property of the central finite differencing. Indeed, it is easy to see that
for $\xn \ge \frac{\r\h}{2}$ we have
\[
\deltah^\r\LRs{\RePU}\LRp{\xn} = \deltah^\r\LRs{\xn^\q} = 0.
\]
\end{proof}

\begin{proof}[Proof of Theorem \theoref{RePUnD}]
The first three assertions are direct consequences of Lemma \lemref{RePU}. For the fourth assertion, since $\Bc\LRp{\x} \ge 0$ it is sufficient to show $\intR \Bc\LRp{\x}\,d\x \le 1$ 
and we do so in three steps. Let $\r = \q+1$ and define $\x = \LRp{\x_{\n+1},\x_\n,\hdots,\x_1} = \LRp{\x_{\n+1},\y}$. We first show by induction that $\B\LRp{\xn,1} \le 1$ for $\q \in \mathbb{N}$ and $\xn \in \R$. The claim is clearly true for $\q = \LRc{0,1}$. Suppose the claim holds for $\q$, then \eqnref{Bandf} implies
\[
\f_{\r}\LRp{\xn} \le 1, \quad \forall \xn \in \R.
\]
For $\q+1$ we have
\[
\B\LRp{0} = \f_{\r+1}\LRp{0} = \f_{\r} * \f_1\LRp{0} = \int_\R \f_{\r}\LRp{-\yn}\f_1\LRp{\yn}\,d\yn \le \int_\R\f_1\LRp{\yn}\,d\yn = 1.
\]
By the second assertion, we conclude that  $\B\LRp{\xn,1} \le 1$ for any $\q \in \mathbb{N}$ and $\xn \in \R$. 

In the second step, we show $\Bc\LRp{\x} \le 1$ by induction on $\n$ for any $\x \in \Rn$ and any $\q \in \mathbb{N}$. The result holds for $n=1$ due to the first step. Suppose the claim is true for $\n$. For $\n+1$, we have
\[
    \Bc\LRp{\x} = \B\LRp{\x_{\n+1},\Bc\LRp{\y}} = \LRs{\Bc\LRp{\y}}^\q \f_\r
    \LRp{\frac{\x_{\n+1}}{\Bc\LRp{\y}}} \le 1,
\]
where we have used \eqnref{Bandf} and in the last equality, and the first step together with the induction hypothesis in the last inequality.

In the last step, we show $\intR \Bc\LRp{\x}\,d\x \le 1$ by induction on $\n$. For $\n = 1$, $\int_\R \Bc\LRp{\xn}\,d\xn = 1$ is clear using by the Irwin-Hall probability density function. Suppose the result is true for $\n$. For $\n+1$, applying the Fubini theorem gives
\begin{multline*}
 \int_{\R^{\n+1}} \Bc\LRp{\x}\,d\x = \intR\LRp{\int_{\R} \B\LRp{\x_{\n+1},\Bc\LRp{\y}}\,d\x_{\n+1}}\,d\y \\= 
 \intR\LRs{\Bc\LRp{\y}}^\q\LRp{\int_{\R} \f_{\r}\LRp{\frac{\x_{\n+1}}{\Bc\LRp{\y}}}\,d\x_{\n+1}}\,d\y = 
 \intR\LRs{\Bc\LRp{\y}}^\r\LRp{\int_{\R} \f_{\r}\LRp{t}\,dt}\,d\y \\
 =  \intR\LRs{\Bc\LRp{\y}}^\r\,d\y \le \intR\Bc\LRp{\y}\,d\y \le 1,
\end{multline*}
where we have used \eqnref{Bandf} in the second equality, the result of the second step in the second last inequality, and the induction hypothesis in the last inequality.
\end{proof}

\begin{proof}[Proof of Lemma \lemref{sigmoid1D}]
The proof for $\B_s\LRp{\xn,\h}$ is a simple extension of those in \cite{costarelli13,CHEN2009}, and the proof for $\B_s\LRp{\xn,\h}$ follows similarly. Note that $\int_\R\B\LRp{\xn,\h}\,d\xn = \h$ for sigmoid and hyperbolic tangent.
For  $\B_p\LRp{\xn,\h}$, due to the global convexity of $\sigma_t\LRp{\xn}$ we have
\[
\ln\LRp{1+e^{\xn}} \le \frac{\ln\LRp{e^\h+e^{\xn}} + \ln\LRp{e^{-\h}+e^{\xn}}}{2}, \quad \forall \xn \in \R,
\]
which is equivalently to $\B_p\LRp{\xn,\h}\ge 0$ for $\xn \in \R$. The fact that $\B_p\LRp{\xn,\h}$ is even and $\lim_{\snor{\xn} \to \infty}\B_p\LRp{\xn,\h}$ = 0 are obvious by inspection. 
Since the derivative of $\B_p\LRp{\xn,\h}$ is negative for $\xn \in (0,\infty)$ and $\B_p\LRp{\xn,\h}$ is even, $\B_p\LRp{\xn,\h}$ is unimodal. 
It follows that $\B_p\LRp{\xn,0} = \max_{\xn \in R}\B_p\LRp{\xn,\h}$. 

Next integrating by parts gives
\begin{multline}
    \int_{-\infty}^\infty \B_p\LRp{\xn,\h}\,d\xn = 
    2\int_{0}^\infty\frac{\LRp{e^\h-1}^2\LRp{1-e^{-\xn}}}{\LRp{e^\h+e^{-x}}\LRp{e^{-\h}+e^{-x}}\LRp{1+e^{-x}}}e^{-\xn}\xn\,d\xn \\
    \le \LRp{\frac{1 - e^{-\h}}{1+e^{-\h}}}^2\int_{0}^\infty e^{-\xn}\xn\,d\xn = \LRp{\frac{1 - e^{-\h}}{1+e^{-\h}}}^2 \le \min\LRc{1,\h^2}.
    \eqnlab{softplus}
\end{multline}
Thus all the assertions for $\B_p\LRp{\xn,\h}$ holds.
\end{proof}

\begin{proof}[Proof of Theorem \theoref{sigmoidnD}]
We only need to show $\intR \Bc\LRp{\x}\,d\x \le 1$ as the proof for other assertions is similar to that of Theorem \theoref{RePUnD}, and thus is omitted. For sigmoid and hyperbolic tangent the result is clear as
\begin{multline*}
\int_{\R^{\n}} \Bc\LRp{\x}\,d\x = \int_{\R^{\n}} \B\LRp{\x_{\n},\B\LRp{\x_{\n-1},\hdots,\B\LRp{\x_1,1}}}\,d\x\\ = \int_{\R^{\n-1}} \B\LRp{\x_{\n-1},\hdots,\B\LRp{\x_1,1}}\,d\x_{\n-1}\hdots d\x_1 = \hdots =
\int_{\R} \B\LRp{\x_1,1}d\x_1 = 1.    
\end{multline*}
For softplus function, by inspection, $\B\LRp{\z,\h} \le \B\LRp{0,\h} \le 1$ for all $\z \in \R$ and $0 < \h\le 1$.  Lemma \lemref{sigmoid1D} gives $\intR \Bc\LRp{\x}\,d\x \le 1$ for $\n = 1$. 
Define $\x = \LRp{\x_{\n+1},\x_\n,\hdots,\x_1} = \LRp{\x_{\n+1},\y}$ and suppose the claim holds for $\n$, i.e.,  $\intR \Bc\LRp{\y}\,d\y \le 1$. Now apply \eqnref{softplus} we have
 \begin{multline*}
     \int_{\R^{\n+1}} \Bc\LRp{\x}\,d\x = \intR\LRp{\int_{\R} \B\LRp{\x_{\n+1},\Bc\LRp{\y}}\,d\x_{\n+1}}\,d\y \le
     \intR\Bc\LRp{\y}\,d\y 
     \le 
     1, 
 \end{multline*}
 which ends the proof by induction. 
\end{proof}

\begin{proof}[Proof of Lemma \lemref{generalizedSigmoid1D}]
The first assertion is clear by Definition \defiref{sigmoid}. For the second assertion, with a telescope trick similar to \cite{costarelli13} we have
\begin{multline*}
s_N\LRp{\xn} = \sum_{k = -N}^N \B\LRp{\xn+k\h,\h} = \frac{1}{2\LRp{L - \ell}} \left[\sigma\LRp{\xn + \LRp{N+1}\h} + \sigma\LRp{\xn + N\h}\right. \\
- \left.\sigma\LRp{\xn -N\h} - \sigma\LRp{\xn - \LRp{N+1}\h}\right] \xrightarrow[]{N \to \infty} 
\text{sign}\LRp{\h}.
\end{multline*}
To show the convergence is uniform, we consider only $\h > 0$ as the proof for  $\h < 0$ is similar and for $\h = 0$ is obvious. We first consider 
the right tail. For sufficiently large $N$, there exists a constant $C > 0$ such that
\begin{multline*}
    \sum_{k = N}^\infty \B\LRp{\xn+k\h,\h} \le 
    C\sum_{k = N}^\infty \LRp{\xn+k\h}^{-1-\alpha} = 
    C\sum_{k = N}^\infty \int_{k-1}^k\LRp{\xn+k\h}^{-1-\alpha}\,dy
    \\ 
    \le C \int_{N-1}^\infty \LRp{\xn+y\h}^{-1-\alpha}\,dy = \frac{C}{\h\alpha}\LRs{\xn + \LRp{N-1}\h} \\
    \le  \frac{C}{\h\alpha}\LRs{m + \LRp{N-1}\h} \xrightarrow[]{N \to \infty} 0 \text{ independent of } \xn.
\end{multline*}
Similarly, the left tail converges to 0 uniformly as
\[
\sum_{k = N}^\infty \B\LRp{\xn-k\h,\h} \le \frac{C}{\h\alpha}\LRs{\LRp{N-1}\h - M} \xrightarrow[]{N \to \infty} 0 \text{ independent of } \xn.
\]
As a consequence, we have
\begin{multline*}
    \int_{\R} \B\LRp{\xn,\h}\,d\xn = \sum_{k=-\infty}^\infty \int_{k\snor{\h}}^{\LRp{k+1}\snor{\h}}\B\LRp{\xn,\h}\,d\xn = 
    \sum_{k=-\infty}^\infty \int_{0}^{\snor{\h}}\B\LRp{y + k\snor{\h},\h}\,dy
    \\
    =  \int_{0}^{\snor{\h}} \sum_{k=-\infty}^\infty \B\LRp{y + k\snor{\h},\h}\,dy = \h,
\end{multline*}
where we have used the uniform convergence in the third equality.

Using the first assertion we have
\begin{multline*}
    \int_{\R} \snor{\B\LRp{\xn,\h}}\,d\xn = \int_{\xn^- - \snor{\h}}^{\xn^+ + \snor{\h}} \snor{\B\LRp{\xn,\h}}\,d\xn + \int_{\xn^+ + \snor{\h}}^\infty \snor{\B\LRp{\xn,\h}}\,d\xn + \int_{-\infty}^{\xn^- - \snor{\h}} \snor{\B\LRp{\xn,\h}}\,d\xn \\
    \le \LRp{\xn_p - \xn_n + 2\snor{\h}} + C\int_{\xn^+ + \snor{\h}}^\infty \LRp{\xn + \snor{\h}}^{-1-\alpha}\,d\xn + C\int_{-\infty}^{\xn^- - \snor{\h}} \LRp{-\xn + \snor{\h}}^{-1-\alpha} \,d\xn  \\
    = \LRp{\xn_p - \xn_n + 2\snor{\h}} + \frac{C}{\alpha}\LRs{\LRp{\xn^+ + 2\snor{\h}}^{-\alpha}+\LRp{-\xn^- + 2\snor{\h}}^{-\alpha}} <\infty.
\end{multline*}
\end{proof}

\begin{proof}[Proof of Theorem \theoref{generalizedSigmoidnD}]
The proof is the same as the proof of Theorem \theoref{sigmoidnD} for the standard sigmoidal unit, and thus is omitted. 
\end{proof}

\begin{proof}[Proof of Lemma \lemref{ELU1D}]
The expression of $\B\LRp{\xn,\h}$ and direct integration give  $\int_{\R} \B\LRp{\xn,\h}\,d\xn = \h^2\gamma$. Similarly, simple algebra manipulations yield
\[
\gamma\int_{\R} \snor{\B\LRp{\xn,\h}}\,d\xn \le
\h^2 + 2\snor{\alpha}\snor{\h} + 2\snor{\alpha}\LRp{e^{-\snor{\h}}-1}^2 
< \infty,    
\]
and thus $\B\LRp{\xn,\h} \in \L^1\LRp{\R}$. The fact that $\B\LRp{\xn,\h} \le 1$ for $\snor{\h} \le 1$ holds is straightforward by inspecting the extrema of $\B\LRp{\xn,\h}$.
\end{proof}

\begin{proof}[Proof of Theorem \theoref{ELUnD}]
From the proof of Lemma \lemref{ELU1D} we infer that  $\int_{\R} \snor{\B\LRp{\xn,\h}}\,d\xn \le \snor{\h}$ for all $\snor{\h} \le 1$: in particular, $\int_{\R} \snor{\B\LRp{\x_1,1}}\,d\xn \le 1$. Define $\x = \LRp{\x_{\n+1},\x_\n,\hdots,\x_1} = \LRp{\x_{\n+1},\y}$ and suppose the claim holds for $\n$, i.e.,  $\intR \snor{\Bc\LRp{\y}}\,d\y \le 1$. We have
\[
 \int_{\R^{\n+1}} \snor{\Bc\LRp{\x}}\,d\x = \intR\LRp{\int_{\R} \snor{\B\LRp{\x_{\n+1},\Bc\LRp{\y}}}\,d\x_{\n+1}}\,d\y \le
     \intR\snor{\Bc\LRp{\y}}\,d\y 
     \le 
     1, 
\]
which, by induction, concludes the proof.
\end{proof}

\begin{proof}[Proof of Lemma \lemref{GELU1D}]
The first assertion is straightforward. For the second assertion, it is sufficient to  consider $\xn \ge 0$ and $\h > 0$. Any root of $\B\LRp{\xn,\h}$ satisfy the following equation
\[
f\LRp{\xn} := \frac{\Phi\LRp{\xn+h} - \Phi\LRp{\xn}}{\Phi\LRp{\xn} - \Phi\LRp{\xn-h}} = \frac{\xn - \h}{\xn+\h} =: g\LRp{\xn}.
\]
Since, given $\h$, $f\LRp{\xn}$ is a positive decreasing function and $g\LRp{\xn}$ is an increasing function (starting from $-1$), they can have at most one intersection. Now for sufficiently large $\xn$, using the the mean value theorem it can be shown that
\[
f\LRp{\xn} \approx e^{-\xn\h} \xrightarrow[]{\xn \to \infty} 0, \text{ and } g\LRp{\xn} \xrightarrow[]{\xn \to \infty} 1,
\]
from which we deduce that that there is only one intersection, and hence only one positive root $\xn^*$ for $\B\LRp{\xn,\h}$. 
Next, we notice $\g\LRp{\xn} \le 0 < f\LRp{\xn}$ for $0\le \xn \le h$. For $\h \ge 1$ it is easy to see $\f\LRp{2\h} < g\LRp{2\h} = 1/3$. Thus, $\h < \xn^* < 2h$ for $h\ge 1$. For $0 < \h < 1$, by inspection we have $f\LRp{2} < g\LRp{2}$, and hence $ \h < \xn^* < 2$. In conclusion $h < \xn^* < \max\LRc{2,2\h}$. 

The preceding argument
also shows that $\B\LRp{\xn,\h} \ge 0$ for $0 \le \xn \le \xn^*$, $\B\LRp{\xn,\h} < 0$ for $\xn > \xn^*$, and $\B\LRp{\xn,\h} \xrightarrow[]{\xn \to \infty} 0^-$. 
Using the Taylor formula we have 
\[
\snor{\B\LRp{\bar\xn,\h}} \le \frac{1}{\sqrt{2\pi}}\h^2.
\]

For the third assertion, it is easy to verify the following indefinite integral (ignoring the integration constant as it will be canceled out)
\[
\int \xn \Phi\LRp{\xn}\,d\xn = \half\LRp{\xn^2-1}\Phi\LRp{\xn} + \frac{xe^{{-\frac{\xn^2}{2}}}}{2\sqrt{2\pi}},
\]
which, together with simple calculations, yields
\[
    \int_{\R} \B\LRp{\xn,\h}\,d\xn = \h^2. 
\]
From the proof of the second assertion and the Taylor formula we can show
\begin{multline*}
     \int_{\R} \snor{\B\LRp{\xn,\h}}\,d\xn = 2\int_{0}^{\xn^*} \B\LRp{\xn,\h}\,d\xn - 
     2\int_{\xn^*}^{\infty} \B\LRp{\xn,\h}\,d\xn \\=
     -3\h^2 + 2\deltah^2\LRs{\LRp{\LRp{\xn^*}^2-1}\Phi(\xn^*)}
    + \deltah^2\LRs{\frac{\xn^*e^{{-\frac{\LRp{\xn^*}^2}{2}}}}{\sqrt{2\pi}}} \le \frac{37}{10}\h^2.
\end{multline*}
Thus, $ \B\LRp{\xn,\h} \in \L^1\LRp{\R}$.
\end{proof}

\begin{proof}[Proof of Theorem \theoref{GELUnD}]
From the proof of Lemma \lemref{GELU1D} we have   $\int_{\R} \snor{\B\LRp{\xn,\h}}\,d\xn \le C\h^2$ and $\snor{\B\LRp{\xn,\h}} \le M\h^2$ for all ${\h} \in \R$ where $M = \frac{2}{\sqrt{2\pi}}$, $C = \frac{37}{10}$: in particular, $\int_{\R} \snor{\B\LRp{\x_1,1}}\,d\xn \le C$. It is easy to see by induction that
\[
\snor{\Bc\LRp{\x}} \le M^{2^\n-1}.
\]
We claim that
\[
\int_{\R^{\n}} \snor{\Bc\LRp{\x}}\,d\x \le C^\n M^{2^\n-
\n-1},
\]
and thus $\Bc\LRp{\x} \in \L^1\LRp{\Rn}$. We prove the claim by induction.
Clearly it holds for $\n = 1$. Suppose it is valid for $\n$.  Define $\x = \LRp{\x_{\n+1},\x_\n,\hdots,\x_1} = \LRp{\x_{\n+1},\y}$, we have
 \begin{multline*}
     \int_{\R^{\n+1}} \snor{\Bc\LRp{\x}}\,d\x = \intR\LRp{\int_{\R} \snor{\B\LRp{\x_{\n+1},\Bc\LRp{\y}}}\,d\x_{\n+1}}\,d\y \le
     C\intR\snor{\Bc\LRp{\y}}^2\,d\y \\
     \le C M^{2^\n-1}\intR\snor{\Bc\LRp{\y}}d\y \le C^{\n+1}M^{2^{\n+1}-n-2}, 
 \end{multline*}
 where we have used the induction hypothesis in the last inequality, and this ends the proof.
\end{proof}

\begin{proof}[Proof of Lemma \lemref{Mish2ndDiff}]
It is easy to see that for $\xn \ge 2$
\[
\snor{\sigma^{(2)}\LRp{\xn}} \le 12e^{-3\xn}\LRp{\xn-2} +8e^{-2\xn}\LRp{\xn-1} + 8e^{-5\xn}\LRp{\xn+2} + 8e^{-4\xn}\LRp{\xn+4},
\]
and 
\[
\snor{\sigma^{(2)}\LRp{\xn}} \le 12e^{3\xn}\LRp{2-\xn} +8e^{4\xn}\LRp{1-\xn} - 8e^{\xn}\LRp{\xn+2} - 8e^{2\xn}\LRp{\xn+4},
\]
for $\xn \le -4$. That is, both the right and the left tails of $\sigma^{(2)}\LRp{\xn}$ decay exponentially, and this concludes the proof.
\end{proof}

\begin{proof}[Proof of Theorem \theoref{Mish}]
For the first assertion, we note that  $\sigma^{(2)}\LRp{\xn}$ is continuous and thus the following Taylor theorem with integral remainder for $\sigma\LRp{\xn}$ holds
\[
\sigma\LRp{\xn+\h} = \sigma\LRp{\xn} + \sigma'\LRp{\xn}\h +\h^2\int_{0}^1\sigma^{(2)}\LRp{\xn + s\h}\LRp{1-s}\,ds.
\]
As a result,
\begin{multline*}
\snor{\B\LRp{\xn,\h}} \le \h^2\LRs{\int_{0}^1\snor{\sigma^{(2)}\LRp{\xn + s\h}}\LRp{1-s}\,ds + \int_{0}^1\snor{\sigma^{(2)}\LRp{\xn - s\h}}\LRp{1-s}\,ds} \le M\h^2,   
\end{multline*}
where we have used the boundedness of $\sigma^{(2)}\LRp{\xn}$ from Lemma \lemref{Mish2ndDiff} in the last inequality.

For the second assertion, we have
\begin{multline*}
    \intR\B\LRp{\xn,\h}\,d\xn = \h^2 \int_\R\int_{0}^1{\sigma^{(2)}\LRp{\xn +s\h}}\LRp{1-s}\,dsd\xn \\+ \h^2\int_\R\int_{0}^1{\sigma^{(2)}\LRp{\xn - s\h}}\LRp{1-s}\,dsd\xn,
\end{multline*}
whose right hand side is well-defined owing to $\sigma^{(2)}\LRp{\xn} \in \L^1\LRp{\R}$ (see Lemma \lemref{Mish2ndDiff}) and the Fubini theorem. In particular,
\[
    \intR\B\LRp{\xn,\h}\,d\xn \le \nor{\sigma^{(2)}}_{\L^1\LRp{\R}}\h^2.
\]
The proof for $\intR\snor{\B\LRp{\xn,\h}}\,d\xn \le \nor{\sigma^{(2)}}_{\L^1\LRp{\R}}\h^2$ follows similarly.

The proof of the last assertion is the same as the proof of Theorem \theoref{GELUnD} and hence is omitted.
\end{proof}

\section{Figures}
\seclab{figures}
This section provides the plots of the $\B$ functions for various activations in 1, 2, and 3 dimensions.

Figure \figref{Bfunc2D} (left) plots $\B\LRp{\xn,1}$ for RePU unit with $\q =
\LRc{0,1,3, 6,9}$ in one dimension. The non-negativeness, compact-support,
 $\B$ell shape, smoothness, and unimodal of $\B\LRp{\xn,1}$ can be
clearly seen. For $\n = 2$ we plot in Figure \figref{Bfunc2D} (the three right most subfigures) the
surfaces of RePU for $\q = \LRc{0,1,5}$ together with $15$ contours to
again verify the non-negativeness, compact-support, $\B$ell
shape, smoothness, and unimodal of $\Bc\LRp{\x} =
\B\LRp{\xn_2,\B\LRp{\xn_1,1}}$. To further confirm these features for
$\n = 3$, in figure \figref{Bfunc3D} we plot an isosurface for RePU
unit with $\q = \LRc{0,1,3,5}$, and $4$ isosurfaces for the case $\q =
4$ in Figure \figref{Bfunc3Dq4}. Note the supports of the $\B$ functions around the origin: the further away from the origin the smaller the isosurface values. 


\begin{figure}[h!t!b!]
  \begin{subfigure}[b]{0.25\textwidth}
    \includegraphics[trim=1.0cm 1.0 0.0cm 1.0cm,clip=true,width=\textwidth]{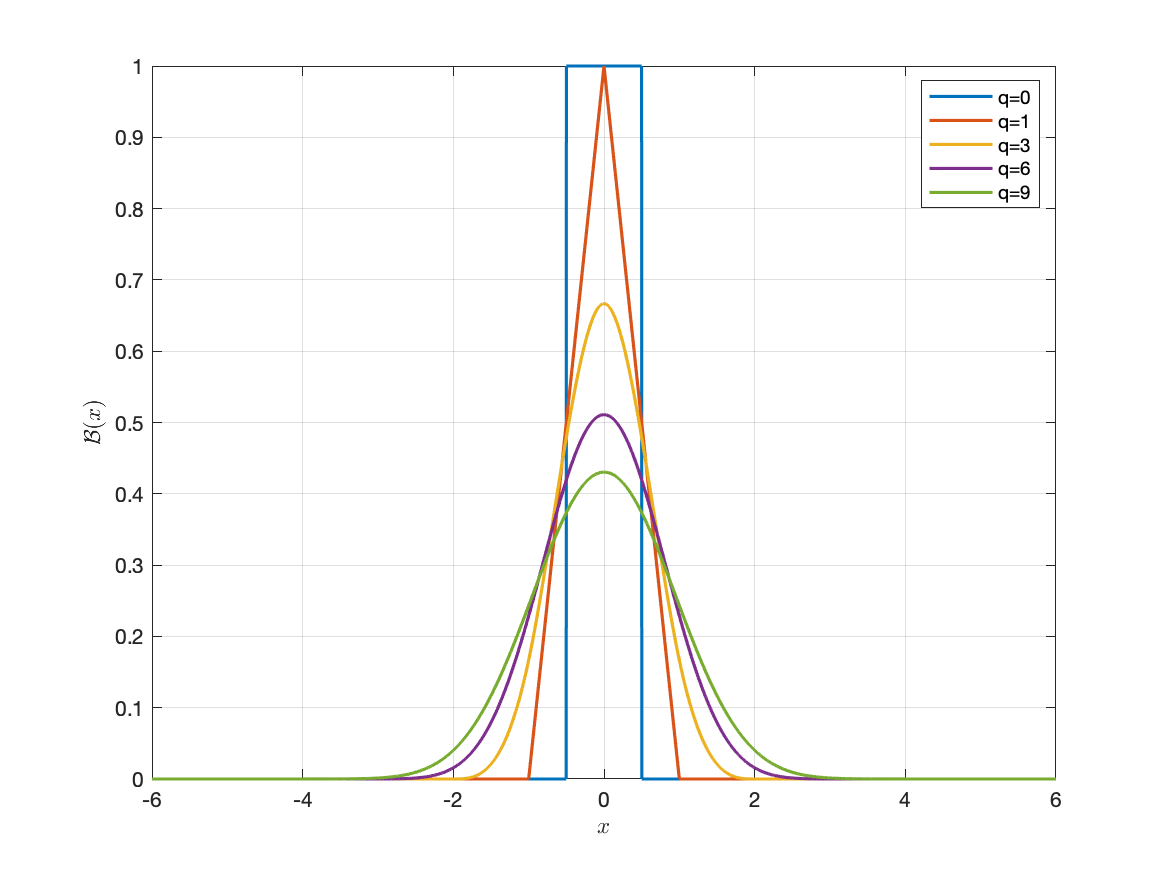}
\end{subfigure}
  \begin{subfigure}[b]{0.24\textwidth}
    \includegraphics[trim=2.0cm 1.0cm 3.0cm 0.0cm,clip=true,width=\textwidth]{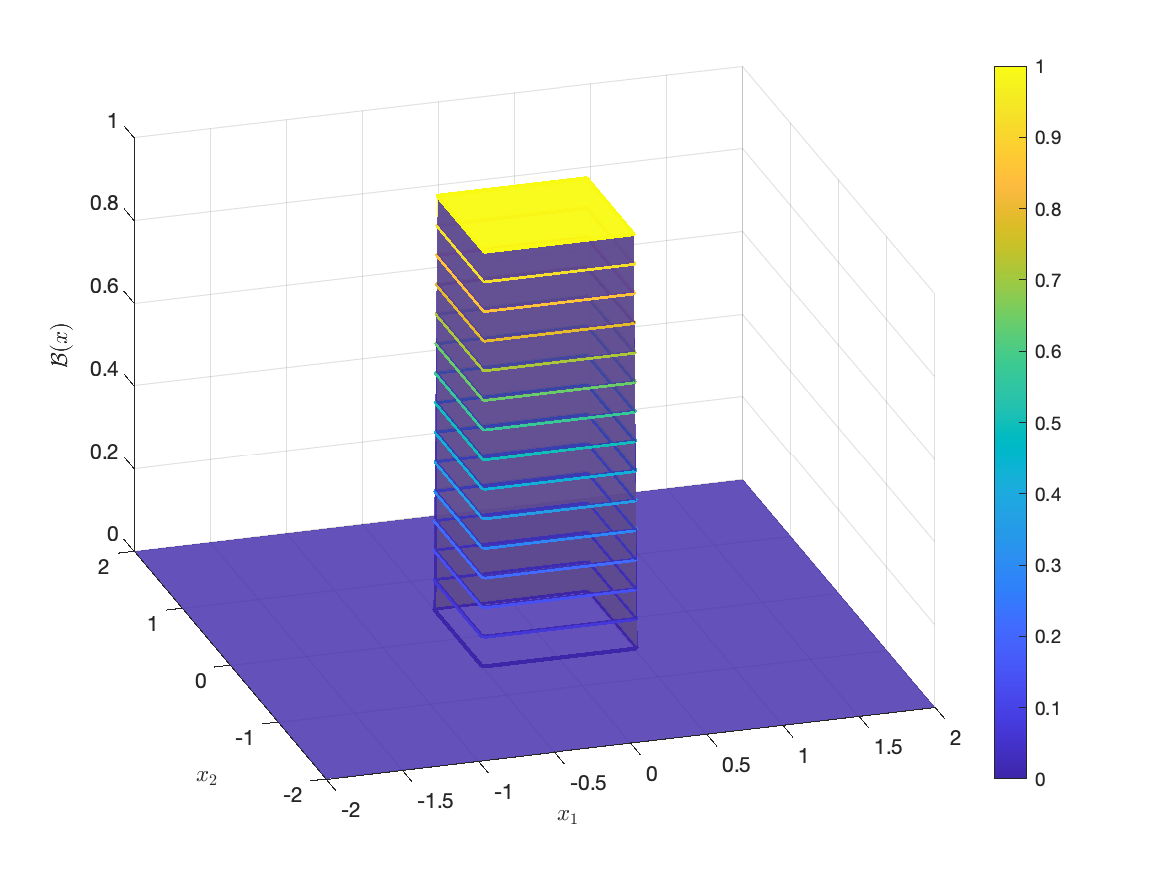}
\end{subfigure}
\begin{subfigure}[b]{0.24\textwidth}
    \includegraphics[trim=2.0cm 0.0 3.0cm 0.0,clip=true,width=\textwidth]{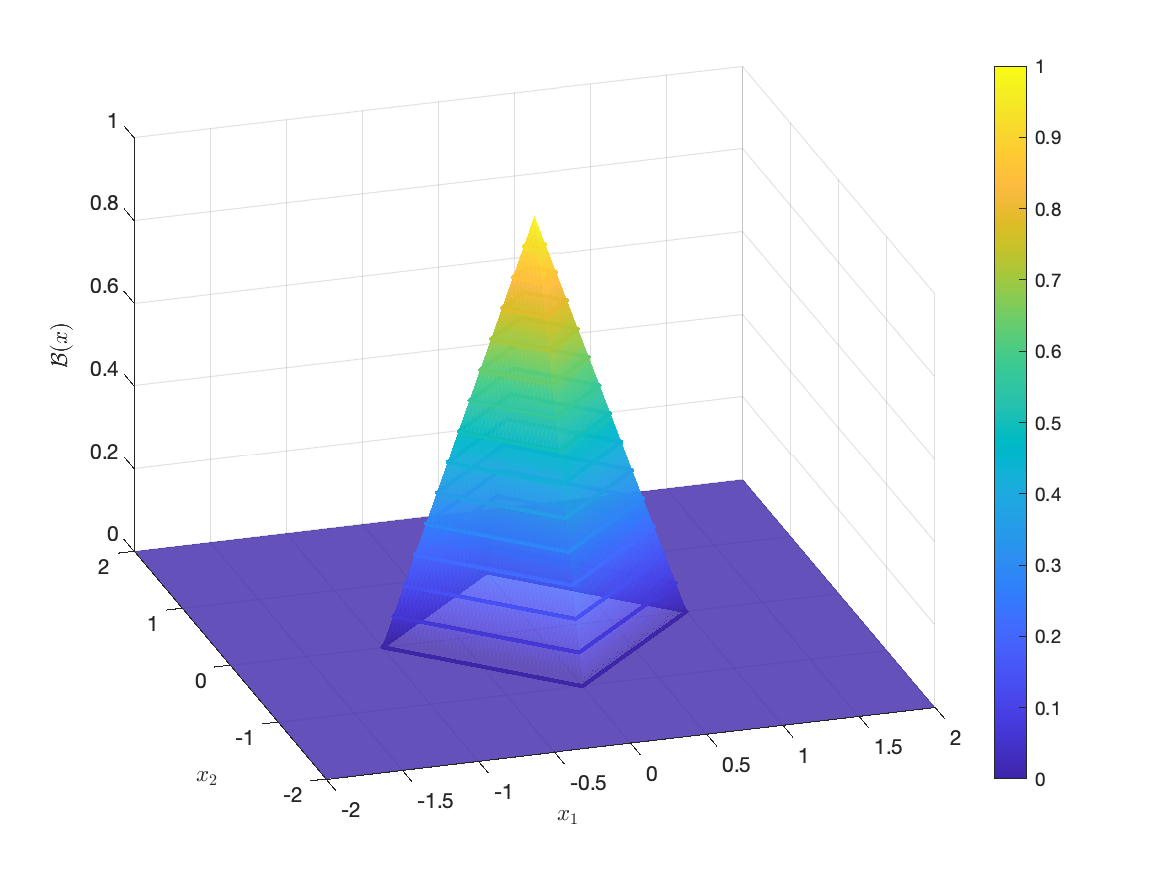}
\end{subfigure}
\begin{subfigure}[b]{0.24\textwidth}
    \includegraphics[trim=2.5cm 0.0 2.1cm 0.0,clip=true,width=\textwidth]{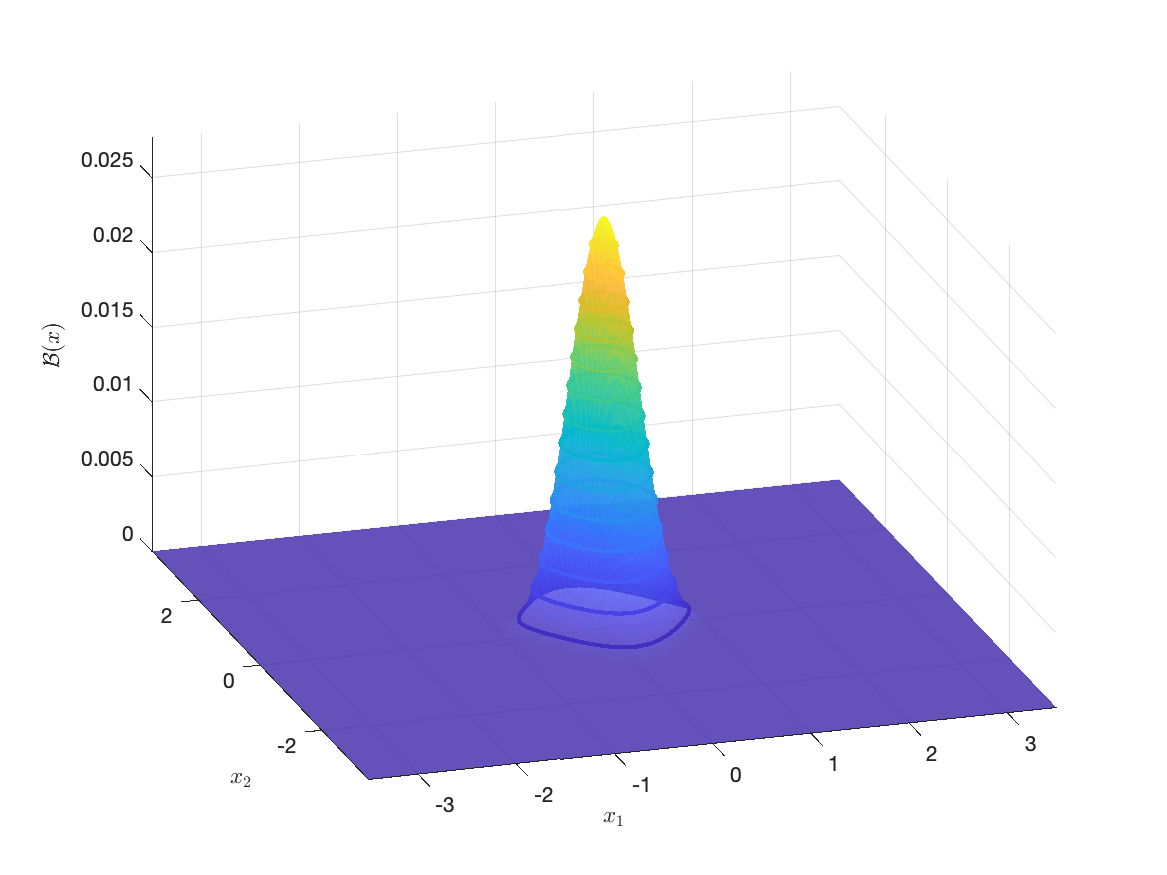}
\end{subfigure}
  		\caption{From left to right: One dimensional $\B\LRp{\x}$ functions for RePU with $\q = \LRc{0,1,3, 6,9}$, and two dimensional $\Bc\LRp{\x}$ functions for RePU with $\q = \LRc{0,1,5}$. The surfaces of $\Bc\LRp{\x}$ are plot with $15$ contours at $15$ values equally space from $10^{-6}$ to $\Bc\LRp{0,\B\LRp{0,1}}$.}
  		\figlab{Bfunc2D}
\end{figure}

\begin{figure}[h!t!b!]
\begin{subfigure}[b]{0.24\textwidth}
    \includegraphics[trim=2.5cm 0.0 3.0cm 0.0,clip=true,width=\textwidth]{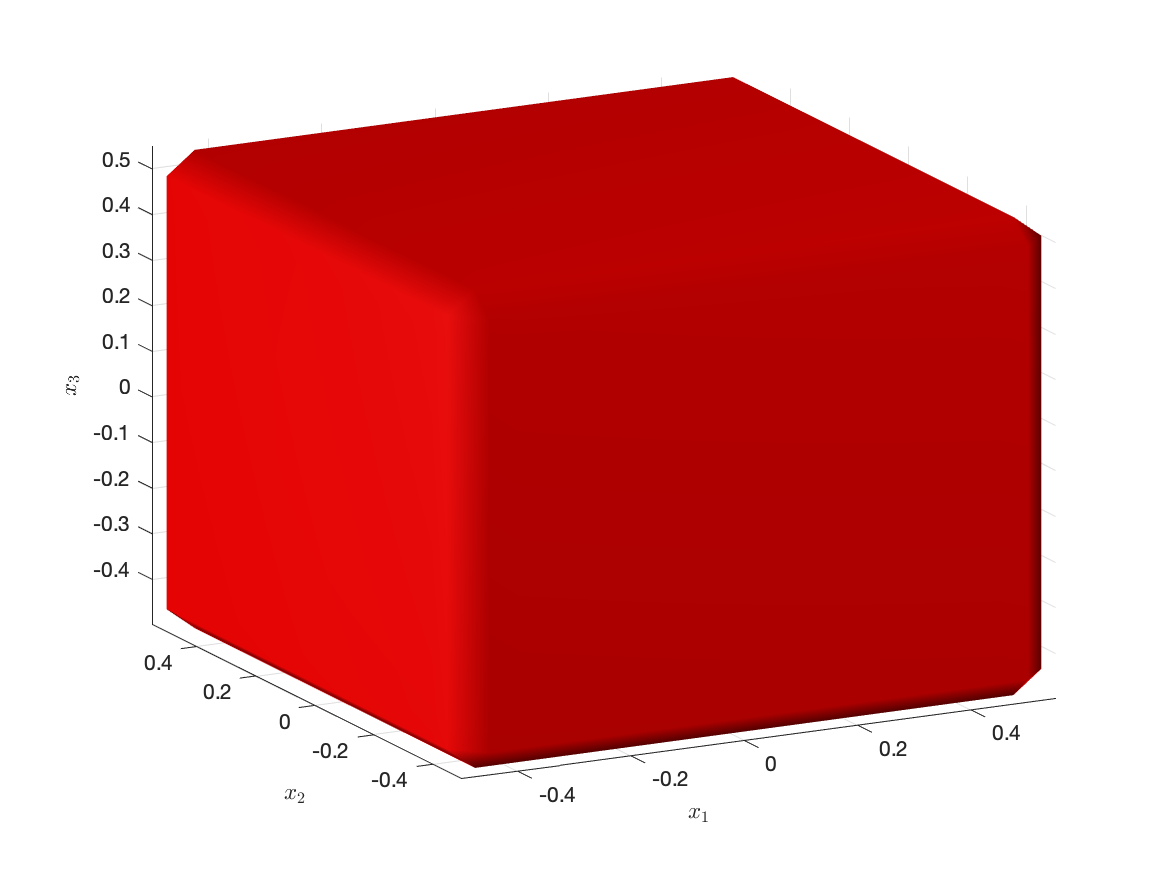}
\end{subfigure}
\begin{subfigure}[b]{0.24\textwidth}
    \includegraphics[trim=2.5cm 0.0 3.0cm 0.0,clip=true,width=\textwidth]{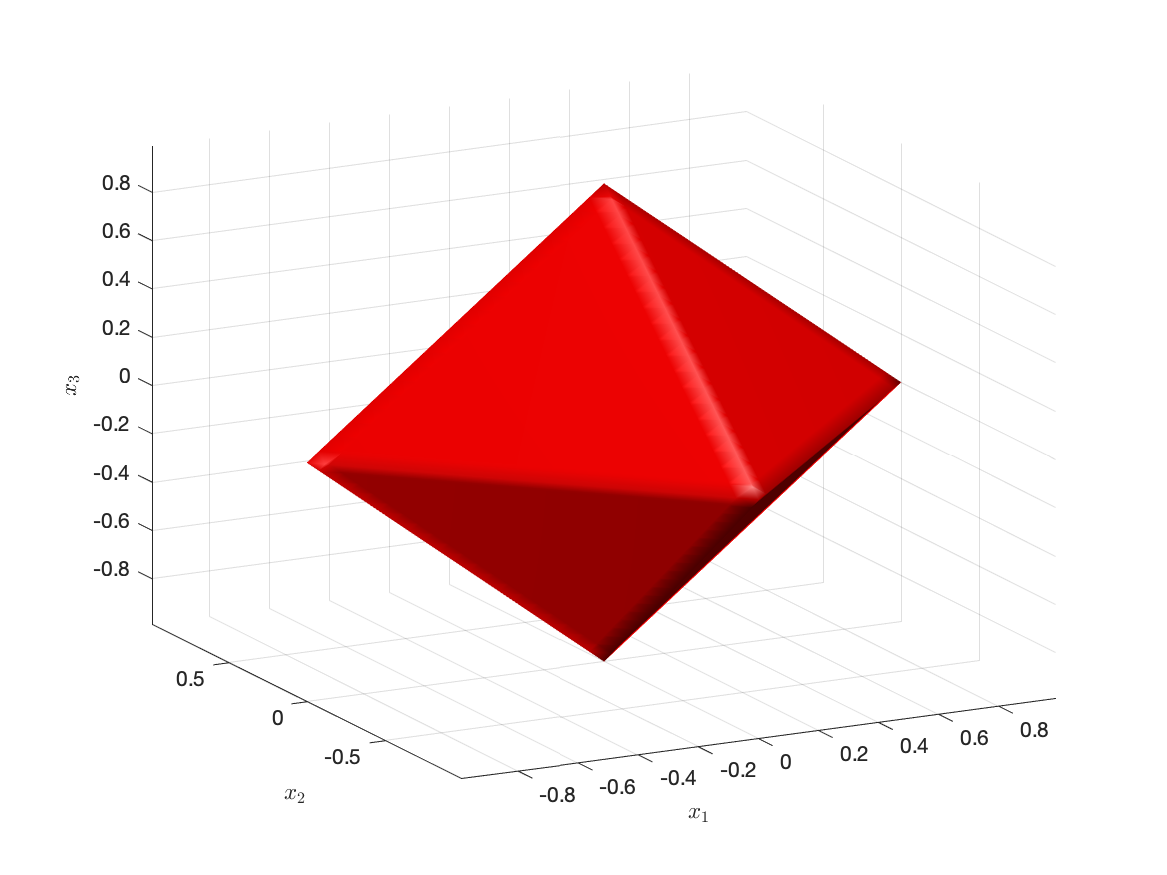}
\end{subfigure}
\begin{subfigure}[b]{0.24\textwidth}
    \includegraphics[trim=2.5cm 0.0 3.0cm 0.0,clip=true,width=\textwidth]{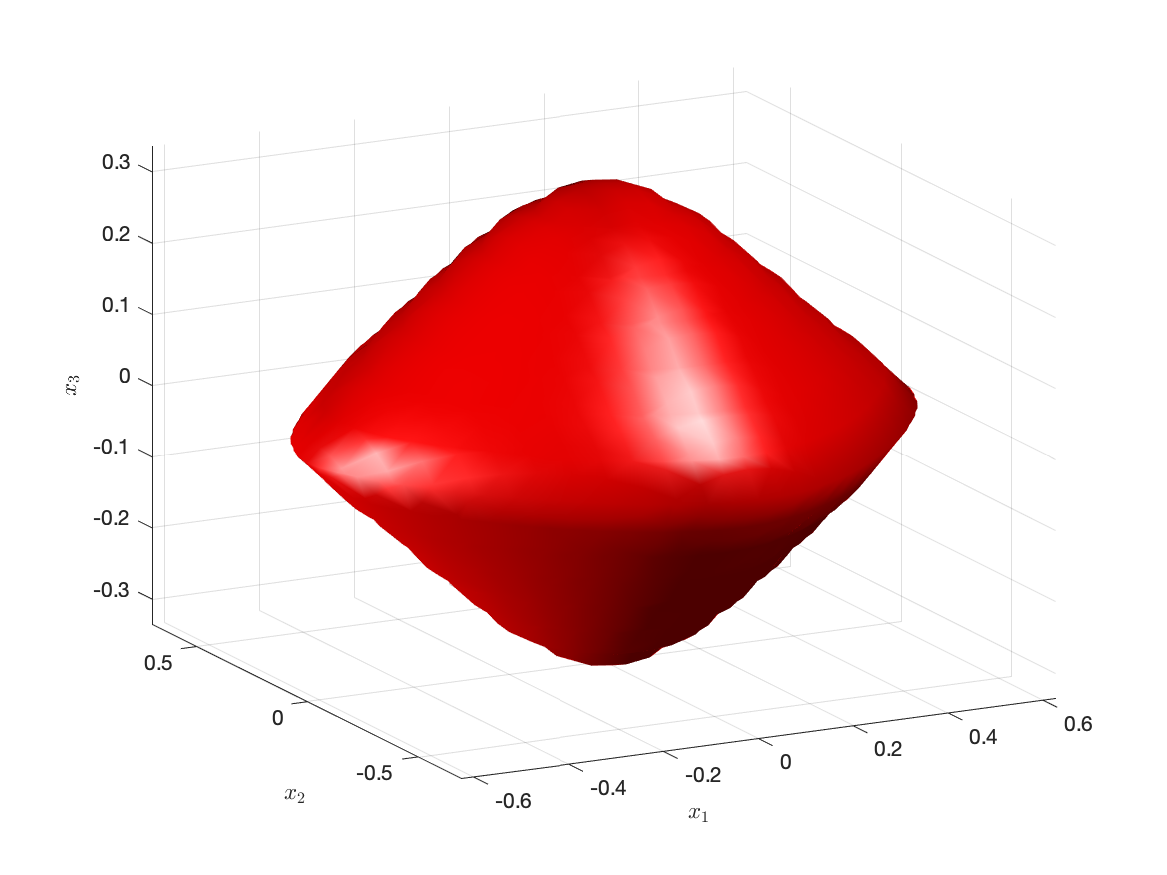}
\end{subfigure}
\begin{subfigure}[b]{0.24\textwidth}
    \includegraphics[trim=2.5cm 0.0 3.0cm 0.0,clip=true,width=\textwidth]{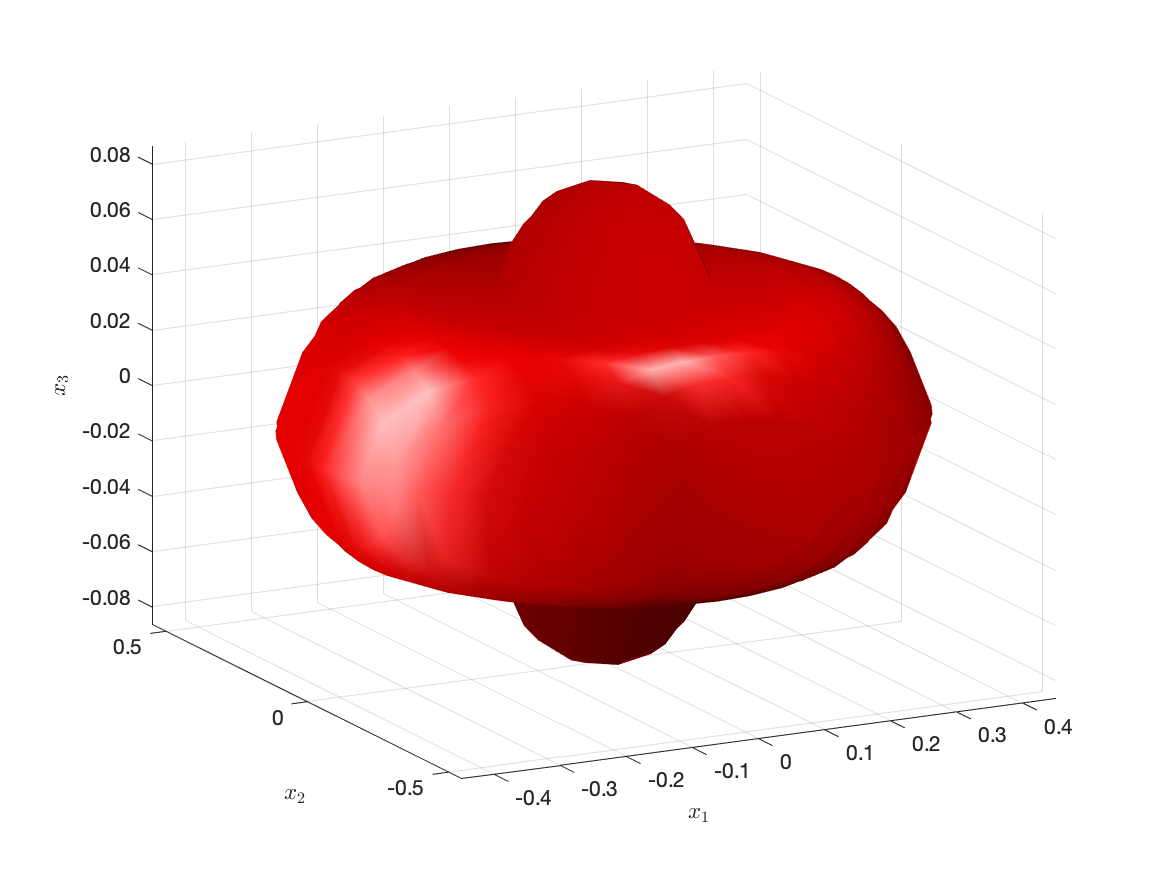}
\end{subfigure}
  		\caption{From left to right: Three dimensional $\Bc\LRp{\x}$ functions for RePU with $\q = \LRc{0,1,3,5}$. The isosurfaces are plotted for  $\Bc\LRp{\x} = \B\LRp{0,\B\LRp{0,1}}\times 10^{-2}$.}
  		\figlab{Bfunc3D}
\end{figure}

\begin{figure}[h!t!b!]
\begin{subfigure}[b]{0.24\textwidth}
    \includegraphics[trim=2.5cm 0.0 3.0cm 0.0,clip=true,width=\textwidth]{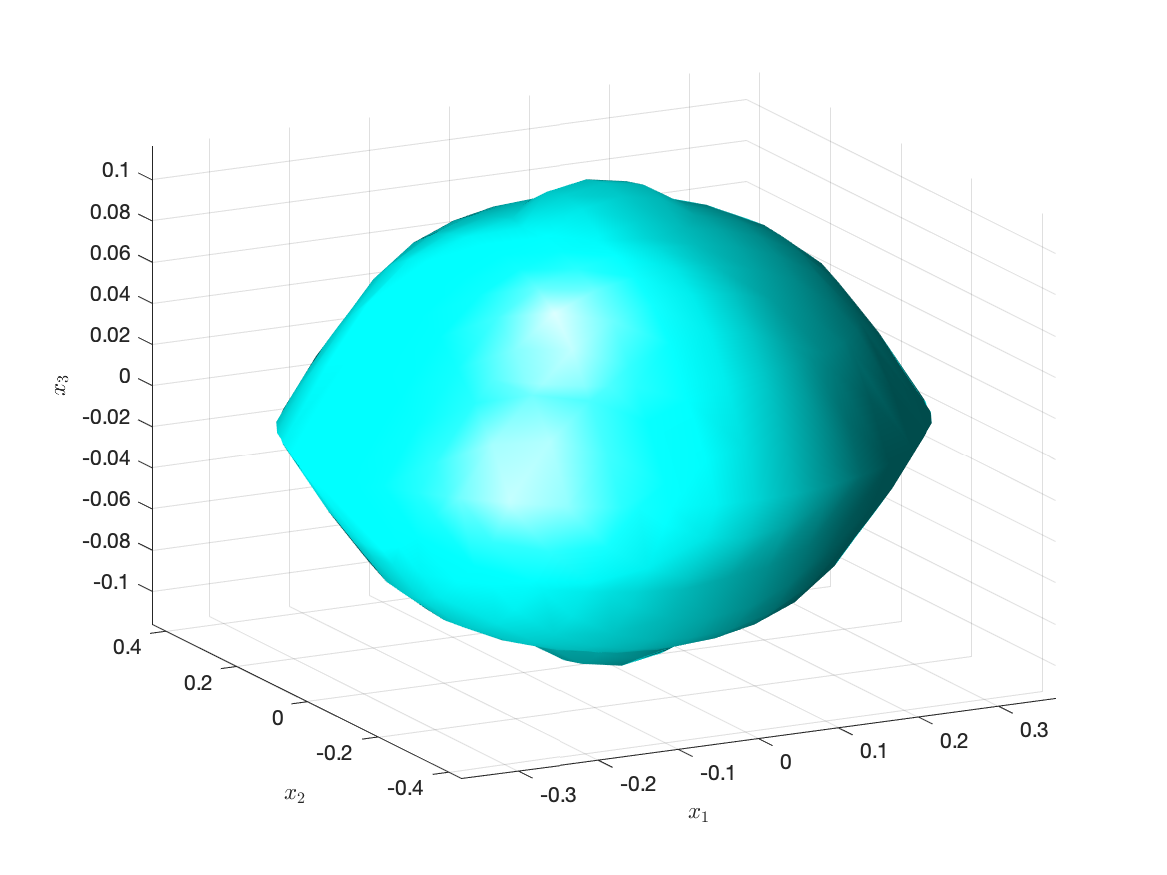}
\end{subfigure}
\begin{subfigure}[b]{0.24\textwidth}
    \includegraphics[trim=2.5cm 0.0 3.0cm 
    0.0,clip=true,width=\textwidth]{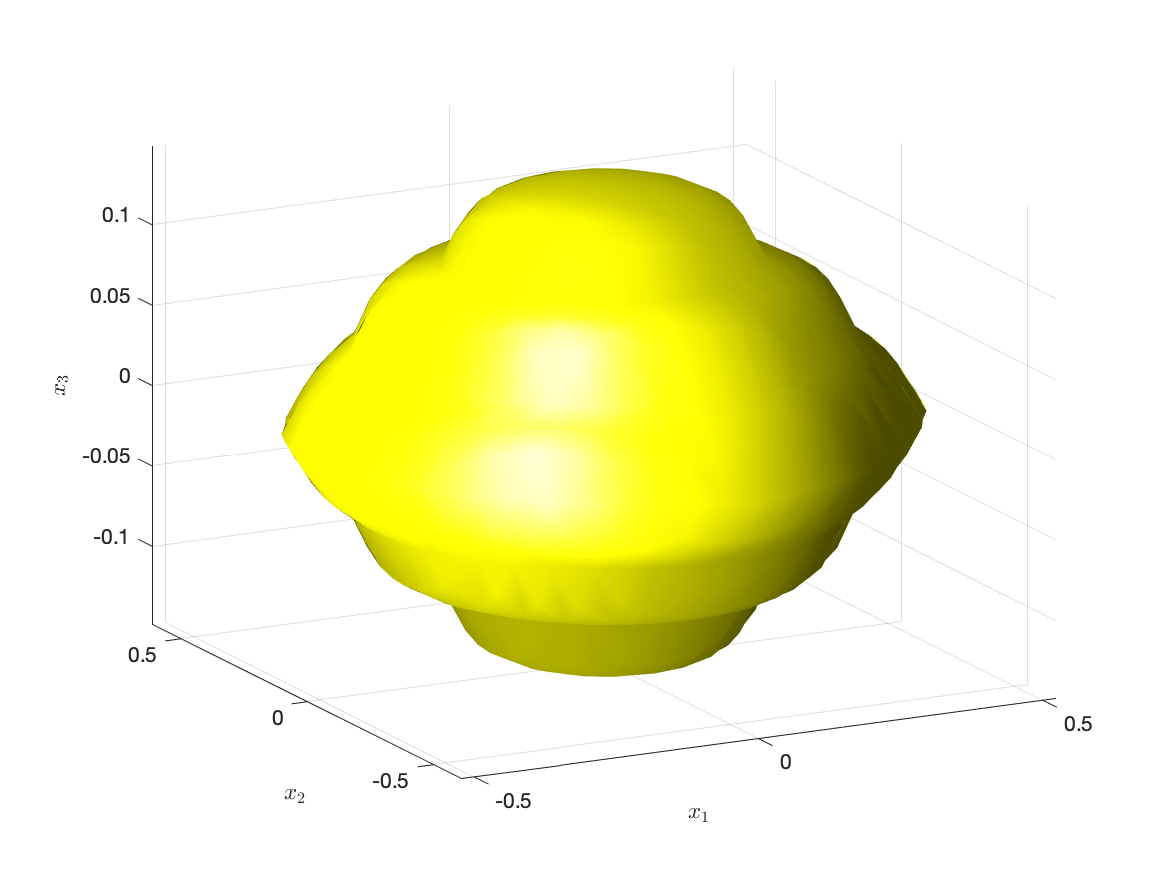}
\end{subfigure}
\begin{subfigure}[b]{0.24\textwidth}
    \includegraphics[trim=2.5cm 0.0 3.0cm 
    0.0,clip=true,width=\textwidth]{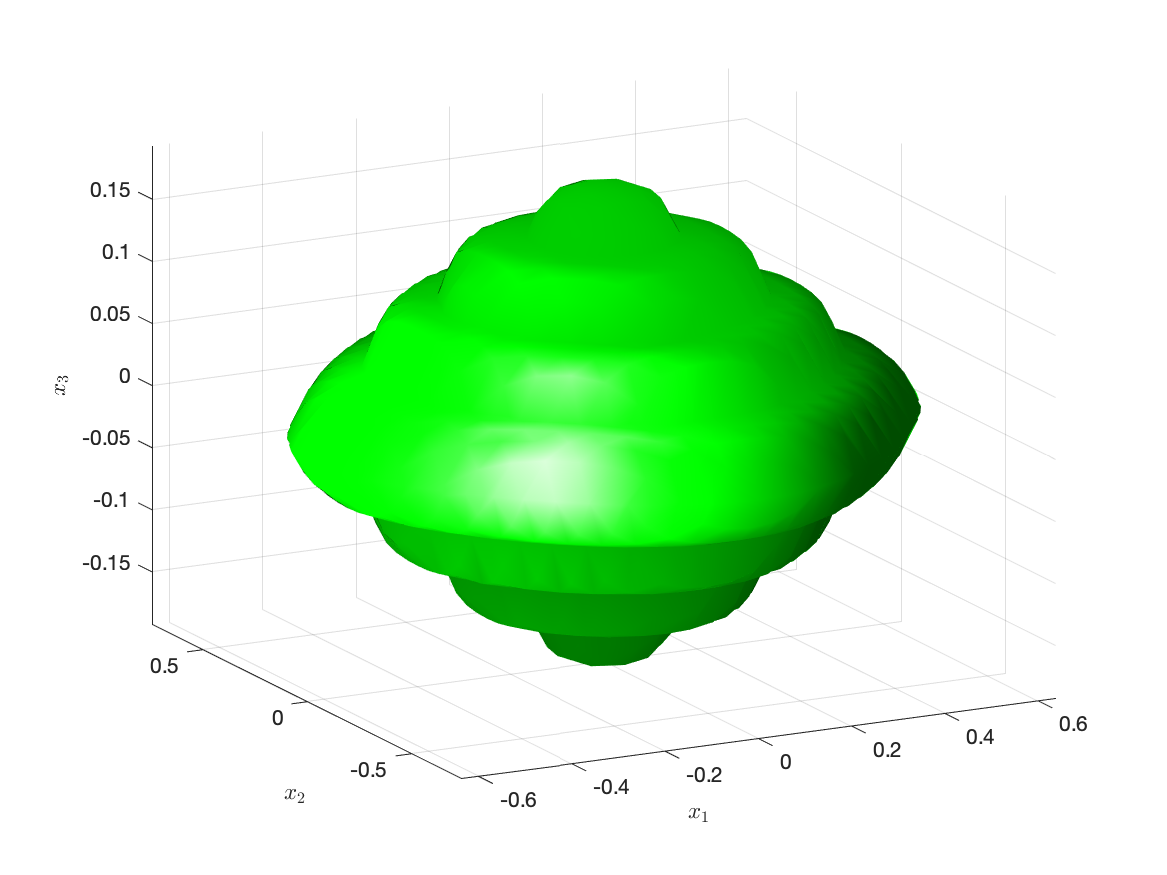}
\end{subfigure}
\begin{subfigure}[b]{0.24\textwidth}
    \includegraphics[trim=4.5cm 0.0 2.0cm 
    0.0,clip=true,width=\textwidth]{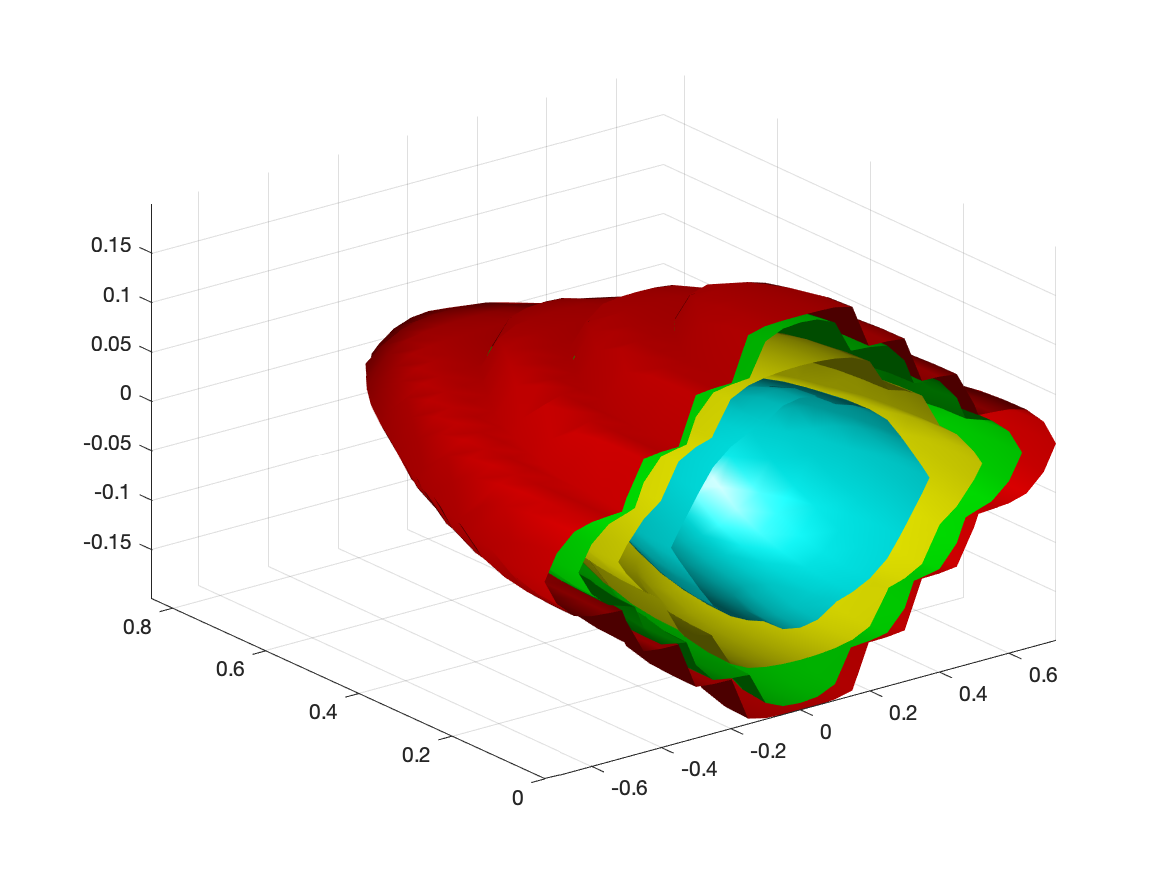}
\end{subfigure}
  		\caption{Three dimensional $\Bc\LRp{\x}$ functions for RePU with $\q = 4$. The first three isosurfaces (from left to right) are plotted at  $ \B\LRp{0,\B\LRp{0,1}}\times \LRc{10^{-1}, 10^{-2},10^{-3}}$, and the right most subfigure shows  half of isosurfaces at  $ \B\LRp{0,\B\LRp{0,1}}\times \LRc{10^{-1}, 10^{-2},10^{-3},10^{-4}}$, where red is for the isosurface at $\B\LRp{0,\B\LRp{0,1}}\times 10^{-4}$.}
  		\figlab{Bfunc3Dq4}
\end{figure}

To verify that activation functions in the generalized sigmoidal class behave similarly from our unified framework we plot the $\B$ functions of the standard sigmoid, softplus, and arctangent activation functions in Figure \figref{sigmoid} for $\n = \LRc{1,2,3}$. As can be seen,  the $\B$ functions, though have different values, share similar shapes. Note that for $\n=3$ we plot the isosurfaces at $5\times 10^{-2}$ times the largest value of the corresponding $\B$ functions.

\begin{figure}[h!t!b!]
    \includegraphics[trim=2.5cm 7.0cm 3.0cm 8.0cm,clip=true,width=0.32\textwidth]{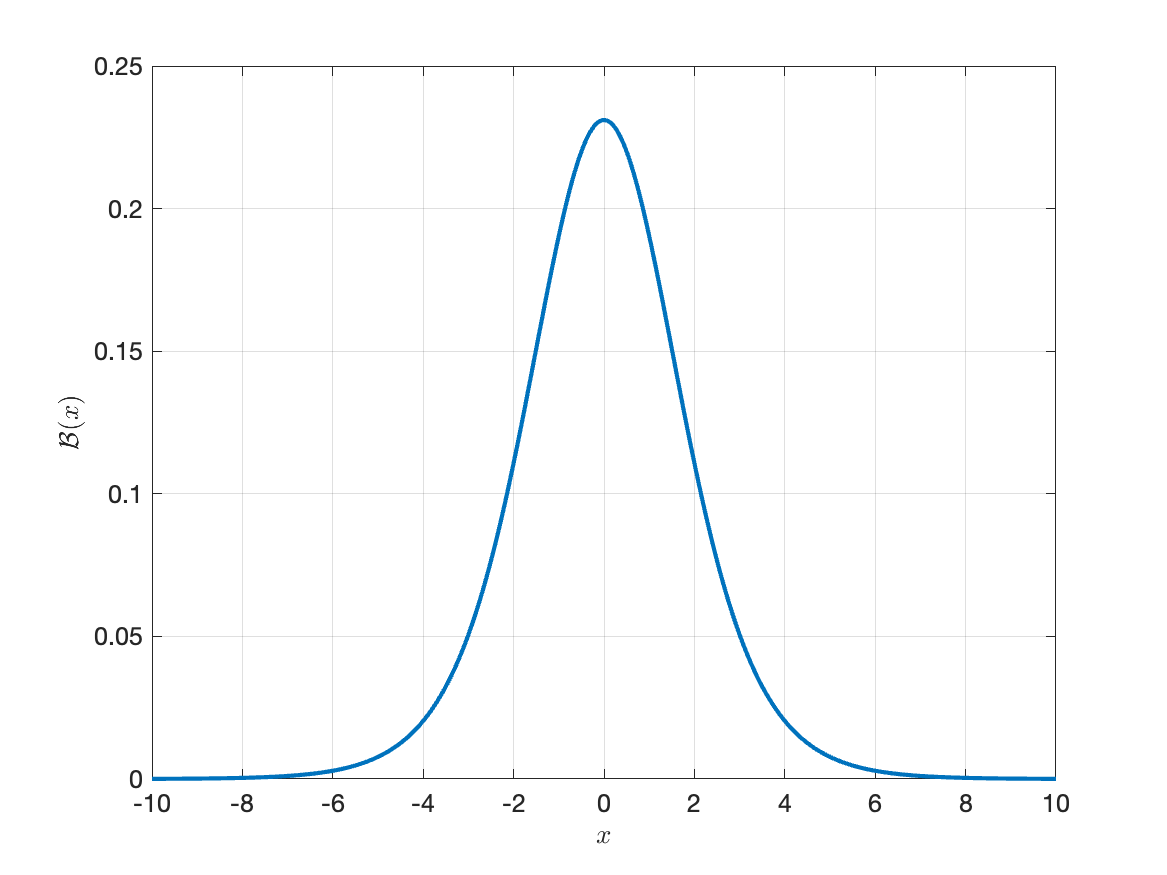}
    \includegraphics[trim=2.0cm 7.0cm 3.0cm
    8.0cm,clip=true,width=0.32\textwidth]{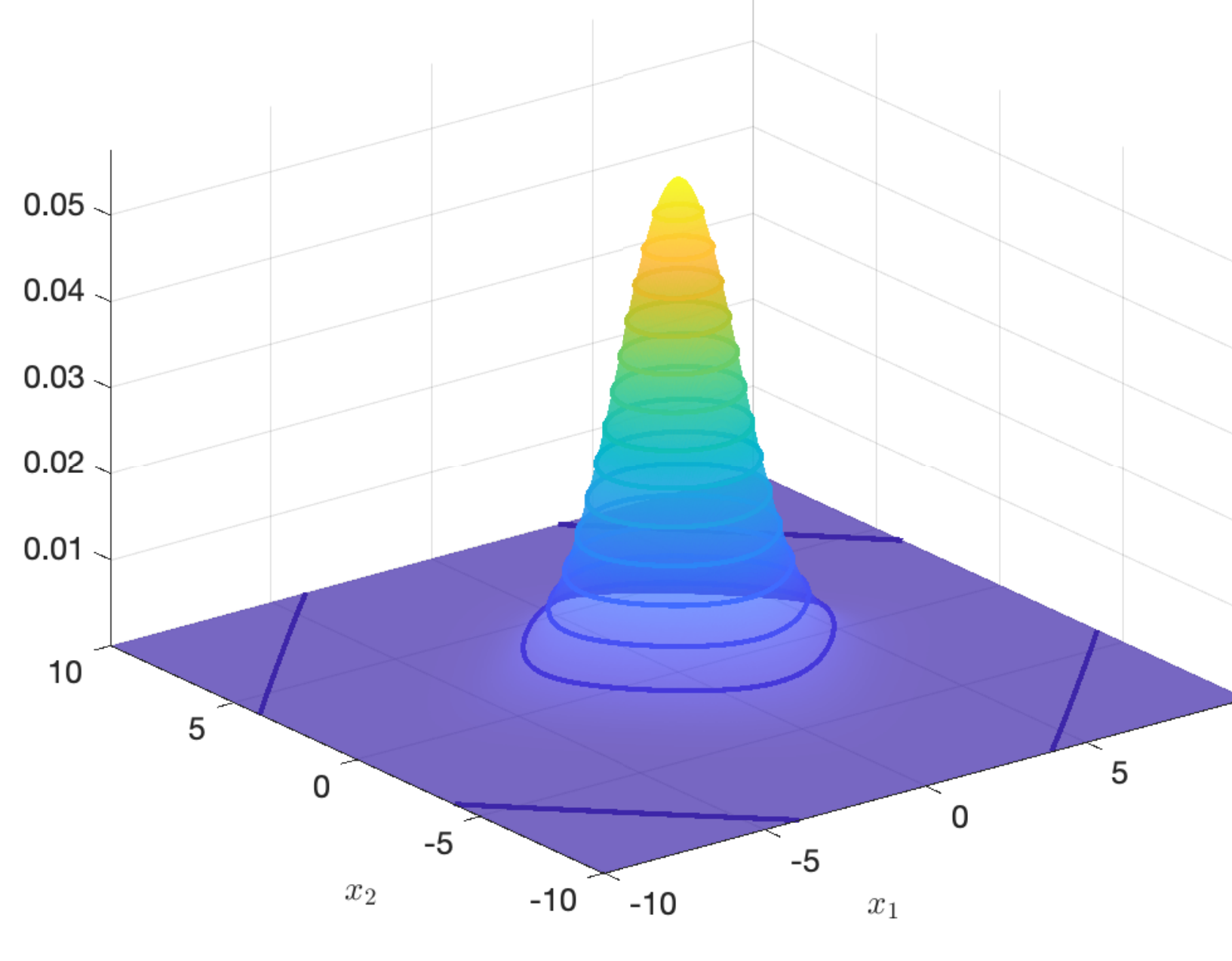}
    \includegraphics[trim=2.0cm 7.0cm 3.0cm 
    8.0cm,clip=true,width=0.32\textwidth]{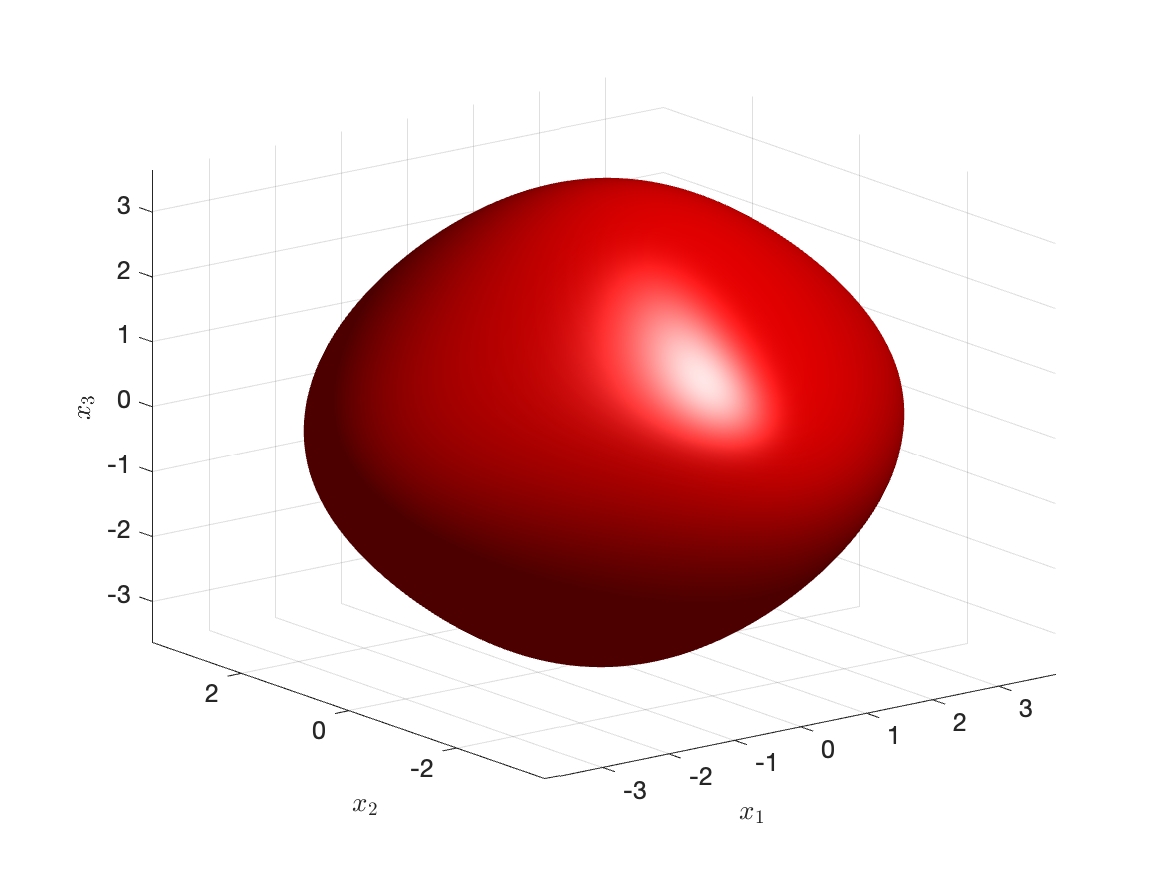}
  
    \includegraphics[trim=2.5cm 7.0cm 3.0cm 8.0cm,clip=true,width=0.32\textwidth]{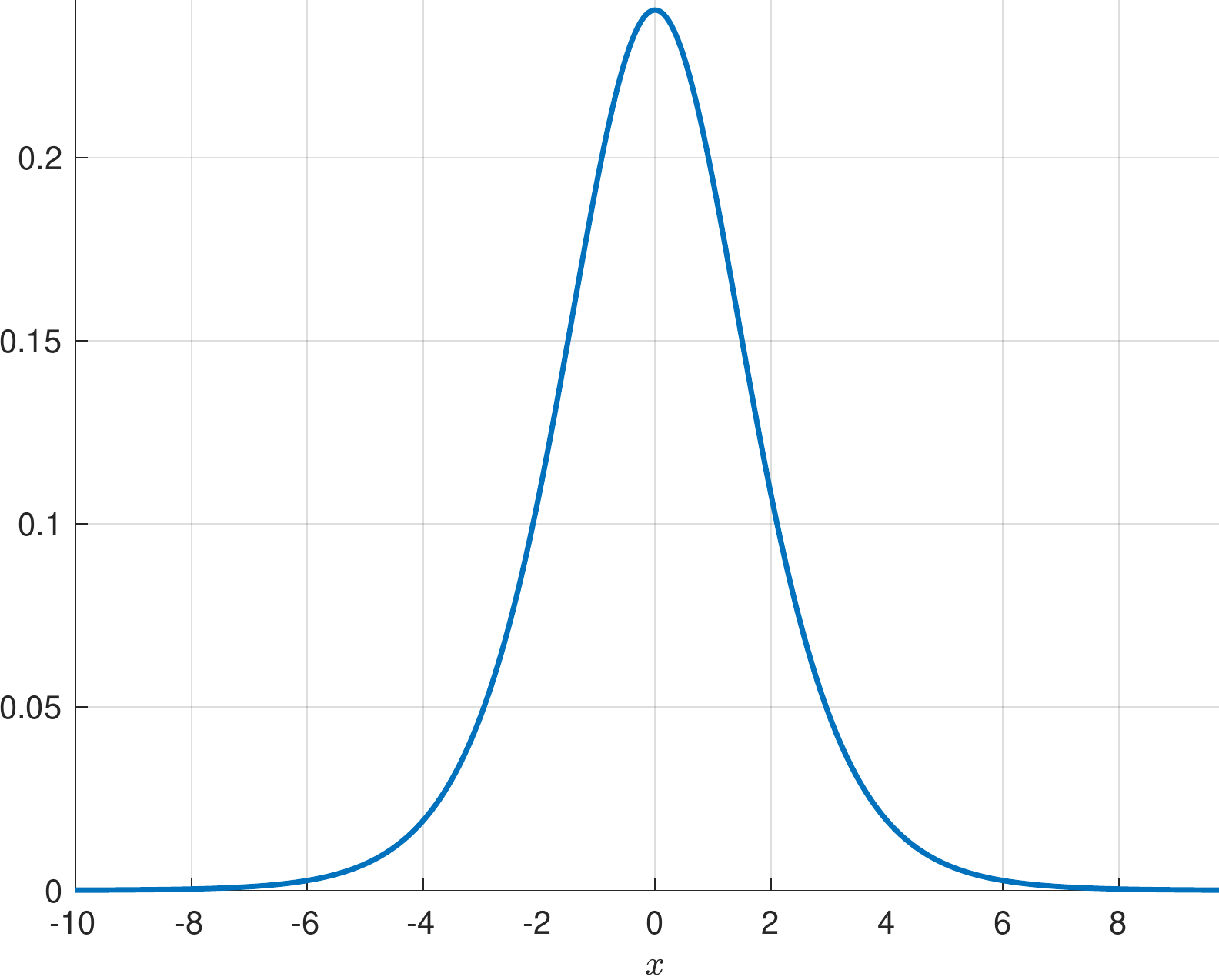}
    \includegraphics[trim=2.0cm 7.0cm 3.0cm
    8.0cm,clip=true,width=0.32\textwidth]{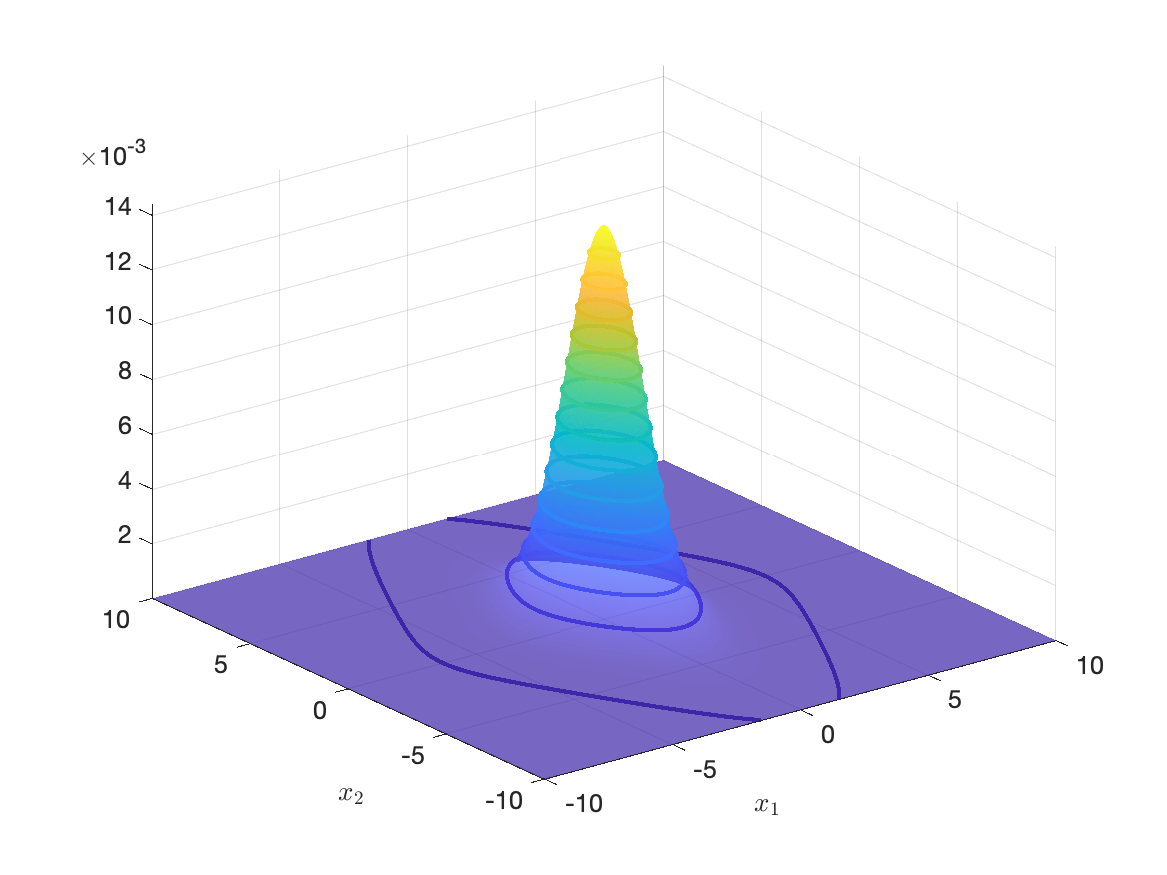}
    \includegraphics[trim=2.0cm 7.0cm 3.0cm 
    8.0cm,clip=true,width=0.32\textwidth]{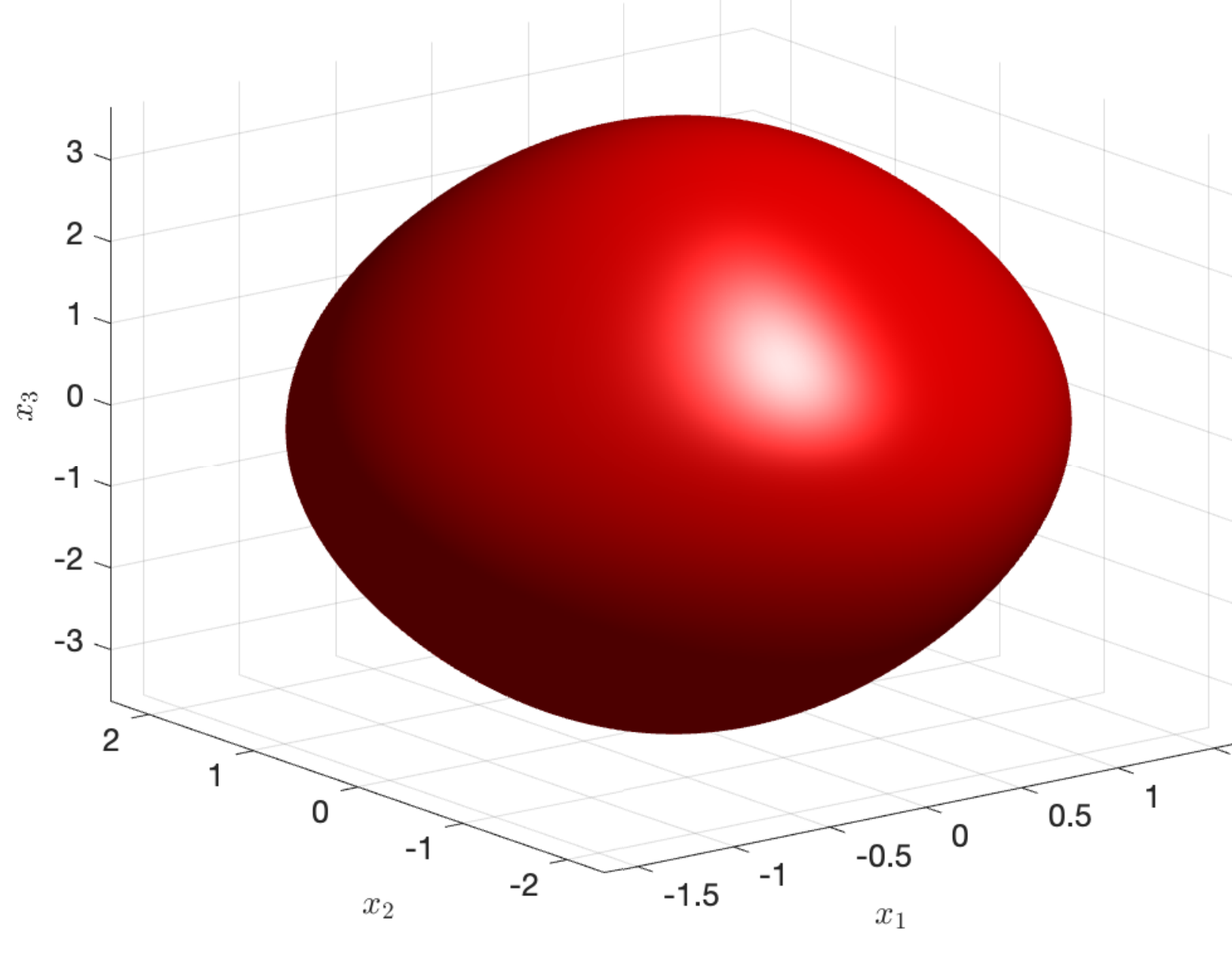}
  	
    \includegraphics[trim=2.5cm 7.0cm 3.0cm 8.0cm,clip=true,width=0.32\textwidth]{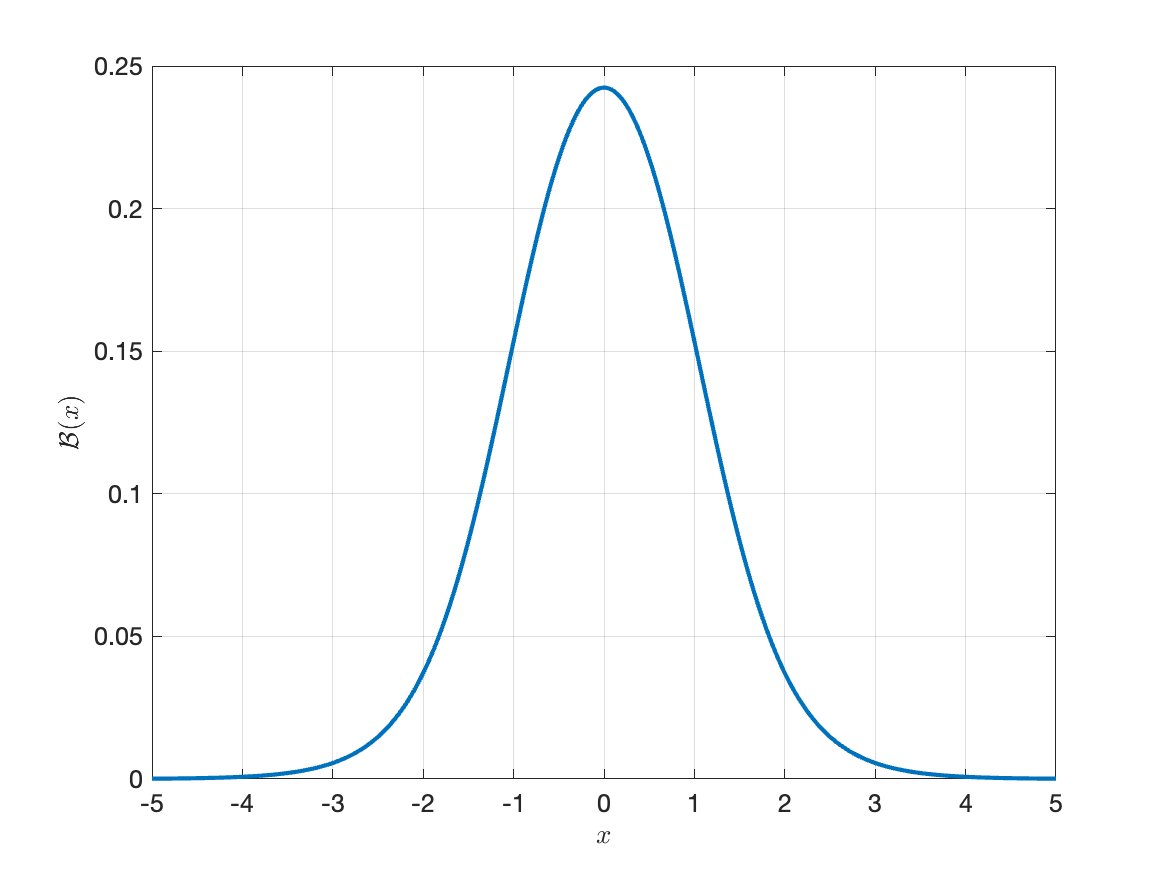}
    \includegraphics[trim=2.0cm 7.0cm 3.0cm
    8.0cm,clip=true,width=0.32\textwidth]{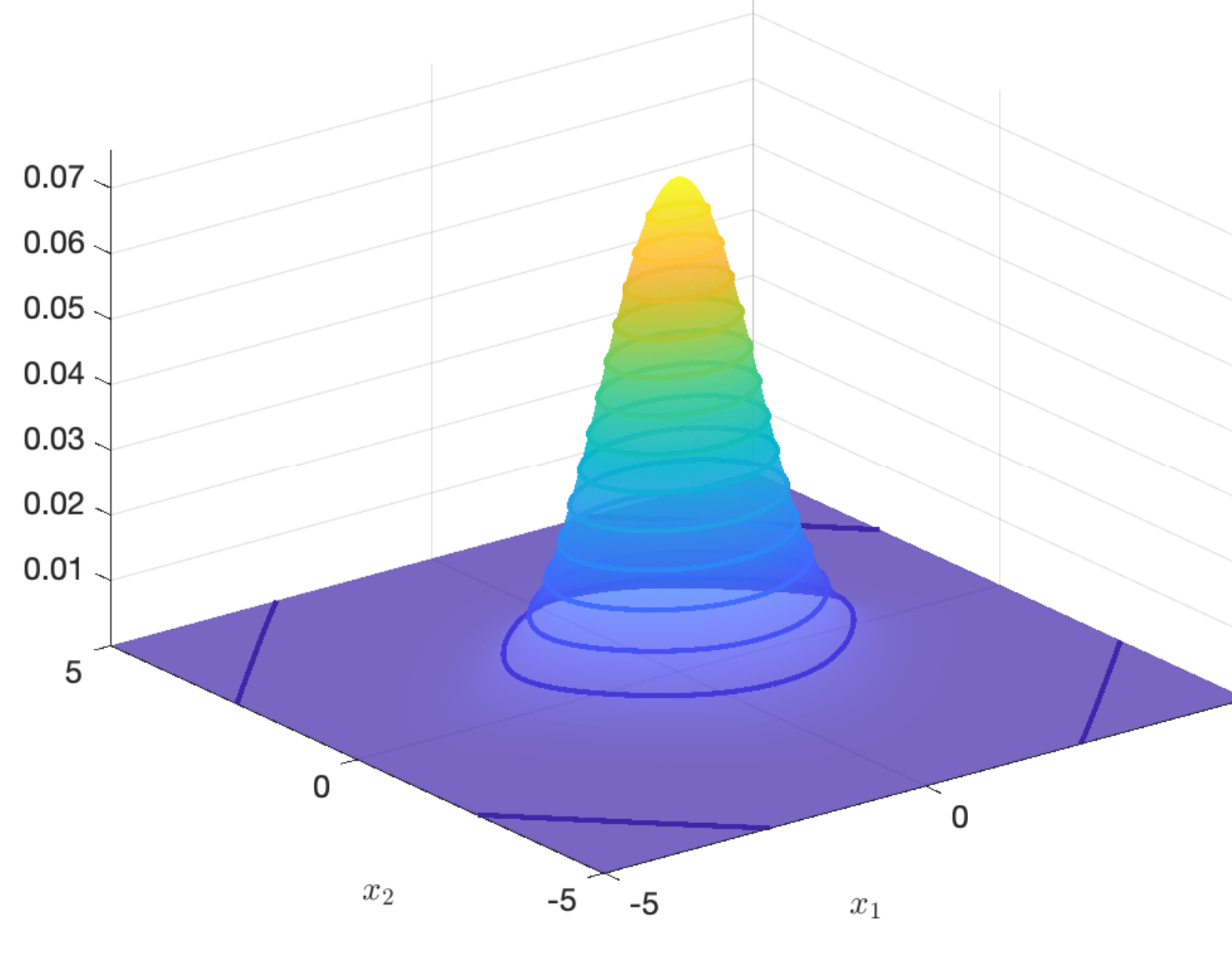}
    \includegraphics[trim=2.0cm 7.0cm 3.0cm 
    8.0cm,clip=true,width=0.32\textwidth]{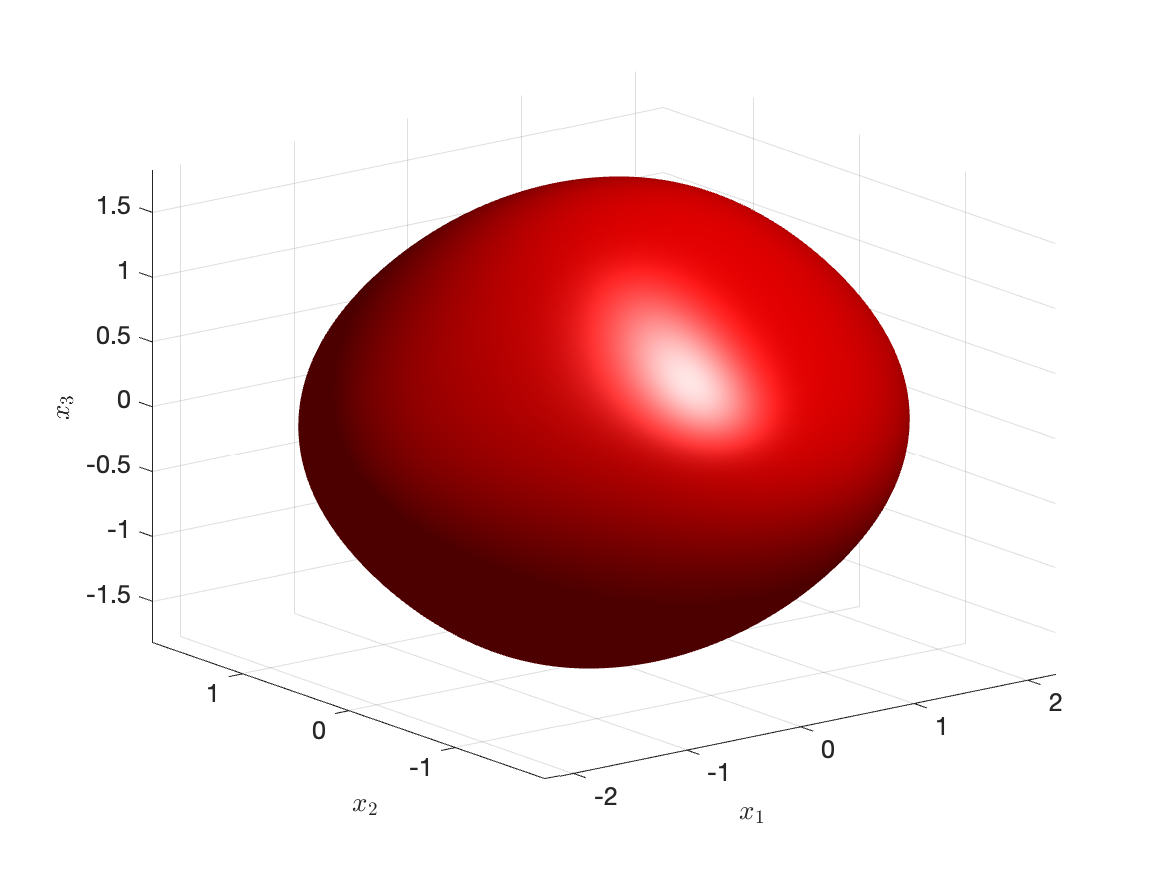}
  		\caption{ From left to right: One dimensional plot, two dimensional surface together with $15$ contours, and a three dimensional isosurface of $\B$ functions. From top row to bottom row: the standard sigmoid, softplus, and arctangent activation functions.
  	}
  		\figlab{sigmoid}
\end{figure}

In figure \figref{ELU} we present a snapshot of the $\B$ function of the ELU activation function in one, two, and three dimensions. While it has common features as other $\B$ functions such as decaying to zero at infinity, it possesses distinct features including asymmetric shape with positive and negative values. 

\begin{figure}[h!t!b!]
    \includegraphics[trim=2.5cm 7.0cm 3.0cm 8.0cm,clip=true,width=0.32\textwidth]{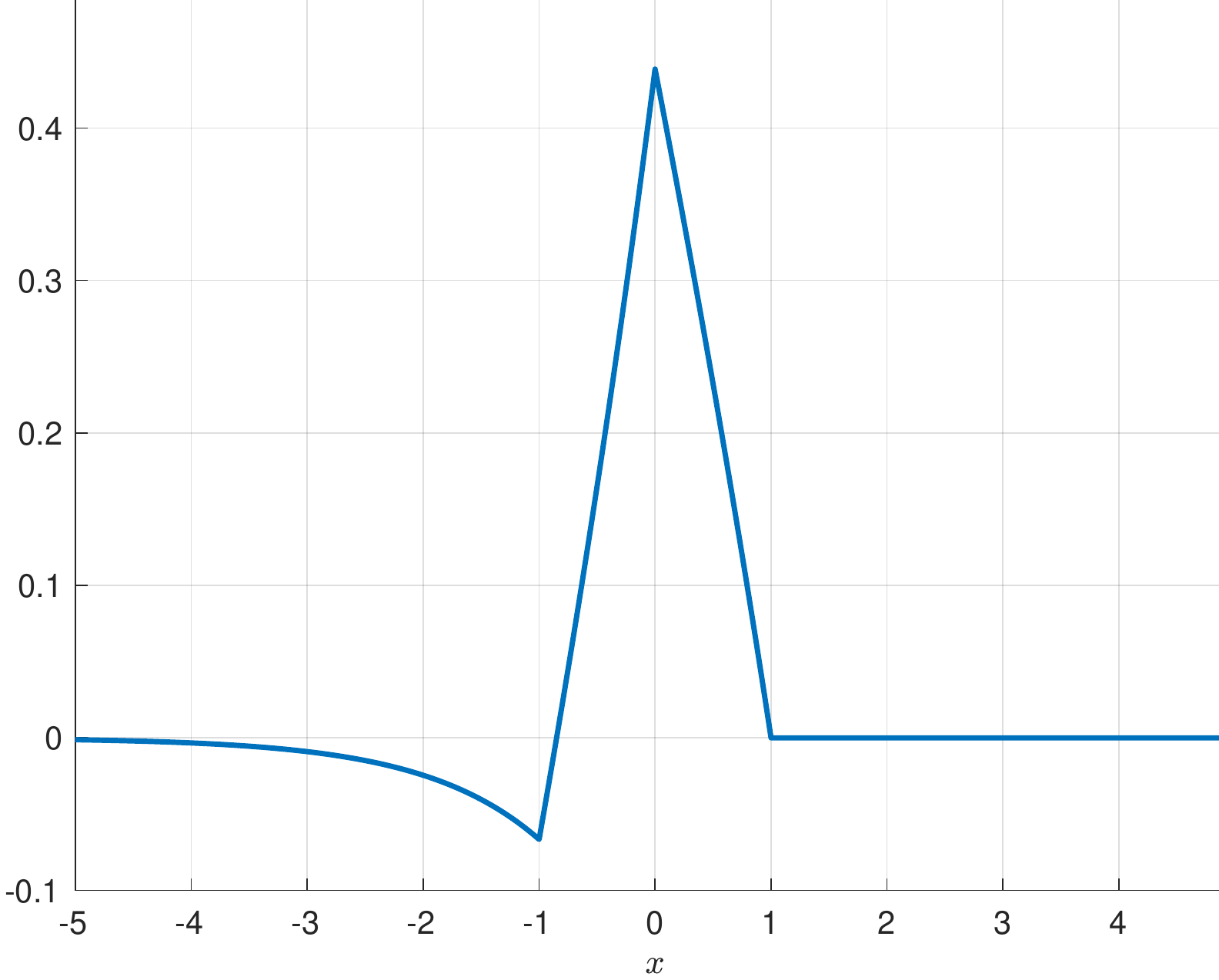}
    \includegraphics[trim=2.0cm 7.0cm 3.0cm
    8.0cm,clip=true,width=0.32\textwidth]{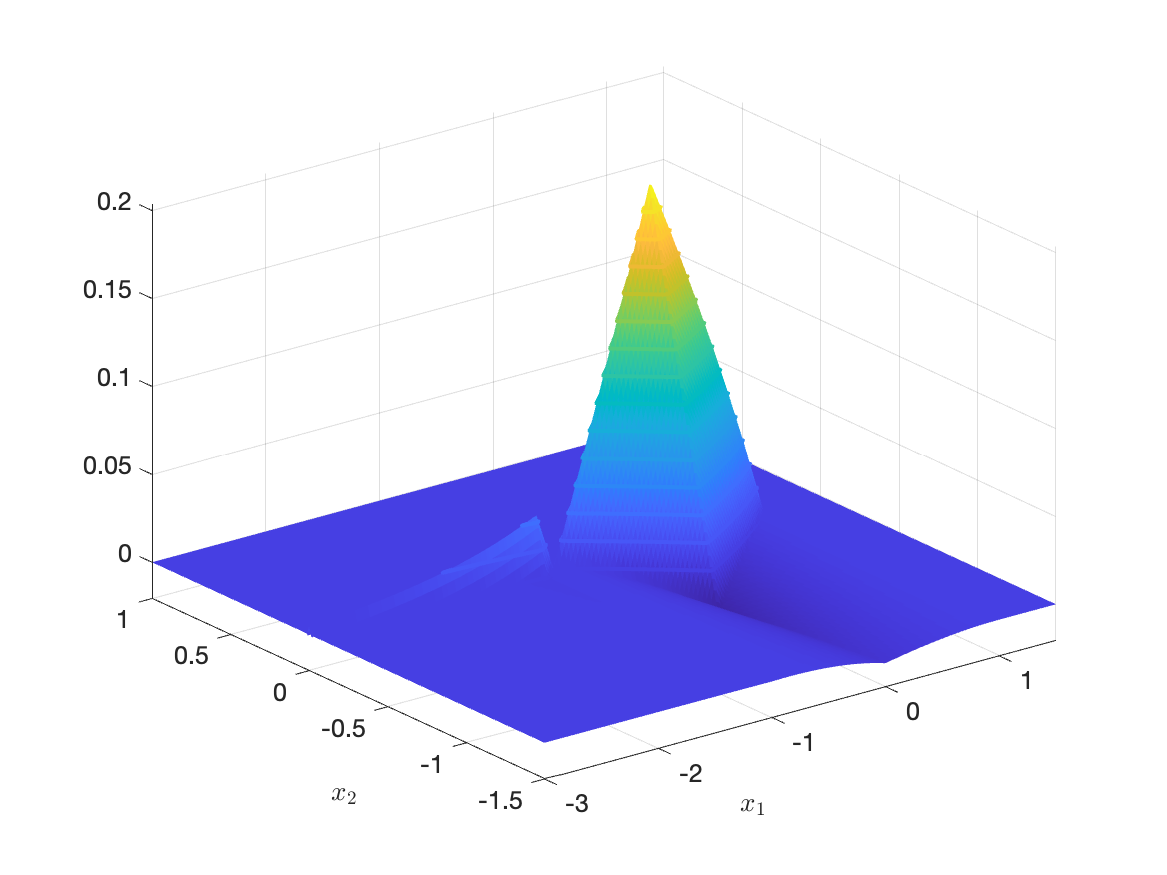}
    \includegraphics[trim=2.0cm 7.0cm 3.0cm 
    8.0cm,clip=true,width=0.32\textwidth]{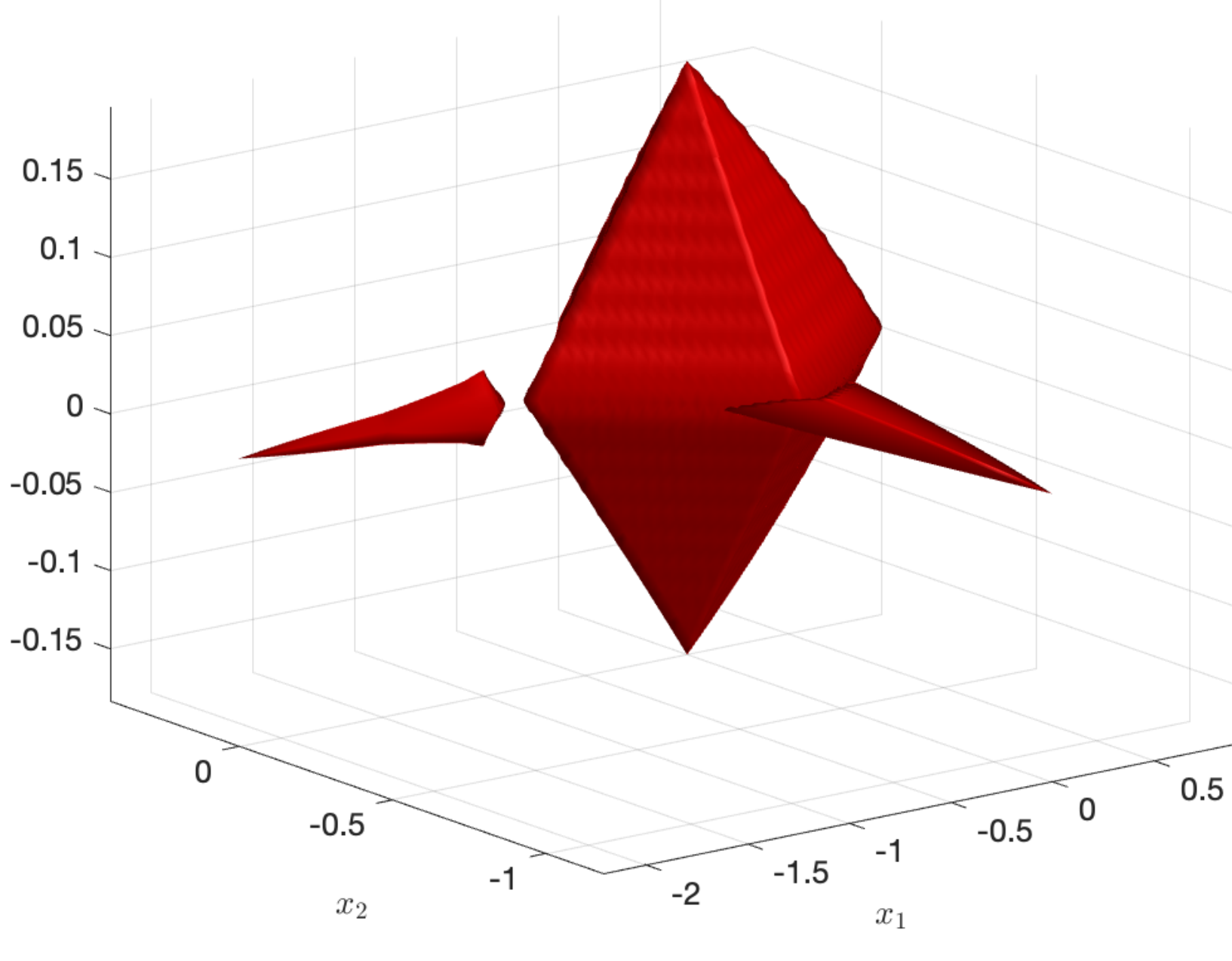}

  		\caption{ From left to right: One dimensional plot, two dimensional surface together with $15$ contours, and a three dimensional isosurface of the $\B$ function of ELU activation function.
  	}
  		\figlab{ELU}
\end{figure}

We have shown the  nAI proof for $\n$ dimensions is the same for GELU, SiLU, and Mish. It turns out that they are geometrically very similar and this can be seen in Figure \figref{GELUtoMish}. Note that for $\n=3$ we plot the isosurfaces at $5\times 10^{-2}$ times the largest value of the corresponding $\B$ functions.

\begin{figure}[h!t!b!]
    \includegraphics[trim=2.5cm 7.0cm 3.0cm 7.8cm,clip=true,width=0.32\textwidth]{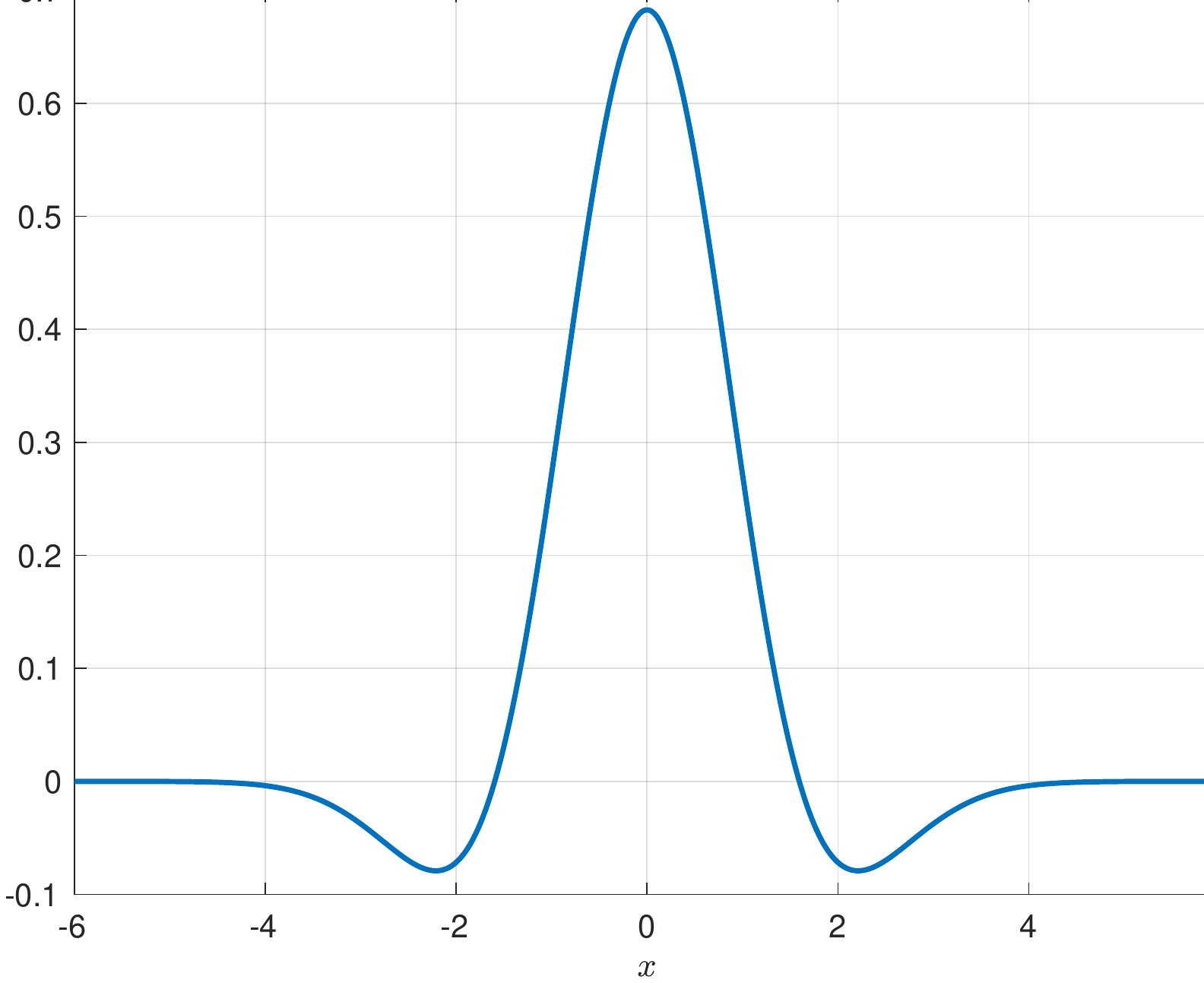}
    \includegraphics[trim=2.0cm 7.0cm 3.0cm
    8.0cm,clip=true,width=0.32\textwidth]{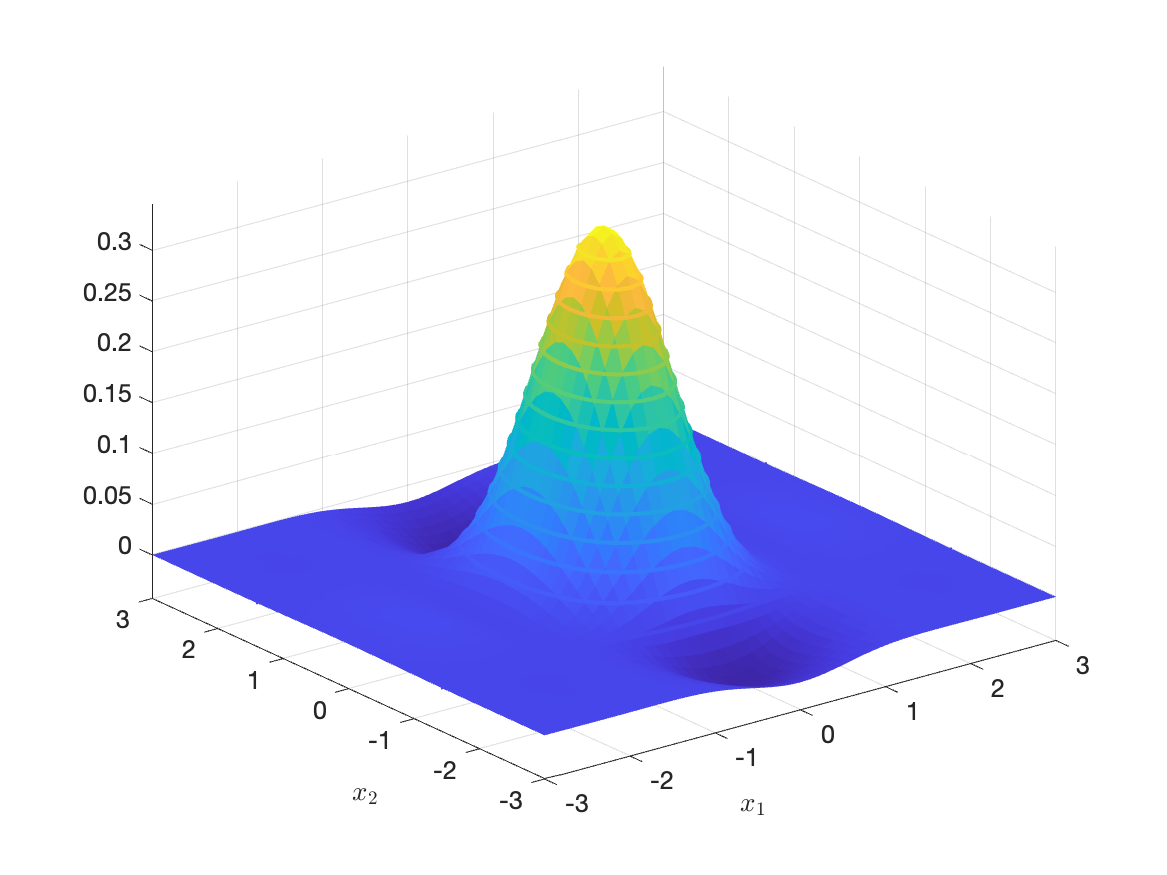}
    \includegraphics[trim=2.0cm 7.0cm 3.0cm 
    8.0cm,clip=true,width=0.32\textwidth]{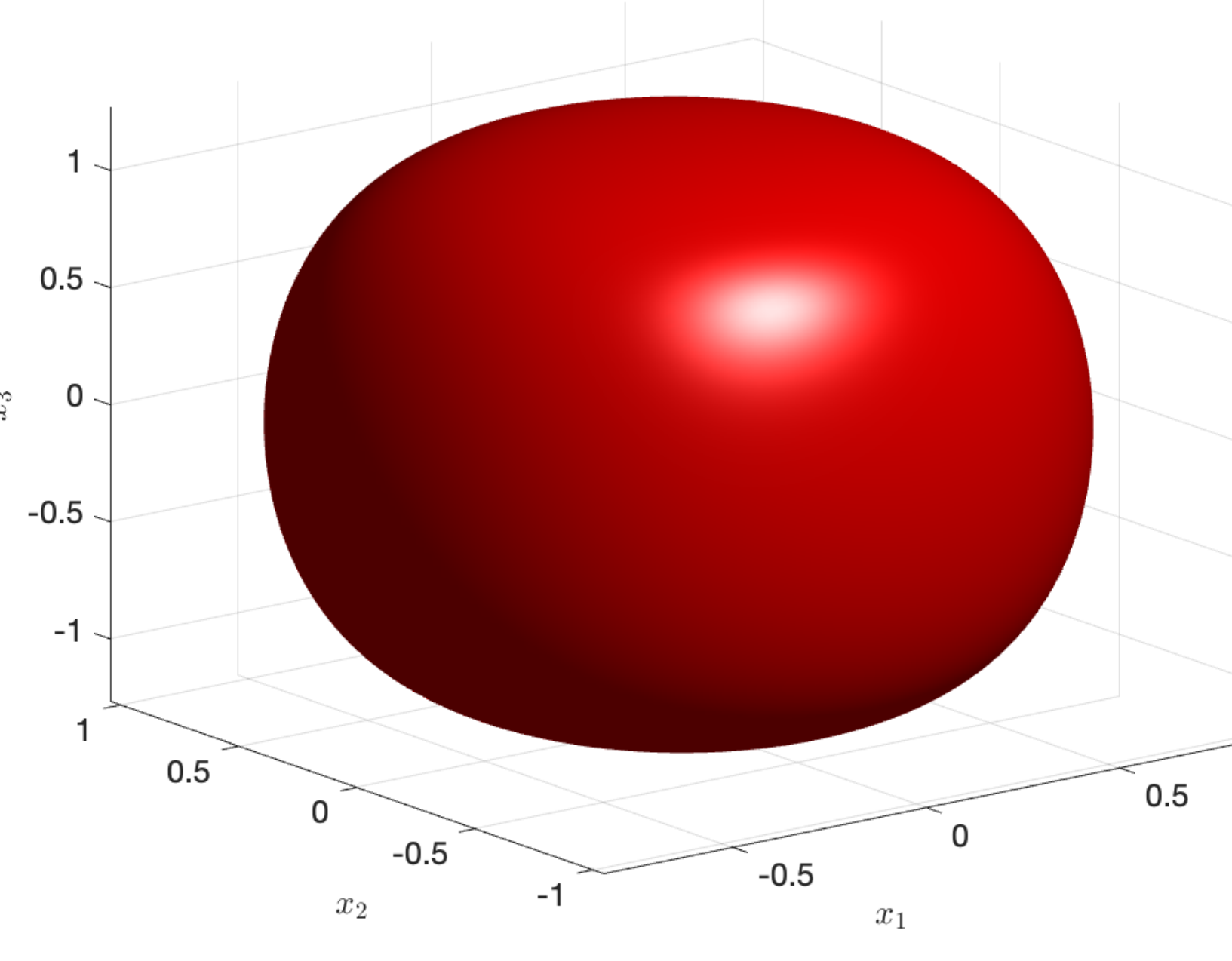}
  
    \includegraphics[trim=2.5cm 7.0cm 3.0cm 8.0cm,clip=true,width=0.32\textwidth]{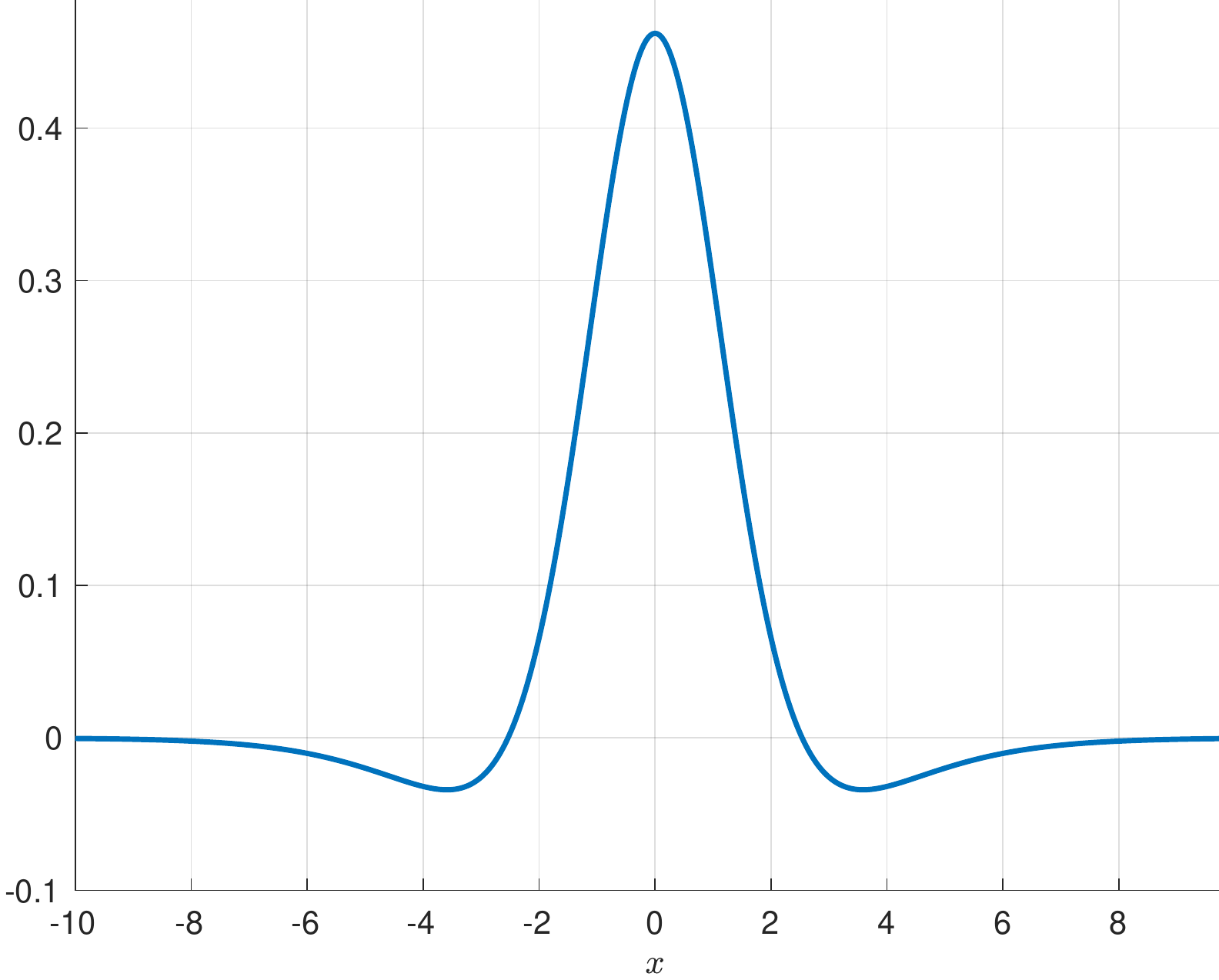}
    \includegraphics[trim=2.0cm 7.0cm 3.0cm
    8.0cm,clip=true,width=0.32\textwidth]{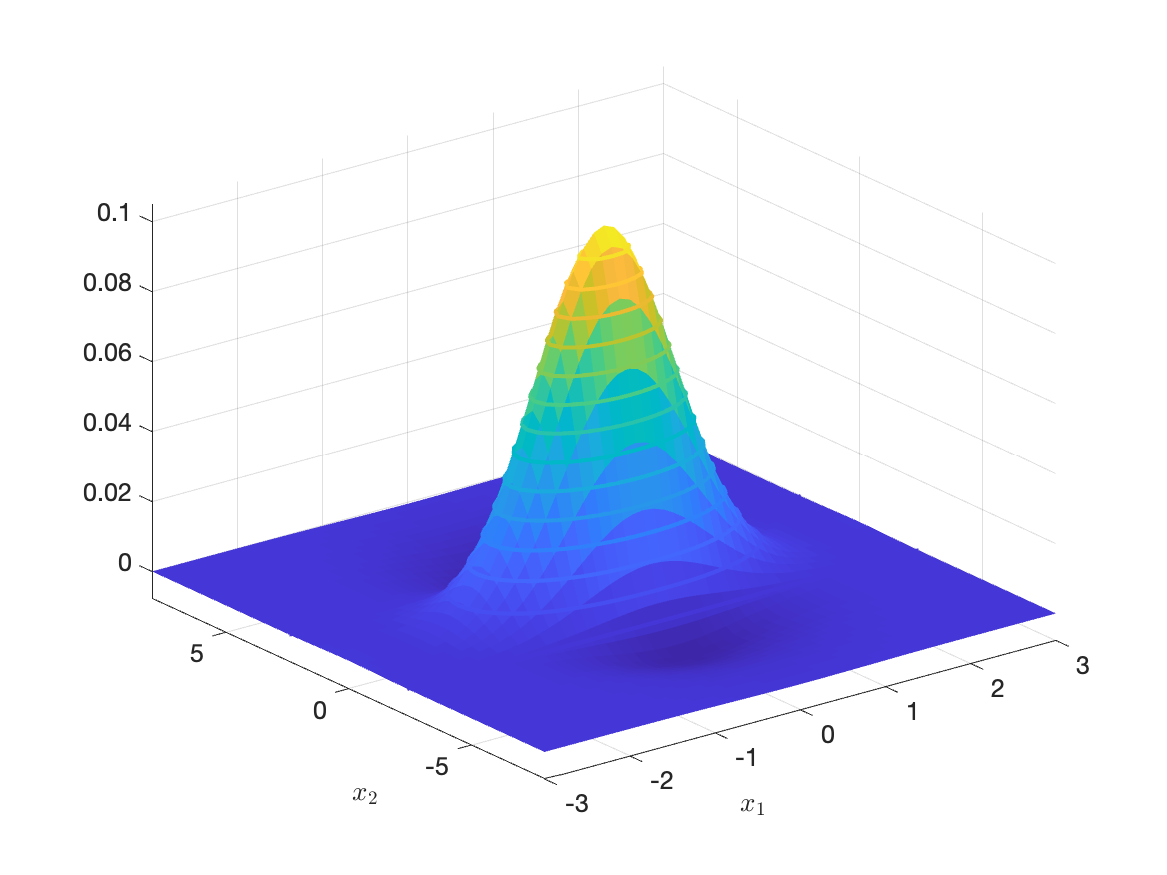}
    \includegraphics[trim=2.0cm 7.0cm 3.0cm 
    8.0cm,clip=true,width=0.32\textwidth]{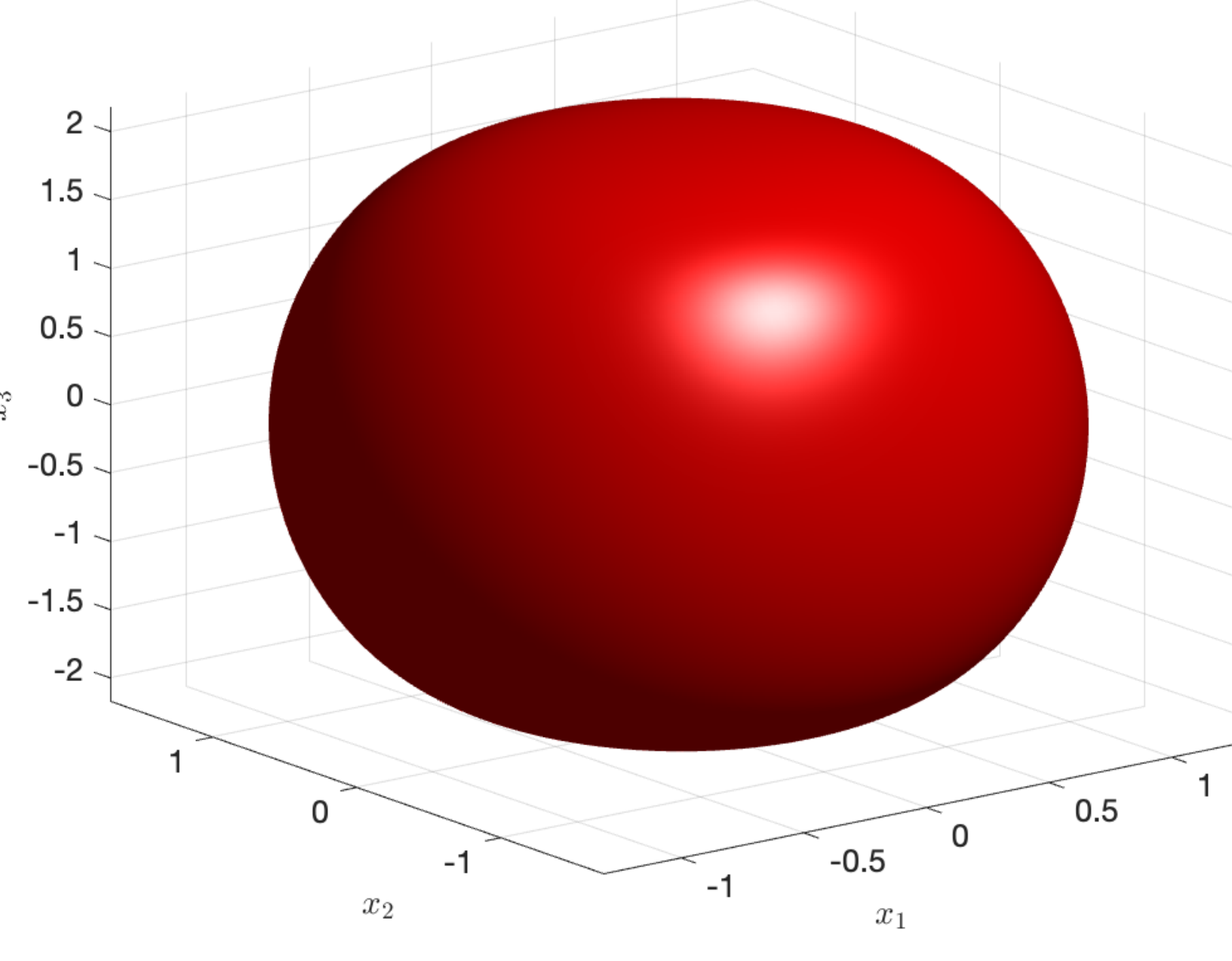}
  	
    \includegraphics[trim=2.5cm 7.0cm 3.0cm 8.0cm,clip=true,width=0.32\textwidth]{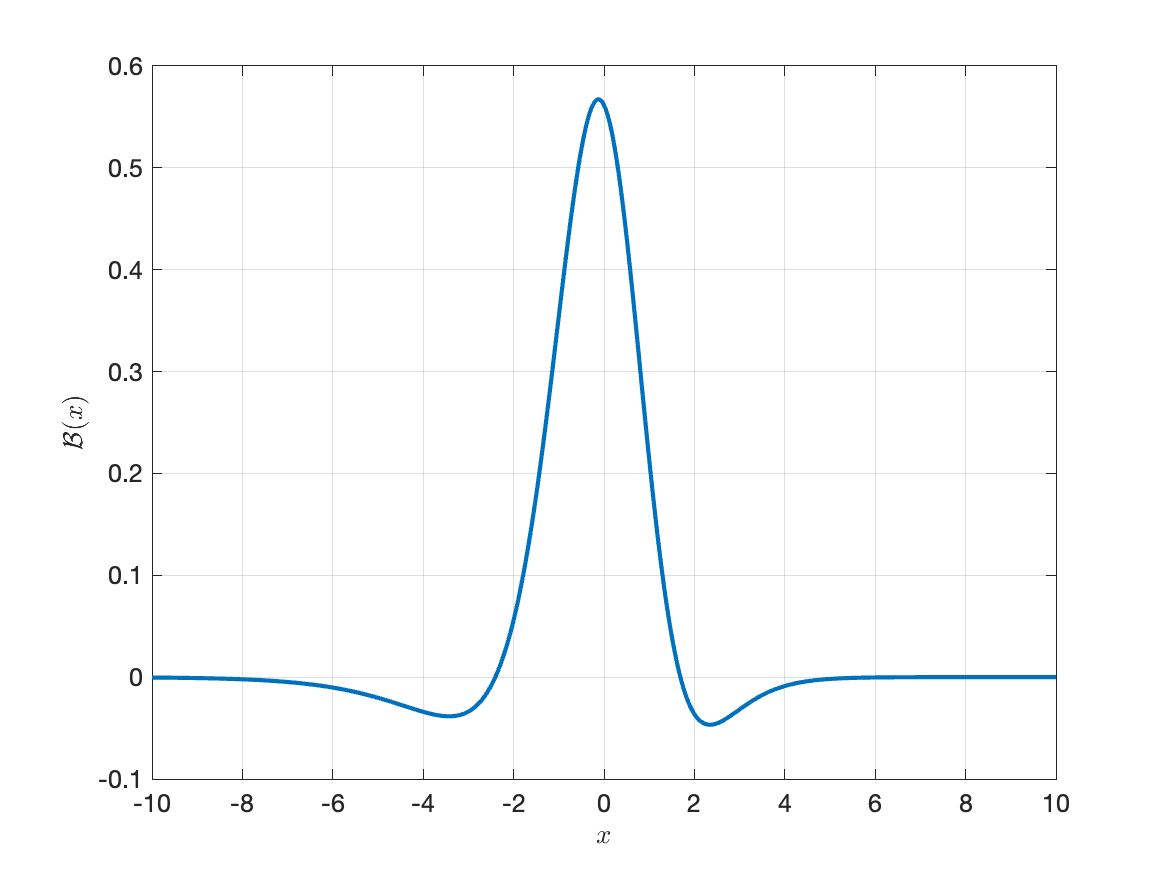}
    \includegraphics[trim=2.0cm 7.0cm 3.0cm
    8.0cm,clip=true,width=0.32\textwidth]{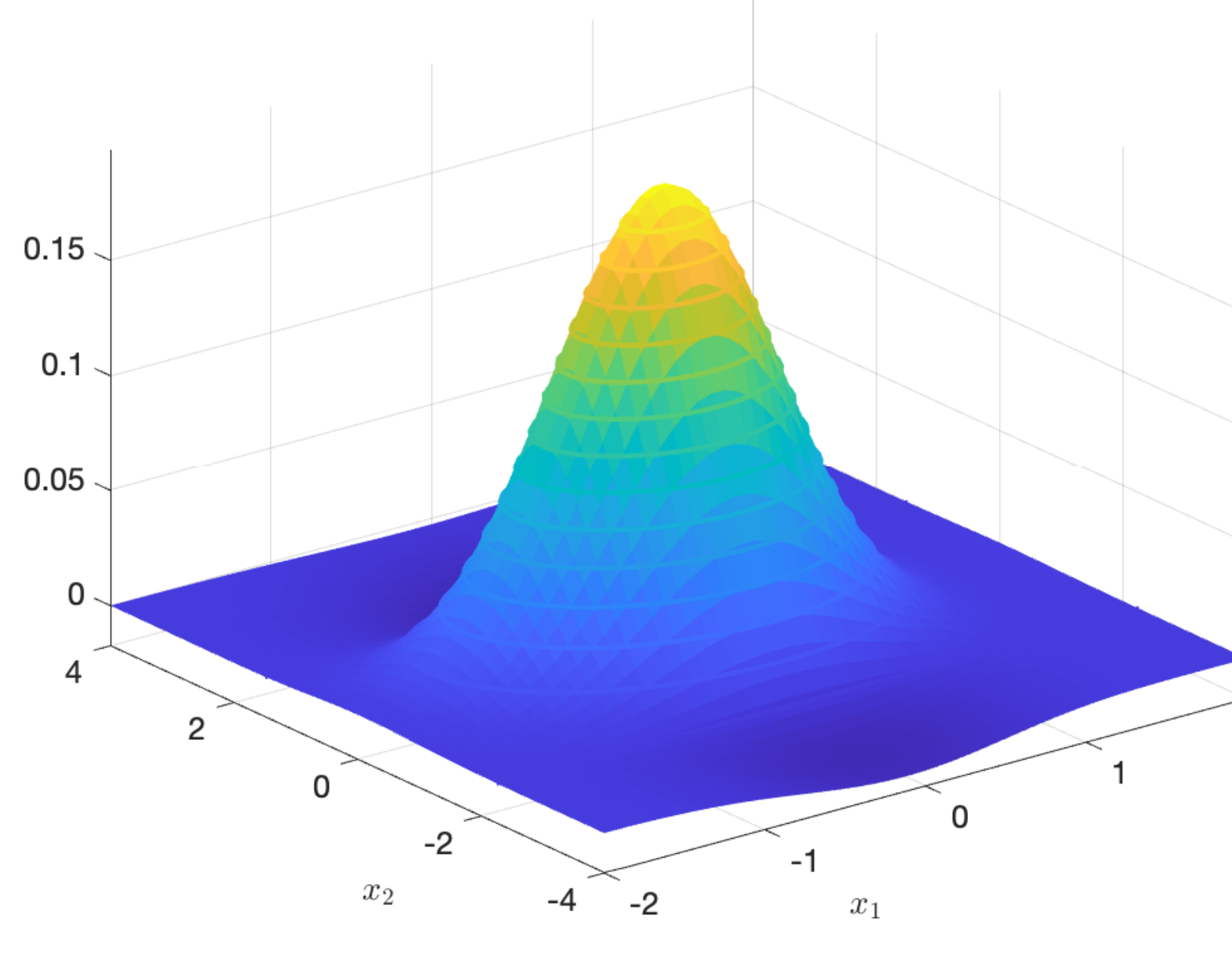}
    \includegraphics[trim=2.0cm 7.0cm 3.0cm 
    8.0cm,clip=true,width=0.32\textwidth]{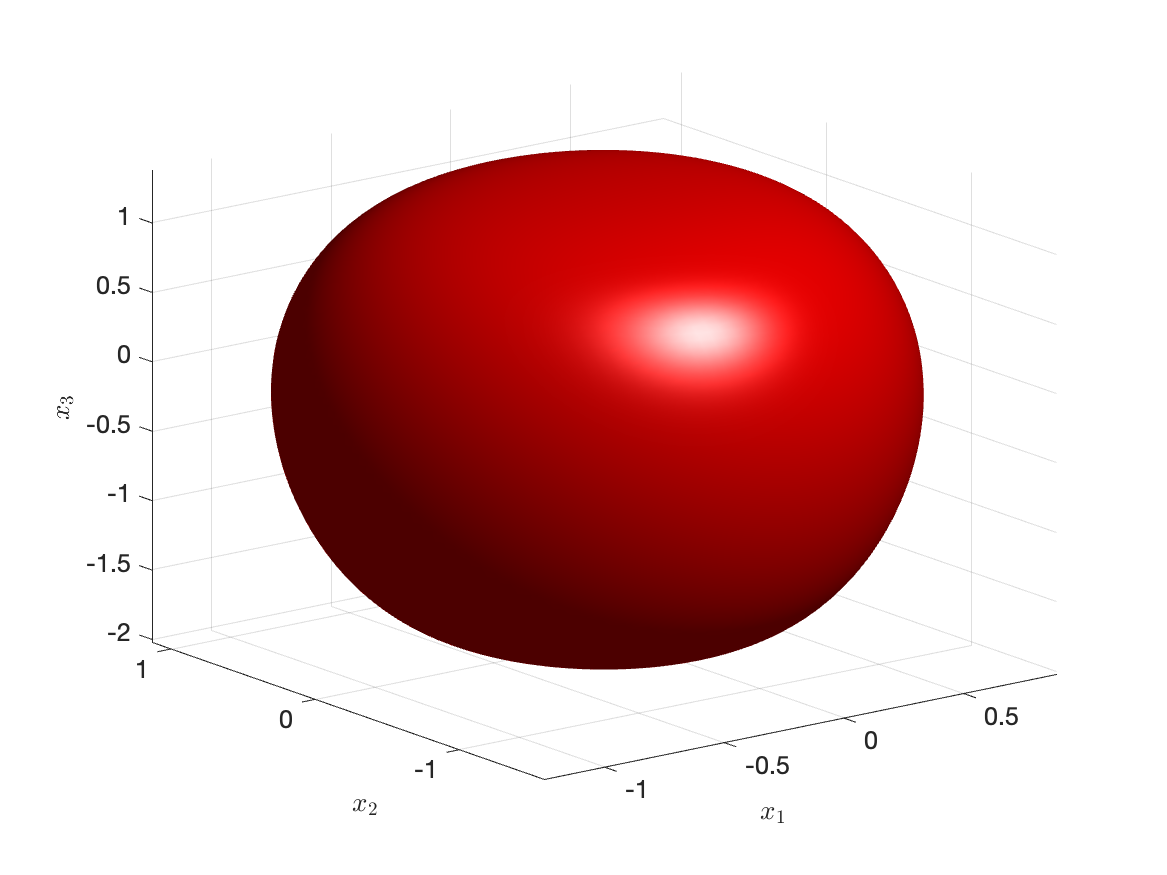}
  		\caption{ From left to right: One dimensional plot, two dimensional surface together with $15$ contours, and a three dimensional isosurface of $\B$ functions. From top row to bottom row: GELU, SiLU, and Mish activation functions.
  	}
  		\figlab{GELUtoMish}
\end{figure}



\bibliographystyle{elsarticle-num} 
\bibliography{references,NewRefs}

\end{document}

%% file: macrosJMLR.tex
\newcommand{\mc}[1]{\mathcal{#1}}

\newcommand{\mbb}[1]{\mathbb{#1}}

\newcommand{\nor}[1]{\left\| #1 \right\|}
\newcommand{\snor}[1]{\left| #1 \right|}
\newcommand{\LRp}[1]{\left( #1 \right)}
\newcommand{\LRs}[1]{\left[ #1 \right]}

\newcommand{\LRc}[1]{\left\{ #1 \right\}}

\newcommand{\bs}[1]{\boldsymbol{#1}}
\newcommand{\floor}[1]{\lfloor #1 \rfloor}

\newcommand{\R}{\mbb{R}}
\newcommand{\intR}{\int_{\R^n}}
\newcommand{\n}{n}
\newcommand{\N}{N}
\renewcommand{\k}{k}
\newcommand{\m}{m}
\newcommand{\p}{p}
\newcommand{\q}{q}
\renewcommand{\r}{r}
\newcommand{\Rn}{\R^\n}
\renewcommand{\L}{\mc{L}}

\newcommand{\C}{\mc{C}}
\renewcommand{\S}{\mc{S}}

\renewcommand{\P}{\mc{P}}
\newcommand{\Q}{\mc{Q}}

\newcommand{\B}{\mc{B}}
\newcommand{\Bc}{\mathfrak{B}}

\newcommand{\Beps}{\B_\theta}
\newcommand{\K}{\mc{K}}
\newcommand{\U}{U}
\newcommand{\X}{X}

\newcommand{\Ind}{\mathds{1}}
\newcommand{\h}{h}
\newcommand{\supp}{\text{supp}}
\newcommand{\T}{\mc{T}}
\newcommand{\RePU}{\text{RePU}}
\newcommand{\half}{{\frac{1}{2}}}
\newcommand{\E}{\mathbb{E}}
\newcommand{\Prob}{\mathbb{P}}

\newcommand{\eval}[2][\right]{\relax \ifx#1\right\relax \left.\fi#2#1\rvert}

\newcommand{\f}{f}
\newcommand{\g}{g}
\newcommand{\deltah}{\delta_{\h}}

\newcommand{\xn}{{x}}
\newcommand{\x}{{\bs{\xn}}}
\newcommand{\yn}{{y}}
\newcommand{\y}{{\bs{\yn}}}
\newcommand{\zn}{{z}}
\newcommand{\z}{\bs{\zn}}
\newcommand{\xib}{{\bs{\xi}}}
\newcommand{\w}{{w}}
\newcommand{\wb}{{\bs{\w}}}
\renewcommand{\b}{{b}}

\newcommand{\figlab}[1]{\label{fig:#1}}
\newcommand{\eqnlab}[1]{\label{eq:#1}}
\newcommand{\theolab}[1]{\label{theo:#1}}

\newcommand{\propolab}[1]{\label{propo:#1}}
\newcommand{\lemlab}[1]{\label{lem:#1}}
\newcommand{\defilab}[1]{\label{defi:#1}}
\newcommand{\remalab}[1]{\label{rema:#1}}

\newcommand{\figref}[1]{\ref{fig:#1}}
\newcommand{\theoref}[1]{\ref{theo:#1}}
\newcommand{\defiref}[1]{\ref{defi:#1}}
\newcommand{\remaref}[1]{\ref{rema:#1}}

\newcommand{\proporef}[1]{\ref{propo:#1}}
\newcommand{\lemref}[1]{\ref{lem:#1}}
\newcommand{\eqnref}[1]{\eqref{eq:#1}}

\newcommand{\seclab}[1]{\label{sect:#1}}
\newcommand{\secref}[1]{\ref{sect:#1}}
